\documentclass[letterpaper, 10 pt, conference]{ieeeconf}  

\usepackage{amsmath} 
\usepackage{amssymb}  
\usepackage{caption}
\usepackage{algorithm}
\usepackage{algpseudocode}
\usepackage{booktabs}
\usepackage[bookmarks=true]{hyperref}
\usepackage{xcolor}
\usepackage{colortbl}
\algrenewcommand\algorithmiccomment[1]{\hfill \textcolor{gray}{\textit{#1}}} 

\usepackage[colorinlistoftodos,prependcaption,textsize=tiny]{todonotes}

\title{\LARGE \bf
Achieving Human Level Competitive Robot Table Tennis
}

\author{David B.\ D'Ambrosio$^{1,*}$, Saminda Abeyruwan$^{1,*}$, Laura Graesser$^{1,*}$, Atil Iscen$^{1}$, Heni Ben Amor$^{2}$, Alex Bewley$^{2}$,\\
Barney J.\ Reed$^{2,\dagger}$, Krista Reymann$^{2}$, Leila Takayama$^{2,\S}$, Yuval Tassa$^{2}$, Krzysztof Choromanski, Erwin Coumans,\\
Deepali Jain, Navdeep Jaitly, Natasha Jaques, Satoshi Kataoka, Yuheng Kuang, Nevena Lazic,\\
Reza Mahjourian, Sherry Moore, Kenneth Oslund, Anish Shankar, Vikas Sindhwani, Vincent Vanhoucke,\\
Grace Vesom, Peng Xu, and Pannag R.\ Sanketi$^{1}$\\
\\%
\textbf{Google DeepMind}\\%
\small $^{1}$Primary contributors, $^{*}$Corresponding authors (order randomized, equal contributions), $^{2}$Core contributors (Alphabetized),\\$^{\dagger}$Work done at Google DeepMind via Stickman Skills Center LLC, $^{\S}$Work done at Google DeepMind via Hoku Labs.
}

\begin{document}

\twocolumn[{%
\renewcommand\twocolumn[1][]{#1}%
\maketitle
\begin{center}
    \centering
    \captionsetup{type=figure}
    \includegraphics[width=0.192\textwidth,trim={0 5cm 0 5cm},clip]{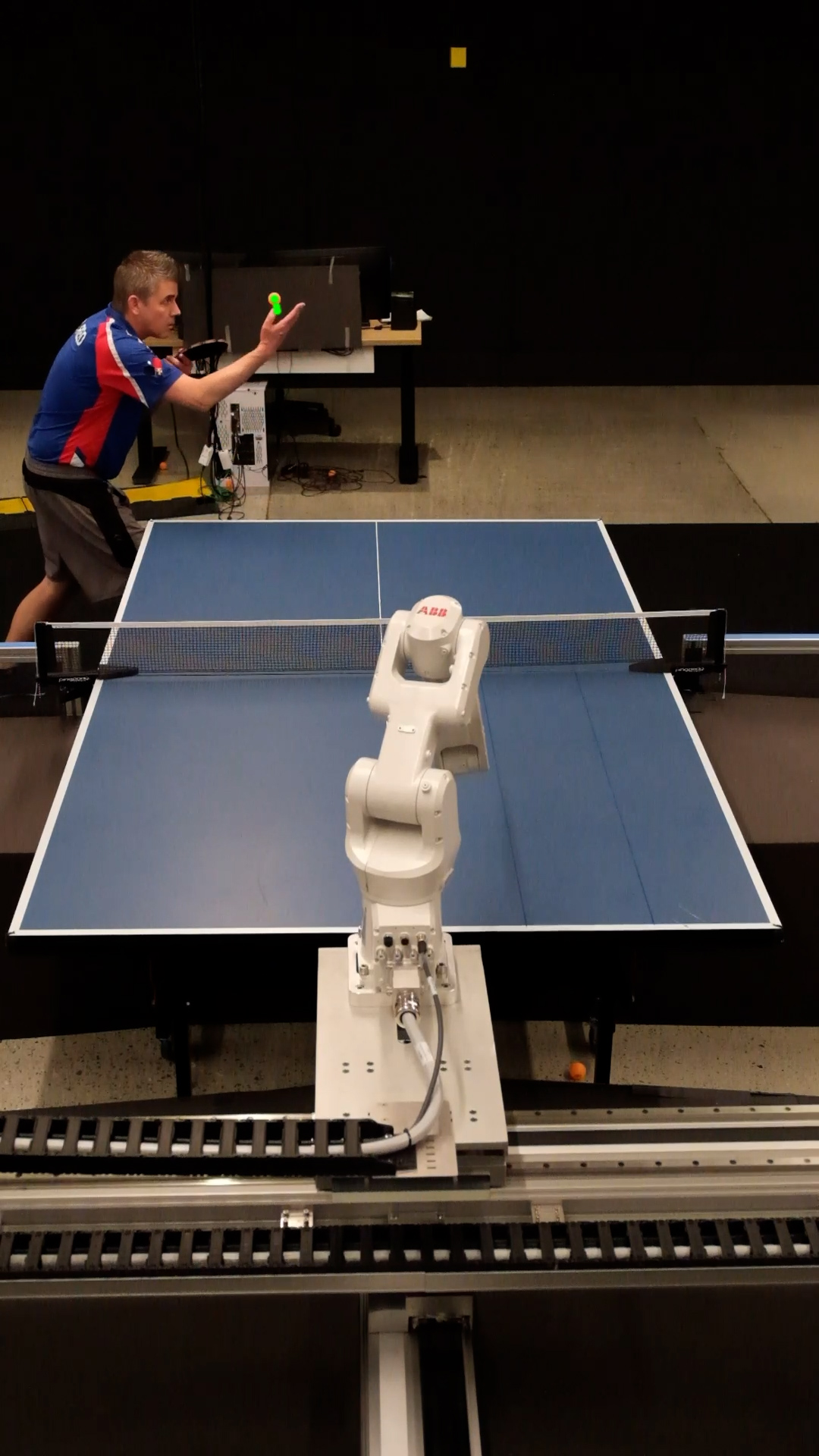}
    \includegraphics[width=0.192\textwidth,trim={0 5cm 0 5cm},clip]{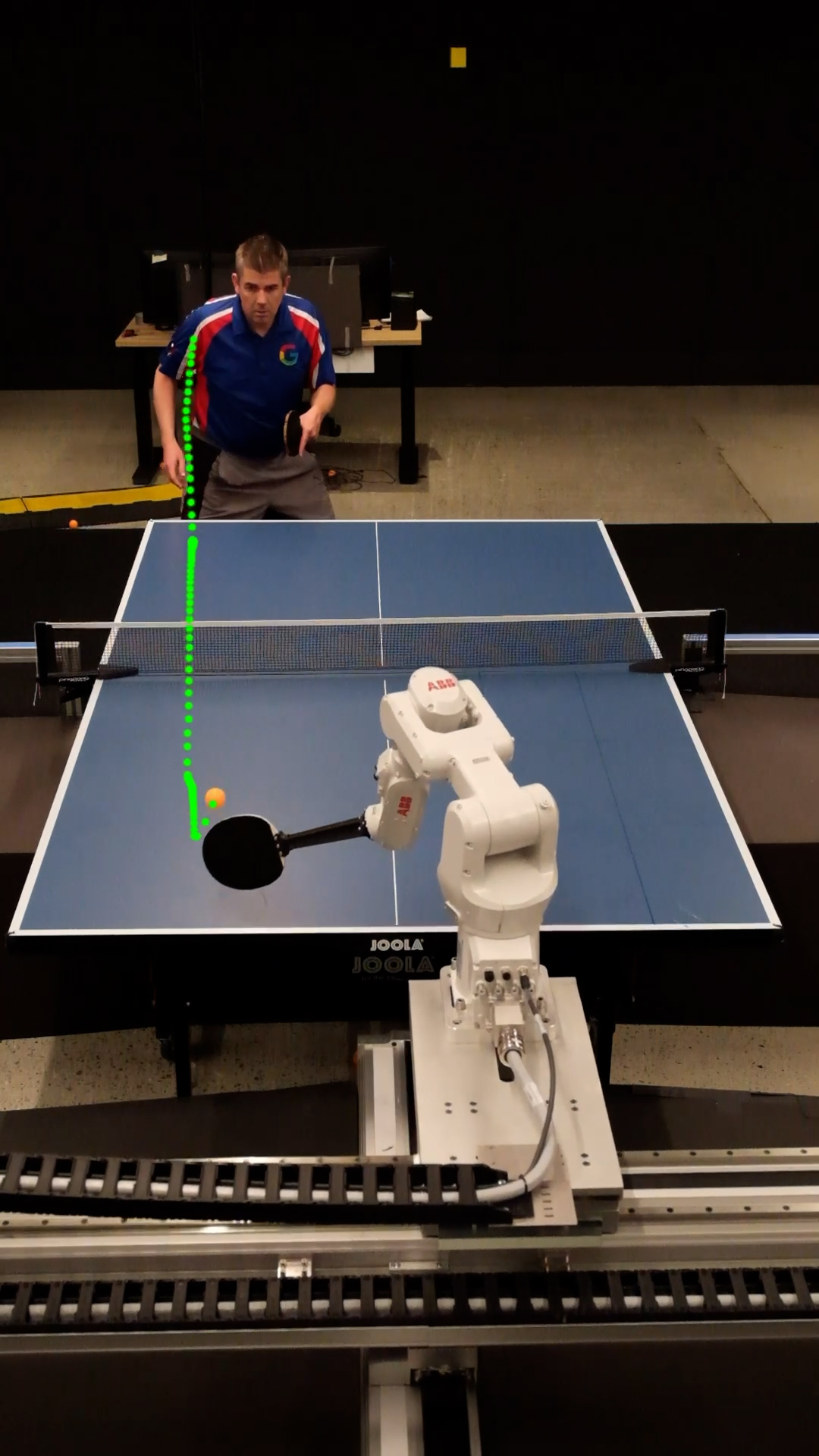}
    \includegraphics[width=0.192\textwidth,trim={0 5cm 0 5cm},clip]{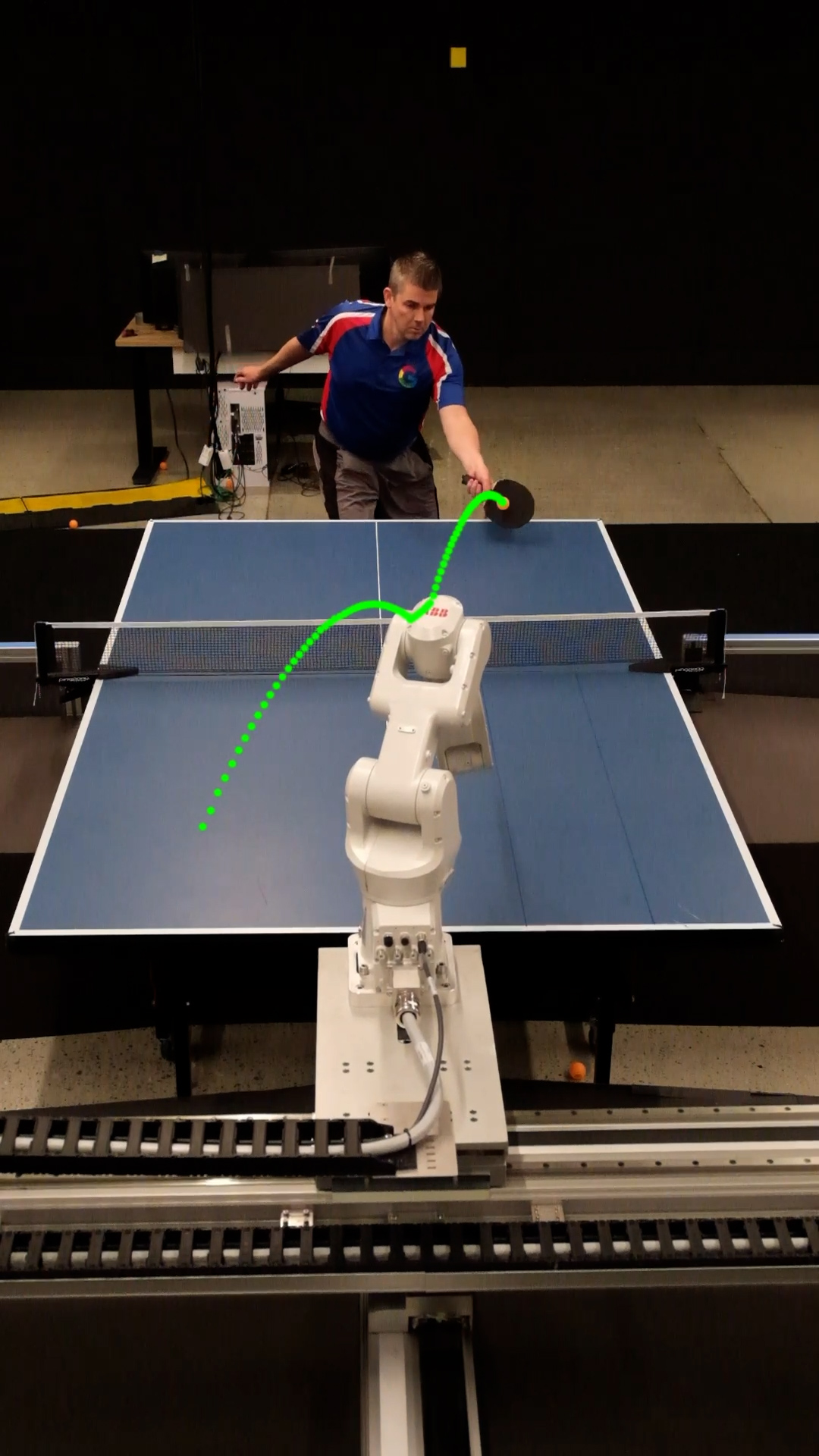}
    \includegraphics[width=0.192\textwidth,trim={0 5cm 0 5cm},clip]{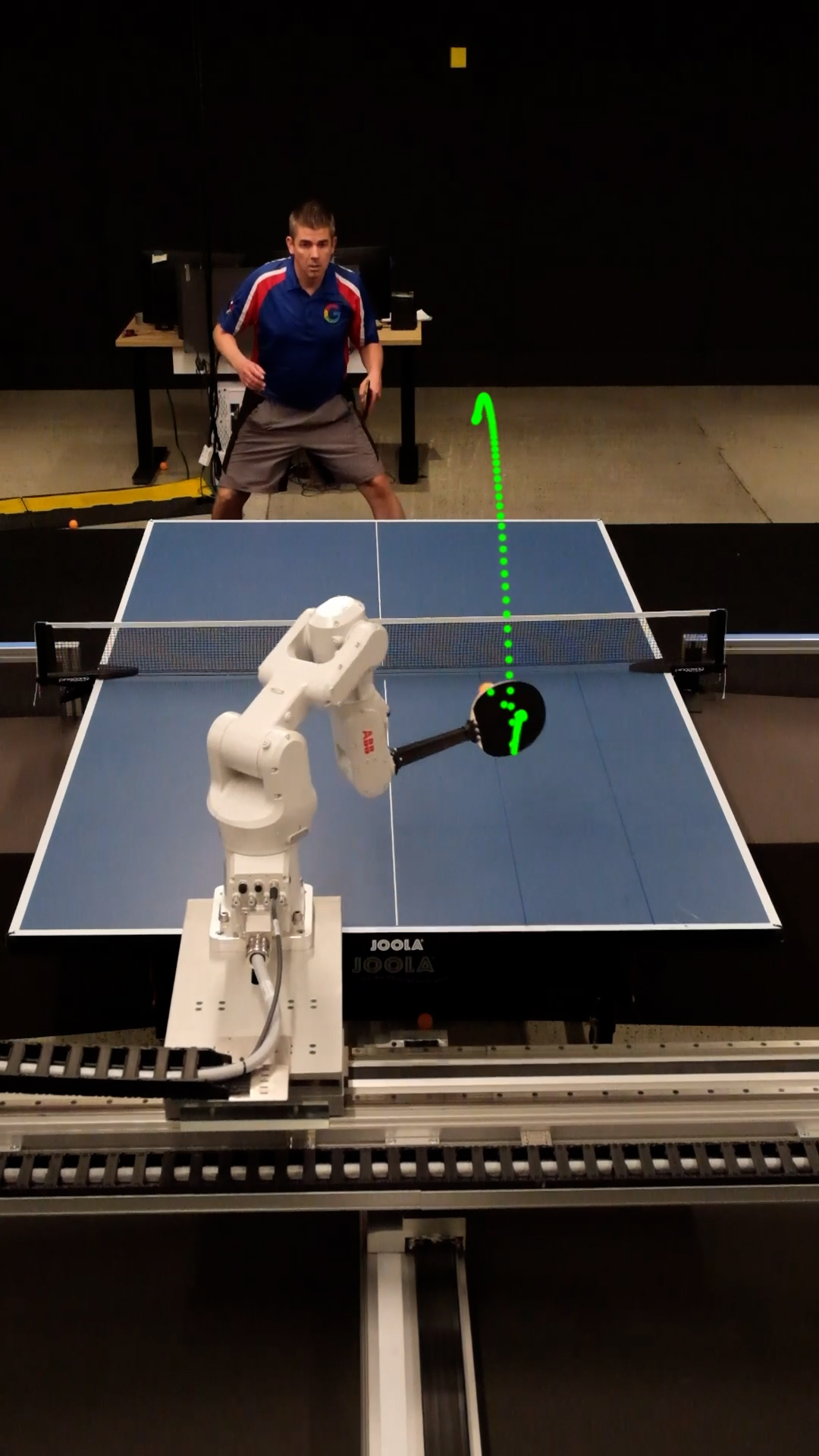}
    \includegraphics[width=0.192\textwidth,trim={0 5cm 0 5cm},clip]{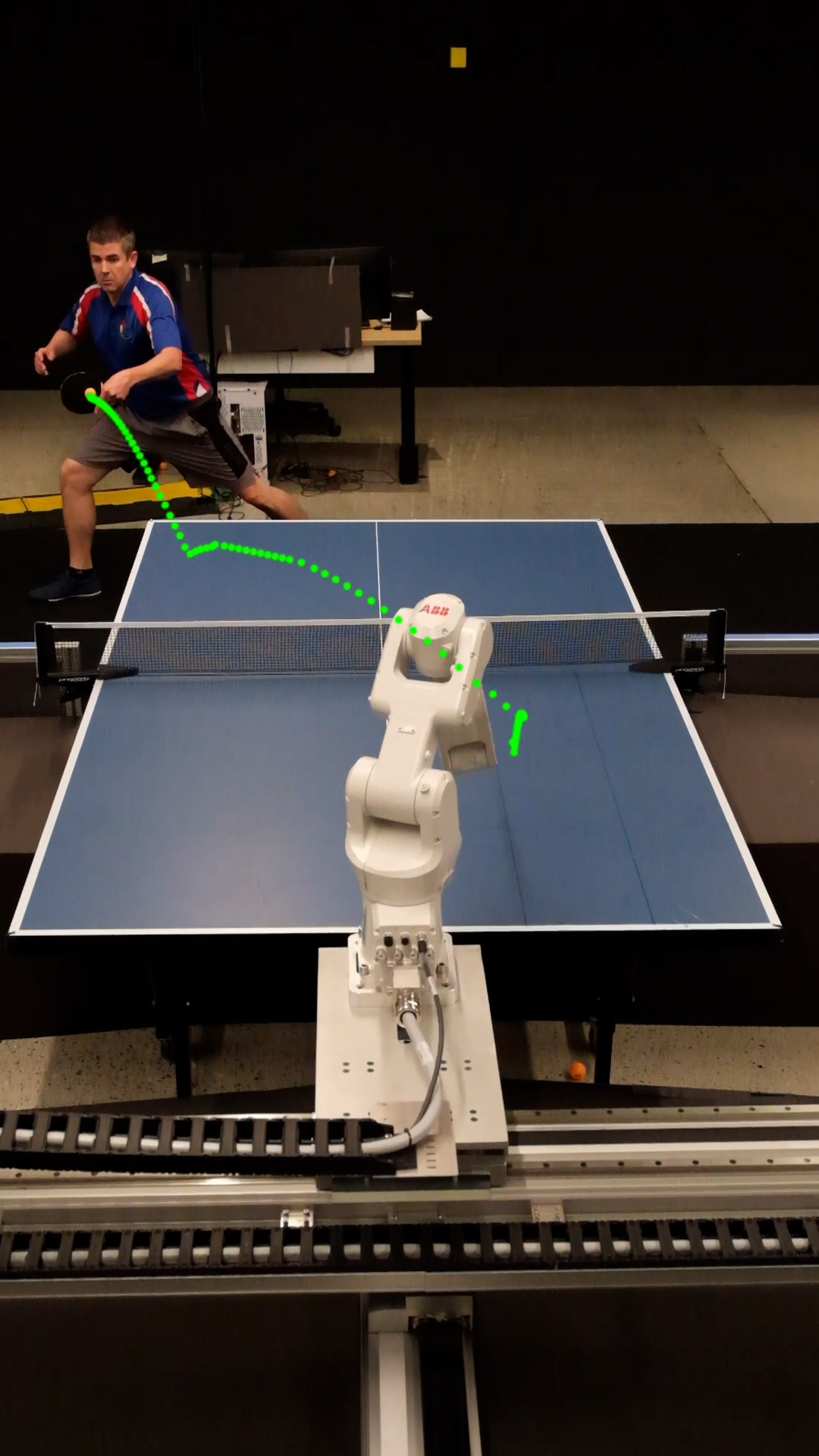}
    \captionof{figure}{Our table tennis robot playing against a professional coach. The green dots show the trajectory of the ball during the rally. The table tennis robot is a 6 DoF ABB 1100 arm mounted on top of two Festo linear gantries, enabling motion in the 2d plane. The x gantry, which moves side to side across the table, is 4m long and the y gantry, which moves towards and away from the table, is 2m long. A 3d printed paddle handle and paddle with short pips rubber is attached to the arm.}
    \label{fig:robot_photo}
\end{center}
}]

\begin{abstract}

Achieving human-level speed and performance on real world tasks is a north star for the robotics research community. This work takes a step towards that goal and presents the first learned robot agent that reaches amateur human-level performance in competitive table tennis. Table tennis is a physically demanding sport which requires human players to undergo years of training to achieve an advanced level of proficiency. In this paper, we contribute (1) a hierarchical and modular policy architecture consisting of (i) low level controllers with their detailed skill descriptors which model the agent's capabilities and help to bridge the sim-to-real gap and (ii) a high level controller that chooses the low level skills, (2) techniques for enabling zero-shot sim-to-real including an iterative approach to defining the task distribution that is grounded in the real-world and defines an automatic curriculum, and (3) real time adaptation to unseen opponents. Policy performance was assessed through 29 robot vs. human matches of which the robot won 45\% (13/29). All humans were unseen players and their skill level varied from beginner to tournament level. Whilst the robot lost all matches vs. the most advanced players it won 100\% matches vs. beginners and 55\% matches vs. intermediate players, demonstrating solidly amateur human-level performance. Videos of the matches can be viewed \textcolor{blue}{\href{https://sites.google.com/view/competitive-robot-table-tennis}{here}}\footnote{See \href{https://sites.google.com/view/competitive-robot-table-tennis}{sites.google.com/view/competitive-robot-table-tennis}}.

\end{abstract}

\section{INTRODUCTION}
Robot learning has made inspiring progress in recent years, resulting in robots that can cook~\cite{fu2024mobile}, clean up~\cite{tidybot2023}, or even perform backflips~\cite{pmlr-v205-li23b}. While the capabilities of learned robot policies have increased dramatically, achieving human-level performance in terms of accuracy, speed and generality still remains a grand challenge in many domains. One such domain is table tennis -- a physically demanding sport which requires human players to undergo years of training to achieve an advanced level of proficiency. Indeed, competitive matches are often breathtakingly dynamic, involving complex motion, rapid eye-hand coordination, and high-level strategies that adapt to the opponent's strengths and weaknesses. For a robotic table tennis system to mimic these abilities it needs high-speed motion, precise control, real-time decision-making and human-robot interaction. Thanks to these demanding requirements, table tennis provides a rich environment to advance robotic capabilities and has served as a benchmark for robotics research since the 1980s \cite{Billingsley83}. Numerous table tennis robots have been developed since and progress has been made on returning the ball to the opponent's side~\cite{Huang2015LearningOS}, 
hitting to a target position~\cite{ding2022learning}, smashing~\cite{LFSD-GT}, cooperative rallying~\cite{abeyruwan2022sim2real}, and many other critical aspects of table tennis~\cite{D_Ambrosio_2023}. \emph{Yet no prior work has tackled the competitive game in which a robot plays a full game of table tennis against a previously unseen human opponent}.

\textbf{In this paper, we present the first learned robot agent that can play competitive table tennis at human level, as depicted in Figure}~\ref{fig:robot_photo}. The robot uses a combination of techniques (known and novel) in order to acquire skills at different levels of abstraction. Table tennis players must be prepared to return balls across a wide variety of positions, speeds, and spins (i.e.\ angular velocities) and competitive players must know how to manipulate these factors to set up advantageous plays or exploit opponent weaknesses.  Thus, there are two levels of play: the high level strategic decisions and the low level physical skills required to execute those strategies.  This organization adds an additional layer of challenge to robotic sports where, unlike a purely strategic game like chess or go, the policy not only needs to decide the most advantageous move, but also needs to have the physical skills to perform it and may even have to choose a less strategically optimal action if it is not confident in successful execution. To address this challenge, we propose a hierarchical and modular policy architecture. Our system consists of multiple low-level skill policies and a high-level controller that selects between them. Each low-level skill policy specializes in a specific aspect of table tennis, such as forehand topspin, backhand targeting, or forehand serve. Training is efficient — each skill builds on top of the same foundation policy for a given category (e.g. forehand, backhand), and once a good skill has been trained it can always be subsequently specialized. In addition to learning the policy itself, we collect and store information both offline and online about the strengths, weaknesses, and limitations of each low-level skill. The resulting \emph{skill descriptors} provide the robot with important information regarding its abilities and shortcomings. In turn, a high-level controller, responsible for orchestrating the low-level skills, selects the optimal skill given the current game statistics, skill descriptors and the  opponent's capabilities.

At its core, our paper aims at scaling robot learning to complex physical tasks which may involve a human partner or adversary. There are two predominant paradigms in the robot learning community. Reinforcement Learning (RL)~\cite{Sutton1998} is the preferred choice for dynamic control tasks, e.g., quadruped locomotion~\cite{caluwaerts2023barkour, kumar2021rma, cheng2023parkour}. Due to its high sample complexity, RL is often first performed in simulation and later transferred to the real world, thereby speeding up training time. However, ensuring the distribution of simulated tasks is grounded in reality can be quite challenging. For example, in table tennis, the set of initial ball conditions at every hit that competitive play will induce is very different from a set of ``all possible" initial ball conditions. By contrast, imitation learning (IL)~\cite{osa2018algorithmic} is the prevalent choice in tasks where demonstrations can be collected from an expert user or process (typically gathered through teleoperation or motion capture), e.g., manipulation~\cite{stepputtis2020language}, \cite{open_x_embodiment_rt_x_2023}. IL from real world data is anchored to examples that are known to solve the desired task. Additionally it bypasses the challenges of simulating the task(s) and the inevitable sim-to-real gap. However, it typically requires multiple demonstrations for every single task~\cite{osa2018algorithmic} and may not cover critical parts of the state space. In this work, we introduce a hybrid training method that combines the best of both worlds. We collect a small amount of human-human play data to seed the initial task conditions. We then train an agent in simulation using RL and employ a number of techniques (known and novel) to deploy the policy zero-shot to real hardware. This agent plays with humans to generate more training task conditions and the training-deployment cycle is repeated. As the robot improves, the standard of play becomes progressively more complex whilst remaining grounded in real-world task conditions. This approach iteratively refines the skill repertoire based on real-world data. As the robot plays, it gathers data and gaps are revealed in its capabilities, which are then addressed through continued training in simulation. This hybrid sim-real cycle creates an automatic task curriculum and enables the robot's skills to improve over time.

A final, yet critical, ingredient of our approach is the ability to rapidly adapt to the unseen human opponent's capabilities and play style. \cite{kumar2021rma} adapts to various environmental conditions for locomotion but not to humans or adversarial conditions. We enable rapid adaptation to the opponent by tracking in real-time the match statistics representing the robot's and opponent's strengths and weaknesses. Additionally, for each low-level skill we estimate \textit{preferences} \cite{Sutton1998} online to augment the offline skill statistics. These two are used to derive the strategy. This online adaptation helps the controller to adapt to novel opponents and allows the robot to learn and refine its decision-making process, leading to improved robustness against a diverse set of human opponents.

\begin{figure*}[!t]
    \centering
    \includegraphics[width=0.99\textwidth]{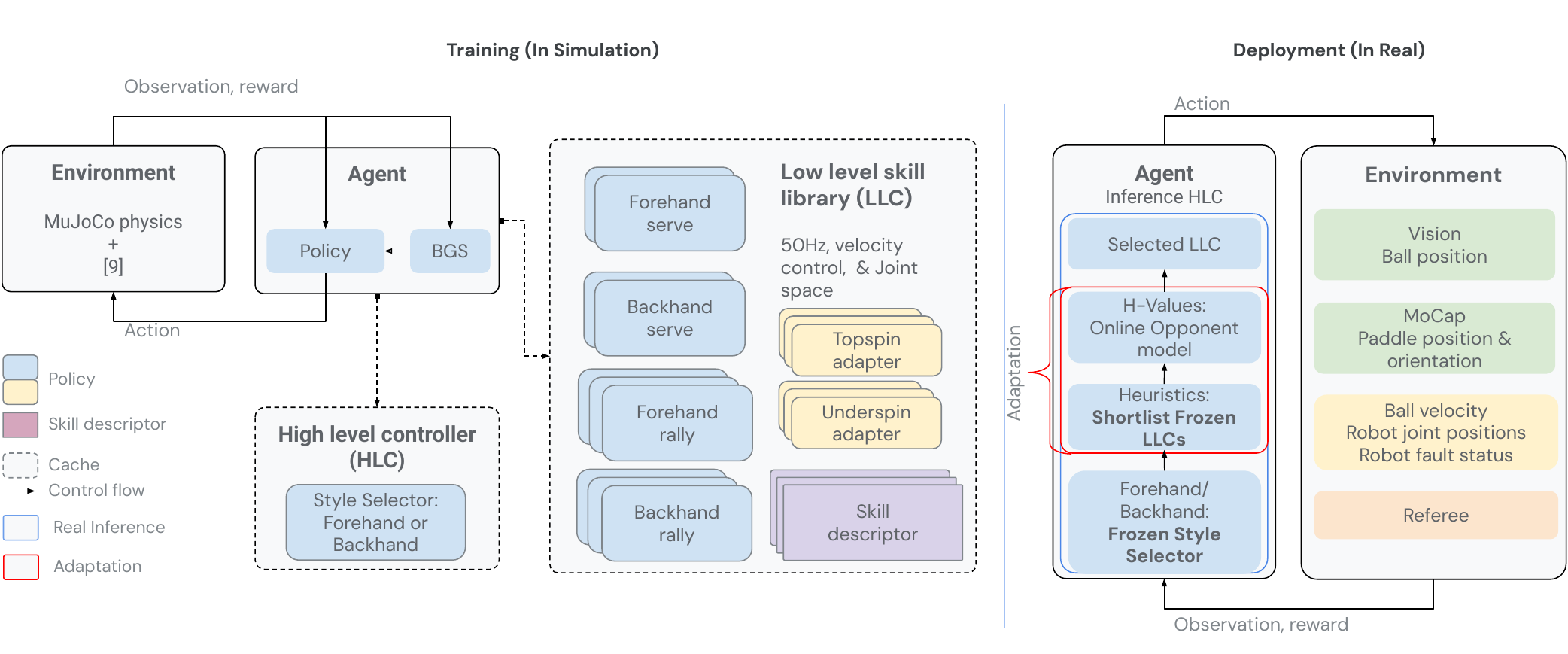}
    \caption{Method overview. We train a skill library of low-level controllers (LLCs), including serving and rallying, and sim-to-sim adapters from a dataset of ball states. Using the same ball states, we train a high level controller (HLC) for style selection. The policies are trained purely in simulation (but using real ball states) using Blackbox Gradient Sensing (BGS) \cite{D_Ambrosio_2023}, \cite{abeyruwan2022sim2real}. The policies transfer zero-shot to the physical world. At deployment time, we freeze the style selector and skills. During inference HLC uses the style selector to select the side. The heuristics module shortlists the most effective skills. H-values (online opponent model) select the most preferred skill, and the skill executes the actions.
}
    \label{fig:system_digram}
\end{figure*}

The components described above ultimately lead to competitive gameplay at human level that humans actually enjoy playing with — the resulting policy is flexible, adaptable, extensible, and more interpretable than a monolithic system. In summary, in this paper we introduce the first robot learning system to achieve amateur human level performance in an interactive competitive sport against unseen human opponents demonstrated through a user study. To achieve this goal, we make the following technical contributions (1) a hierarchical and modular policy architecture, specifically the (i) low level controllers with their skill descriptors and (ii) a high level controller that chooses the low level skills; (2) techniques to enable zero-shot sim-to-real including an iterative approach to defining the training task distribution that is grounded in the real-world and defines an automatic curriculum, (3) real time adaptation to unseen opponents and (4) A user-study to test our model playing actual matches against unseen humans in physical environments.

\section{METHOD}
\subsection{Hardware, problem setting, and environment}
\label{method-setup}

Figure \ref{fig:robot_photo} depicts the physical robot. The table tennis robot is a 6 DoF ABB IRB 1100 arm mounted on top of two Festo linear gantries, enabling motion in the 2d plane. The x gantry, which moves side to side across the table, is 4m long and the y gantry, which moves towards and away from the table, is 2m long. A 3D printed paddle handle and paddle with short pips rubber \cite{Globe889_2024} is attached to the ABB arm. A pair of Ximea MQ013CG-ON cameras operating at 125Hz capture images of the ball and these are used as input into a neural-perception system \cite{D_Ambrosio_2023} which produces ball positions at the same frequency. We use a PhaseSpace motion capture system consisting of 20 cameras mounted around the play area to track the human opponent's paddle. 

We model table tennis as a single-agent sequential decision making problem in which the human opponent is modeled as part of the environment. We use the \textit{Markov Decision Process} (MDP) \cite{puterman2014markov} formalization. This consists of a of a 4-tuple ($\mathcal{S}$, $\mathcal{A}$, $\mathcal{R}$, $p$), whose elements are the state space $\mathcal{S}$, action space $\mathcal{A}$, reward function $\mathcal{R}: \mathcal{S} \times \mathcal{A} \rightarrow \mathbb{R}$, and transition dynamics $p: \mathcal{S} \times \mathcal{A} \rightarrow \mathcal{S}$. An episode $(s_0, a_0, r_0, ..., s_n, a_n, r_n)$ is a finite sequence of $s\in\mathcal{S}$, $a\in\mathcal{A}$, $r\in\mathcal{R}$ elements, beginning with a start state $s_0$ and ending when the environment terminates. An episode consists of a single incoming ball, which is hit and returned, beginning at the moment the opponent's paddle contacts the ball and ends when either of the following conditions occur 1.) the robot returns the ball,  2) the ball goes out of play, or 3) the robot misses the ball. A ball return means that the robot hits the ball such that it bounces on the opponent's side of the table without first bouncing on the robot's side. The objective for the robot is to maximize the expected return rate over the ball distribution.

The real environment consists of the neural-perception system which tracks the ball, the motion capture system which tracks the human player paddle pose, a state machine that tracks the state of the game, and an observation module which provides data such as ball position and velocity, robot position, etc., to the policy. In addition, we also built a corresponding simulation environment built on top of the MuJoCo \cite{conf/iros/TodorovET12} physics engine. A detailed description of the basic system can be found in \cite{D_Ambrosio_2023}. Below, we describe changes to this system that were made to enable real-time competitive play with humans.

\subsection{Hierarchical agent architecture and training overview}

Our agent architecture and approach to training are both designed to address the numerous challenges presented by competitive table tennis with humans. The table tennis agent shown in Figure \ref{fig:system_digram} consists of two levels of control which we refer to as the high level controller (HLC) and the low level controllers (LLCs). LLCs are policies representing different table tennis skills and are trained to produce joint velocity commands at 50Hz. For example LLCs may represent playing with a forehand style and hitting cross-court balls, playing backhand conservatively, or playing forehand to return underspin serves.

The HLC is responsible for selecting which LLC should be executed every incoming ball episode. The HLC does not have a fixed control frequency but instead is triggered to act once every time the opponent hits the ball. Within the HLC, there are six components that are combined to produce the choice of LLC; \textbf{(1) \textit{Style:}} this is a policy trained to choose the play-style, forehand or backhand given the incoming ball, \textbf{(2) \textit{Spin classifier:}} this classifier provides information about the spin of the incoming ball identifying topspin or underspin, \textbf{(3) \textit{LLC skill descriptors:}} these are a model of the agent's own capabilities. They provide performance metadata for every LLC such as estimated return rate, ball hit velocity, and land position, conditioned on the specific incoming ball, \textbf{(4) \textit{Match statistics}} about the opponent and robot's performance, \textbf{(5) \textit{Strategies:}} these take (1), (3) and (4) as input and output a shortlist of LLCs, and \textbf{(6) \textit{LLC preferences (H-values):}} these estimate the performance of each LLC for the current player and are updated after every shot. The HLC combines (3), (5), and (6) to produce the final choice of LLC. The entire control flow within the HLC happens within 20ms.

We chose to train multiple, modular LLCs instead of a single monolithic LLC for a number of reasons: \textit{Avoiding catastrophic forgetting} --- once a good skill has been learned it is never forgotten, whilst still serving as an initialization point for further skill learning. \textit{Extensibility} --- new skills are straightforward to incorporate by adding a new LLC. \textit{Evaluation efficiency} --- which in turn speeds up experimental velocity. Once a low level skill has been tested in the real world, its capabilities are well understood and it does not have to be re-tested. In contrast a monolithic learned system will need to be tested on the full suite of expected capabilities every time the model weights change. \textit{Fast inference} --- inference for each LLC takes 3ms on a CPU.

The LLCs and HLC style policy were trained iteratively, alternating between simulated training and zero-shot deployment in the real world during which human opponents play with the robot. The human-robot interactivity inherent in the task motivated doing all training entirely in simulation. Fine-tuning complex skills with humans in the loop in the real world is too time consuming to be feasible, especially since prior work  \cite{abeyruwan2022sim2real} showed it took 6 hours of real-world fine tuning to train a policy to cooperate with a single human. This led to a substantial effort to narrow the sim-to-real gap.

\subsection{LLC training}
\label{method-llc-training}

\begin{figure*}[!t]
    \centering
    \includegraphics[width=0.99\textwidth]{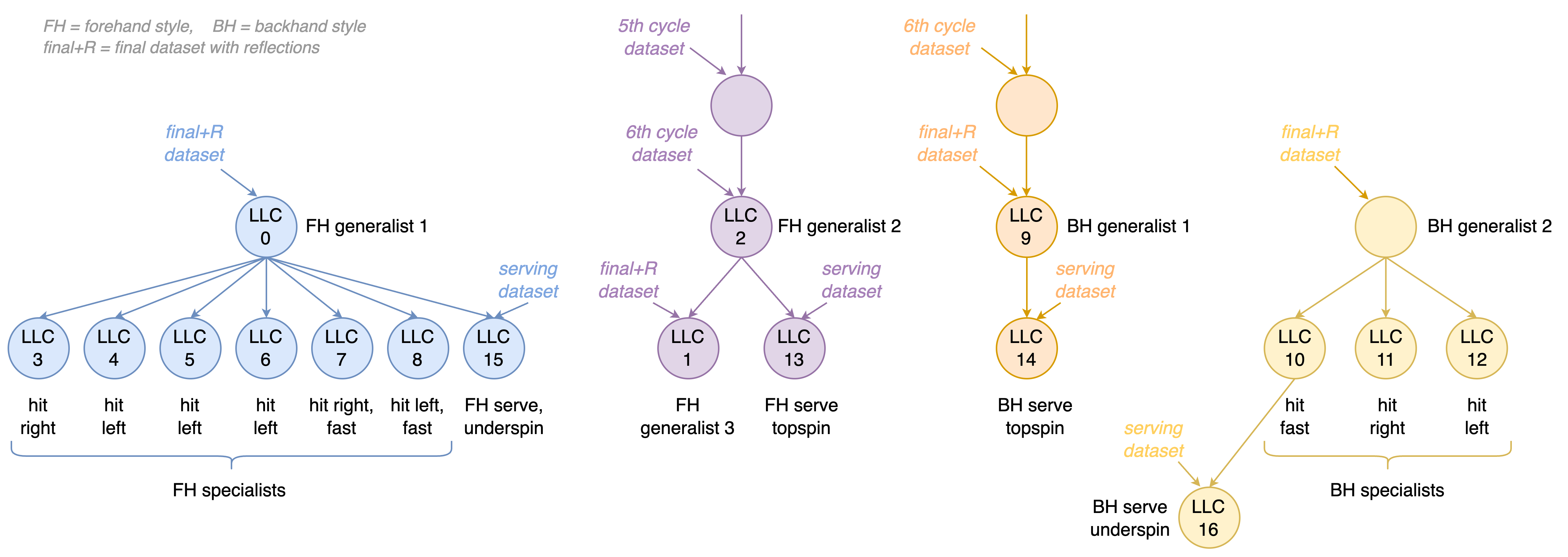}
    \caption{LLC training lineage. LLC x = ID of the LLC in the final system. The forehand (FH) and backhand (BH) LLCs were each developed from two independently trained generalists. One of the generalists was developed along with the dataset cycles, whilst the other was trained only after finalizing the dataset. Both the seed forehand generalists were deployed (LLC 0 and LLC 2) whilst for the backhand only one of the seed generalists was deployed (LLC 9).}
    \label{fig:llc-lineage}
\end{figure*}

LLCs are meant to provide a library of skills that our HLC can deploy in its strategies. Our approach to training them can be summarized in three steps. \textbf{(1)} Train two generalist base policies, one for each main play style (forehand, backhand) and add this to the set of LLCs. \textbf{(2)} Specialize policies to different skills by adding reward function components and / or adjusting the training data mix before fine-tuning a new policy initialized from one of the existing LLCs. This is typically one of the generalist base policies but could be any policy in the LLC set. \textbf{(3)} Evaluate the new policies and assess if a policy exhibits the desired characteristics. For example, if a policy is trained to target a particular location on the table, calculate the average error between the ball landing position and target. If successful, add the policy to the set of LLCs.

\textbf{Training algorithm} All policies were trained in simulation with Blackbox Gradient Sensing (BGS) \cite{abeyruwan2022sim2real}, an evolutionary strategies (ES) algorithm, on the task described in Section \ref{method-setup}. The training task distribution of initial ball states is sampled from a real world dataset, gathered iteratively through multiple cycles of policy training and real world evaluations. Creating this dataset is discussed in detail in Section \ref{method-zs-s2r}. We chose BGS because we observed that it produced policies with relatively smooth actions, whereas policies trained with RL algorithms such as PPO \cite{ppo} or SAC \cite{sac-18} produced noticeably jerkier actions. Additionally, it has been shown to have strong sim-to-real transfer performance \cite{D_Ambrosio_2023}. We hypothesize that action smoothness and potentially less overfitting to the simulator are the main reasons why BGS-trained policies exhibit such good transfer.

\textbf{Network architecture} Each policy is a dilated-gated CNN \cite{gated-cnn} following the architecture in \cite{GaoPPOES2020} with 10k parameters and an optional FILM adapter layer of 2.8k parameters designed to aid sim-to-real transfer (see Section \ref{method-zs-s2r}). The CNN contains 1D convolutions, convolving across timesteps. \cite{GaoPPOES2020} found this accelerated learning and led to smoother outputs. The observation space is $(8, 16)$ consisting of 8 consecutive timesteps of ball position and velocity (6), robot joint position (8), and one-hot style; forehand or backhand (2). The style component is an artifact of early experiments and likely could be removed without affecting performance. All policies output actions with dimension $(8,)$ representing joint velocities at 50Hz. 8 timesteps is 0.14 seconds of history which was empirically determined to be sufficient to smooth out noise in the trajectory and give context to the current state.

\textbf{Training generalist base policies} To train for a particular style (forehand, backhand), each ball state in the dataset was annotated with \textit{forehand}, \textit{backhand}, or \textit{center} based on where the ball trajectory intersected with the back of the table on the robot side. \textit{center} was defined as +/- 0.2m around the center of the table, \textit{forehand} as $> 0.2$, and \textit{backhand} as $< -0.2$. Forehand policies were trained on only \textit{forehand} + \textit{center} balls, backhand on \textit{backhand} + \textit{center}. This created an overlap in the center where policies of either style are capable of returning the same balls. The policy was also rewarded for moving towards a reference pose (either forehand or backhand) at the beginning of the shot. Without such a reward we observed that the robot would sometimes employ a backhand pose to hit forehand balls even though it was less efficient.  These base policies are important, not only to have a strong starting polices capable of returning a wide range of balls to branch from, but also to anchor play in specific styles for efficient returns.  Base polices were trained for roughly 2.4 billion simulation steps across 6,000 parallel simulation workers.

\textbf{Training specialists} We experimented with the types of skills to train for based on advice from a table tennis coach and general game intuition, including targeting specific return locations, maximizing return velocity, and specializing to return serves of either topspin or underspin, fast balls, and lobs. We found that we did not need a specialist to handle lobs, and were unable to train a specialist on fast balls due to lack of data and hardware limitations. We therefore focused on developing serving, targeting and fast hitting specialists in addition to the generalists.  Specialists were further trained for roughly 300-1200 million additional simulation steps depending on convergence.

\textbf{Determining the total set of skill policies} The final system contained 17 LLCs. 4 were specialized for returning serves, 13 for rallying. 11 played with a forehand style, 6 with a backhand style. Importantly, each policy had the same initial robot pose, enabling straightforward sequencing of LLC choices, since the initial robot pose will be in-distribution for all LLCs. The training lineage along with brief descriptions of each LLC is shown in Figure \ref{fig:llc-lineage}. We kept training LLCs until we had covered our target set of skills --- consistent generalists, targeting, fast hits, topspin and underspin serves. Beyond that it was desirable to have varied playstyles and to provide options to the HLC. ES training meant that policies trained with the same objective could exhibit different behaviors, and due the modular architecture, there was little downside in including additional LLCs. Thus, if we had a strong LLC, we included it, even if there was already an LLC covering that particular skill. Forehand training produced a wider set of viable LLCs and thus more were included.

\begin{figure*}[!t]
    \centering
    \includegraphics[width=0.99\textwidth]{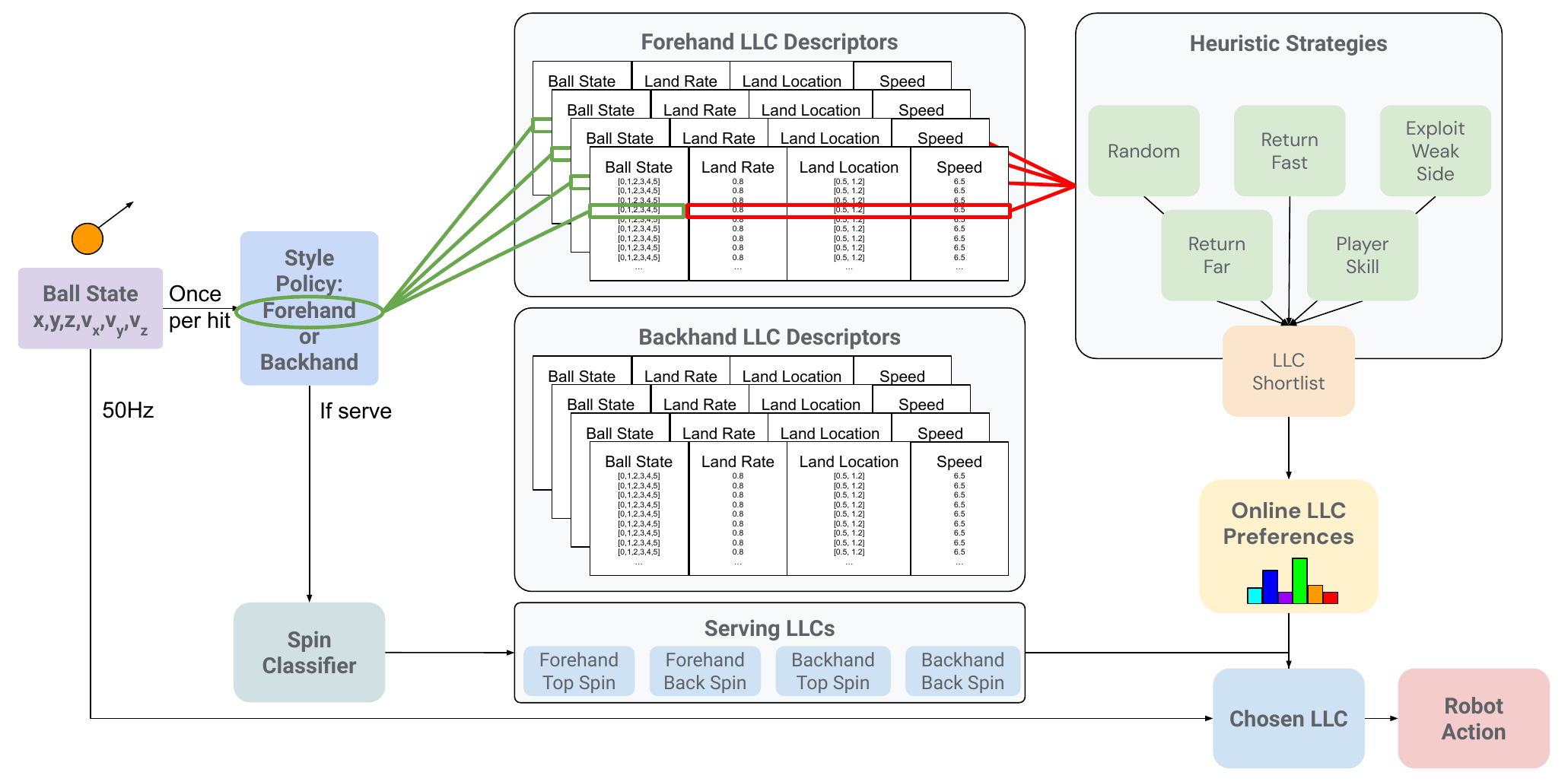}
    \caption{Once per ball hit, the HLC decides which LLC to return the ball with by first applying a style policy to the current ball state to determine forehand or backhand (in this example we demonstrate choosing forehand). If the ball is a serve it will attempt to classify the spin as topspin or underspin and pick the corresponding LLC.  Otherwise it must determine which of the many rallying LLCs will perform best by finding the most similar ball state within the corresponding set of LLC skill tables and getting the return statistics.  Heuristic strategies are applied to these statistics and produce a shortlist of candidate LLCs and the final LLC is chosen through a weighted selection.  The chosen LLC will be queried at 50Hz with the current ball state to determine the robot actions.  }
    \label{fig:inference}
\end{figure*}

\subsection{The High Level Controller (HLC)}
\label{method-hlc}

\begin{algorithm}
\caption{Pseudocode for HLC inference}
{\footnotesize
\begin{algorithmic}[1]
\ttfamily
\Procedure{hlc\_act}{$B, S, \pi^{St}_\theta, \pi^{Sp}_\theta, o^H, o^{Sp}, \Lambda, \Gamma, \Psi, H$}
\State \hspace{0.3cm}
\State \textbf{Input:}
\State \hspace{0.3cm} $B$: $[b_x, b_y, b_z, \dot{b}_x, \dot{b}_y, \dot{b}_z]$, ball state
\State \hspace{0.3cm} $S$: is the ball a serve?
\State \hspace{0.3cm} $\pi^{St}_\theta$: style policy
\State \hspace{0.3cm} $\pi^{Sp}_\theta$: spin policy
\State \hspace{0.3cm} $o^{St}$: $\pi^{Sp}_\theta$ observation, contains $B$
\State \hspace{0.3cm} $o^{Sp}$: $\pi^{Sp}_\theta$ observation
\State \hspace{0.3cm} $\Lambda$: LLCs
\State \hspace{0.3cm} $\Gamma$: LLC skill descriptors
\State \hspace{0.3cm} $\Psi$: set of LLC selection strategies
\State \hspace{0.3cm} $H$: current preferences, one per LLC
\State \textbf{Output:}
\State \hspace{0.3cm} $\lambda$: selected LLC
\State \hspace{0.3cm}
\State $s_t = \pi^{St}_\theta(o^{St}$) \Comment{choose style}
\If{$S$} \Comment{select serving LLC}
\State $s_p = \pi^{Sp}_\theta(o^{Sp}$) \Comment{estimate spin type}
\State $\lambda = \text{get\_serve\_llc}(s_t, s_p, \Lambda)$
\Else \Comment{select rallying LLC}
\State \Comment{get metrics from LLC skill descs.}
\State $\gamma = \text{get\_llc\_metrics}(\Gamma, B, s_t)$
\State $\Lambda^* = []$ \Comment{LLC shortlist}
\State $R = []$ \Comment{prob. of returning ball $B$}
\For{$\psi\; \in \; \Psi$}
\State \Comment{get best LLC per strategy}
\State $\lambda_{\psi}, R_{\psi} = \psi(B, \gamma, s_t, \Lambda)$
\State $\Lambda^* \mathrel{+}= \lambda_{\psi}$ \Comment{add $\lambda_{\psi}$ to shortlist}
\State $R \mathrel{+}= R_{\psi}$ \Comment{add return prob. $R_{\psi}$ to $R$}
\EndFor
\State \Comment{estimate H values for selected LLCs}
\State $\hat{H} = H[\Lambda^*] + R$
\State $p = \text{softmax}(\hat{H})$
\State \Comment{sample LLC proportional to prob. $p$}
\State $\lambda = \text{sample}(\Lambda^*, p)$
\EndIf
\State \hspace{0.3cm}
\State \Return $\lambda$
\State \hspace{0.3cm}
\EndProcedure
\end{algorithmic}
}
\label{algo:hlc_inference}
\end{algorithm}

The HLC is responsible for making strategic decisions --- e.g. where to return the ball, how fast to hit, how much risk to take. An overview of the decision-making process is given in Algorithm \ref{algo:hlc_inference} and Figure \ref{fig:inference}. Currently the HLC is only capable of executing simple strategies and is an initial proof of concept of the entire system. However, the selection strategies sub-component is straightforward to replace with a more expressive implementation, which could even include a fully learned model. Next we give details the HLC components and how they are combined to produce an action.

\subsubsection{\textbf{Event-driven decisions}}
 
The HLC action is triggered by the opponent hitting the ball (i.e. an event external to the agent). One timestep  after the opponent hits the ball, the HLC makes a decision that applies until the opponent hits the ball again. 
The robot only starts moving after the HLC has decided on the strategy for that ball. Waiting one step gives the policy sufficient information to make a decision. We also tried zero and three steps, but three steps did not give the robot enough time to react to faster balls and zero steps did not allow for an accurate estimation of ball velocity. The decision is only made once since switching LLCs mid-swing resulted in policies ending up in states that were outside of their training distribution (i.e. the robot arm was not where they expected it to be because the previous LLC moved it somewhere new), thus performing poorly.

\subsubsection{\textbf{Style policy}}
\label{hlc-style}

The style policy determines if the robot should return the ball with a forehand or backhand style.  A naive heuristic would be to simply divide the table in half and choose a style based on which half the ball will end up on.  However, such an approach neglects many strategic tradeoffs such as forehand shots being easier for the opponent to smash and the ambiguous nature of balls near the center.  Additionally, noise in the real world and our inability to fully capture spin may mean our physics estimates are inaccurate.  By learning a style policy the HLC can understand the strengths of individual LLCs and compensate for systematic inaccuracies, leading to better overall strategic decisions.

The style policy architecture, similar to the LLC but with only 4.5k parameters, has a (8, 128) observation space. We flatten the LLC (8, 16) observation (described in Section \ref{method-llc-training}) and stack the latest 8 observations to form the observation.  This observation space is an artifact of previous iterations of the HLC that required longer histories. For the current implementation that makes a single, early decision per ball a much smaller history would be preferable. The action space is (2,) representing a one-hot categorical choice between forehand and backhand.

To train the style policy, we first selected a general-purpose forehand and backhand LLC and froze their weights. Then, we selected all available ball states (including reflections) and trained the style policy to maximize the expected ball landing rate. Although we trained the style policy with rally ball states, we found that it generalized to serving ball states. Therefore, we used a single policy for both serving and rallying phases of the game.

\subsubsection{\textbf{Spin classifier}}
\label{hlc-spin}

The spin classifier is a binary classifier that determines if the incoming serve was hit by the humans as a topspin or an underspin. This is crucial for selecting the appropriate LLC for services because, unlike the rallying play, we found it very challenging to have a common policy that handled both topspin and underspin serves. To train the model, we built a dataset of paddle and ball states from the serving dataset (see Section \ref{method-zs-s2r}). Specifically, we record a history of the 6 timestamps of ball and paddle states directly before the paddle made contact with the ball. The observation space is (18,) (see Appendix \ref{app:hlc} for details), the policy is a 2-layer MLP and it outputs the probability that the incoming ball is topspin or underspin.

During inference, to increase the precision on underspin (the less common of the two spin types), we required that the classifier predict underspin on at least 4 out of 5 consecutive past queries to be deemed an underspin prediction overall.

\subsubsection{\textbf{LLC skill descriptors}}
\label{hlc-llc-skill-desc}

To excel in interactive sports, it is crucial to understand one's own capabilities. This motivated the development of LLC skill descriptors which provide detailed metrics to the HLC on the estimated performance of each LLC for a given incoming ball. They are the agent's model of its own capabilities and together with a model of the opponent and current game play, are the foundation of all strategic decision making.

To create the descriptors, we evaluated each LLC in simulation on all 28k ball states averaged over ten repetitions, recording the following policy metadata:
\begin{itemize}
\item Initial ball position and velocity.
\item Post-paddle median hit velocity (hit velocity).
\item Ball landing location and standard deviation on the opponent's side. 
\item Ball landing rate (land rate).
\end{itemize}
This metadata was used to construct lookup tables (we used KD-Trees \cite{kd-tree}) with keys representing initial ball position and velocity. Given any ball in play, the table can be queried for information about the likely performance of each LLC were it to be selected by averaging performance of similar balls it has seen in the past.

While we observed high zero-shot transfer rates per LLC, there remained a sim-to-real gap. Hit rates in the real world were high, however ball return rates, whilst good, were lower than the $>80\%$ return rates we typically observed in simulation. One common failure mode was the LLCs hitting the ball just over the edge of the table. This sim-to-real gap meant that building skill descriptors using only simulated data was likely to lead to errors.

To address this, we updated each LLC's skill descriptor using real-world data. Four researchers played with the robot with the HLC set to randomly select an LLC in order to sample them roughly equally. This resulted in 91 - 257 real world ball throws per LLC. For each LLC and for each ball collected, the 25 nearest neighbors in the relevant LLC-specific tree were updated, weighting the simulated metrics and real world metrics for a single ball throw equally on the assumption that the real world data more accurately reflected expected performance.

\subsubsection{\textbf{Strategies and LLC shortlist}}
\label{hlc-strategies}

Every time the HLC acts, five hand-coded heuristics were used to generate a shortlist (one per heuristic) of the most promising LLC candidates, given the output from the style policy and information collected by the HLC about the opponent on their ability to return balls both in total and broken down by forehand, backhand, and center returns. This opponent information is persisted between games with the same opponent. Not all heuristics use all available information.

The set of heuristics we utilized are as follows;
\begin{itemize}
\item Random selection: Randomly select an LLC if its landing rate exceeds 80\%.
\item Prioritization of hit velocity: Select the top $m$ LLCs with the fastest hit velocities, given that their landing rates are among the top $n$.
\item Prioritization of landing distance: Select the top $m$ LLCs with the farthest landing positions from the initial ball state, given that their landing rates are among the top $n$.
\item Exploitation of opponent's weak side (backhand or forehand): Select an LLC that targets the opponent's weaker side.
\item Consideration of opponent's overall skill: For strong opponents, we assume they can hit the ball from any position on the their side. If the opponent has a hit rate above 75\% we select the farthest landing position for the given ball state on the assumption that this forces the opponent to work harder to return the ball. Otherwise, select the LLC with the highest landing rate.
\end{itemize}

From the shortlist we select the LLC that will be used to return the ball with weighted sampling (to make the robot less predictable) described below.

\subsubsection{\textbf{LLC preferences (H-value) \& choosing an LLC}}
\label{hlc-llc-q-vals}

Another key aspect of playing competitive sports is understanding the opponent's capabilities and being able to adapt in response. This motivated learning online preferences for each LLC which, as well as helping to bridge the remaining sim-to-real gap, provide a rudimentary model of the human opponent.

We learned a numerical preference for each LLC, $H(LLC) \in \mathbb{R}$, based on the LLC’s online performance. The agent selects LLCs more often if their preference is higher. However, the preference itself has no connection to reward. Only the relative preference between LLCs matters. We used a simple gradient bandit algorithm \cite{Sutton1998}\footnote{Chapter 2, p37} to learn these preferences, the pseudocode is given in Algorithm \ref{algo:q_update}.

For a given ball, each LLC in the shortlist is associated with an offline return rate. We combined the offline return rate and the online preferences (H-values) to select an LLC. We found that combining learned H-values with information from the skill descriptor tables played an important role in improving performance. These H-values serve two major purposes. (1) Online sim-to-real correction; even though efforts were made through the offline updates to the skill descriptor tables, a sim-to-real gap remained, likely because the sample of real world balls used to update the tables was small and generated by a small number of players. H-values allow the policy to quickly switch away from poor-performing LLCs to more stable ones. (2) To learn player-specific strengths and weaknesses; if the current opponent is able to easily send shots that one LLC struggles to return, the HLC can shift weight to another the opponent can less easily exploit.

Each time an LLC was selected the H-value was updated using the binary ball land signal as the reward function.  For each new opponent, these values were initialized to a set of known baseline preferences, to ensure everyone played against the same initial agent. These preferences were updated and persisted across games for the same opponent.

\subsection{Techniques for enabling zero-shot sim-to-real}
\label{method-zs-s2r}

There are two core challenges in simulating robotic table tennis. First, faithfully modeling the robot, paddle, and ball dynamics. High fidelity is required because advanced table tennis play involves manipulating ball angular velocity (i.e.\ spin) and due to the size and weight of the ball, components such as air friction and paddle material play a much larger role in dynamics than in typical robot tasks. Second, accurately modeling the task distribution, i.e. the distribution over initial states of real-world incoming ball trajectories toward the robotic player.

In contrast to prior work \cite{abeyruwan2022sim2real}, the enhanced simulation components presented in this section facilitated a high degree of zero-shot transfer, obviating the need for real world fine-tuning of low-level policies.

\subsubsection{\textbf{Modeling ball and robot dynamics}}

We enhanced the simulation environment described in \cite{D_Ambrosio_2023} and \cite{abeyruwan2022sim2real} by incorporating the MuJoCo physics engine \cite{conf/iros/TodorovET12}, leveraging its advanced solid state fluid dynamics for ball trajectory simulation, refining model and system identification, and improving the representation of paddle rubber.

Our simulation utilized integrated-velocity actuators, stateful actuators with an activation state coupled with an integrator and a position-feedback mechanism. The activation state corresponds to the position actuator's setpoint, and the control signal represents the velocity of this setpoint. System identification was performed for each actuator-joint pair to ascertain parameters such as position gain (N/rad), actuator damping (N/rad/s), friction loss, joint damping (Nm/(rad/s)), force limits (Nm), and armature inertia (kg m$^2$). Our approach to system identification aligned with the methodology presented in \cite{2304.13653}.

We utilized MuJoCo's ellipsoid-based stateless fluid model to simulate ball trajectories. We measured the Blunt drag coefficient and used the default value for the Slender drag coefficient, while setting the Angular drag coefficient to zero. The Kutta lift coefficient and Magnus lift coefficient were kept at their default values.

\begin{figure}[!t]
    \centering
    \includegraphics[width=0.9\columnwidth]{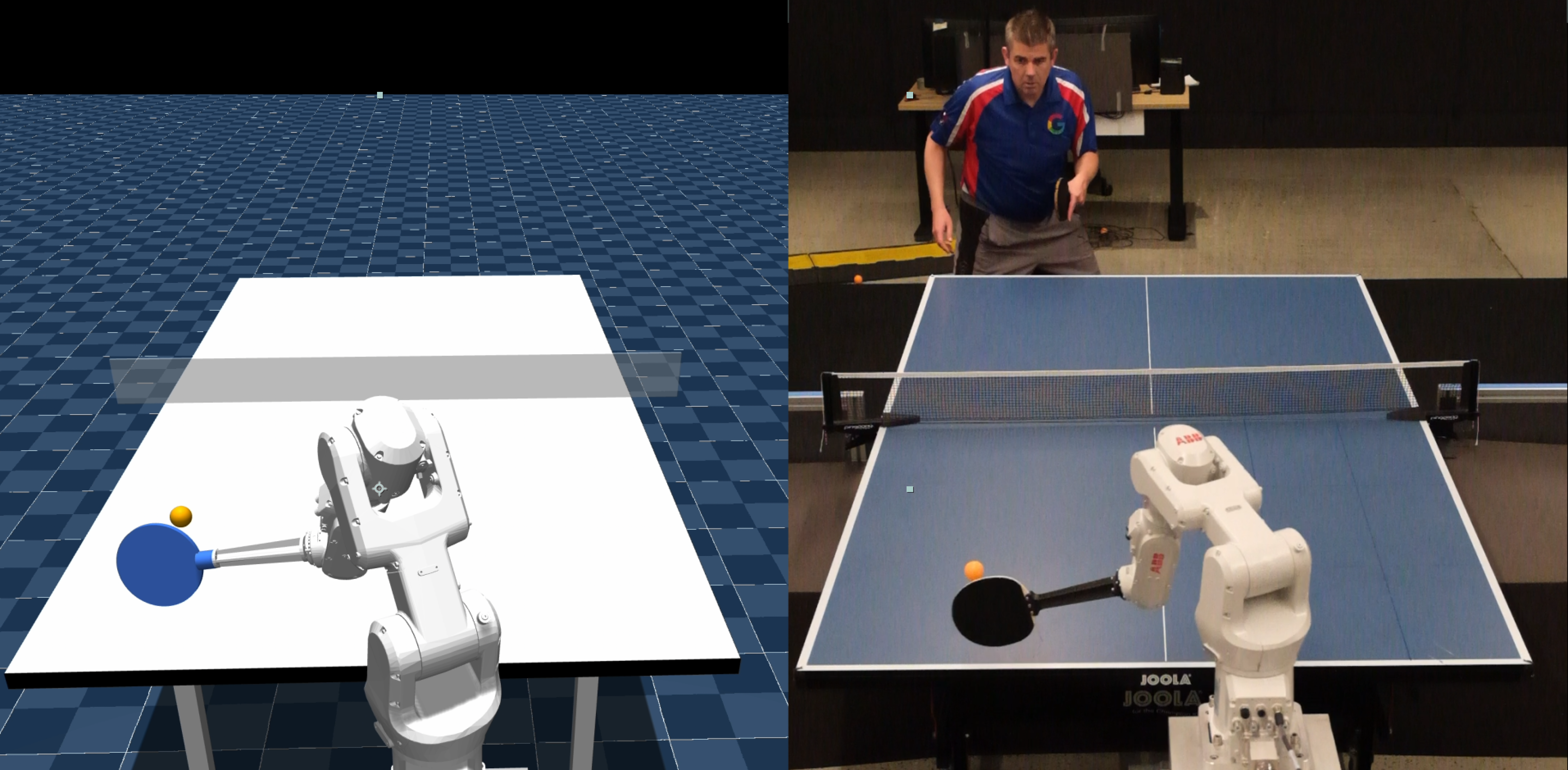}
    \includegraphics[width=0.9\columnwidth]{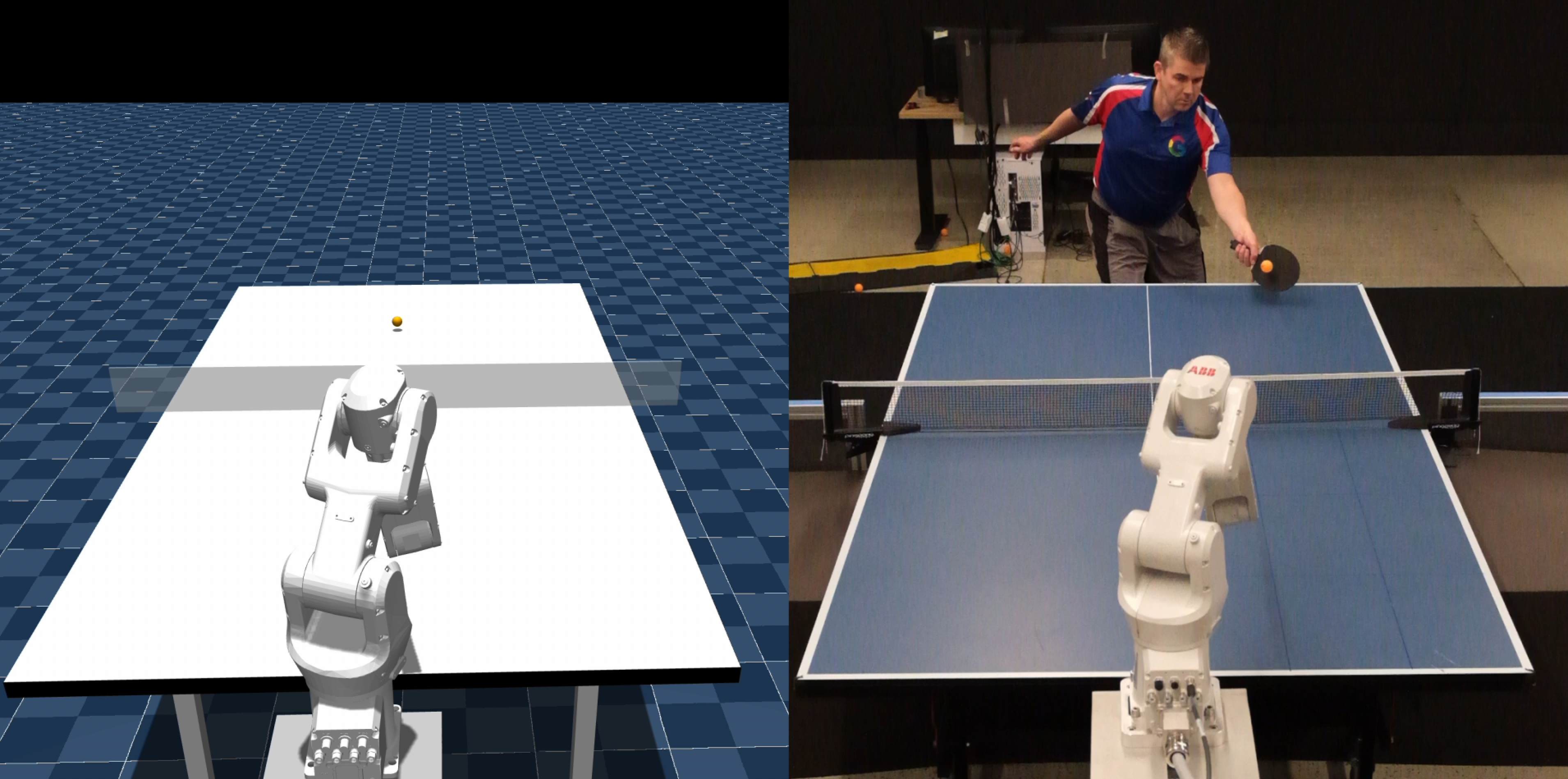}
    \caption{Sample training in simulation and zero-shot transfer to the hardware are shown side by side.}
    \label{fig:simreal}
\end{figure}

Furthermore, the paddle rubber was explicitly modeled using two orthogonal passive joints representing a spring-damper system to approximate a rubber surface. Ball-rubber contact solver parameters (softness, slip, friction) were determined empirically, while joint stiffness, damping, and armature were established through parameter sweeps optimizing for sim-to-real transfer. Analogously, ball-table contact solver parameters were also measured.

We observed a bimodal distribution in contact solver parameters for the paddle rubber restitution when we completed system ID for topspin and underspin ball contacts\footnote{https://mujoco.readthedocs.io/en/latest/modeling.html\#csolver}. Underspin balls exhibit a damping coefficient of $-103$, while topspin balls have a damping coefficient of approximately $-0$. Consequently, during the \textit{topspin correction} phase of policy training (described in the following section), the simulator dynamically selects the appropriate solver parameters based on the ball's pre-contact spin. This bimodality was not observed in the ball-table contact solver parameters.

In addition to modeling and domain randomizing observation noise and latency as described in \cite{D_Ambrosio_2023}, and \cite{abeyruwan2022sim2real}, we randomized table and paddle damping, and friction parameters during training. We employed two shaping rewards, net height reward and a target for the last ABB joint at ball-paddle contact, to mitigate a sim-to-real gap observed due to robot returns overshooting the opponent's side. This approach not only addressed the intended criteria but also promoted competitive robot returns.

\begin{table*}[!t]
\centering
\begin{tabular}{*{13}{c}}
\cmidrule[\heavyrulewidth]{3-13} 
\multicolumn{2}{c}{} & \multicolumn{11}{c}{Number of balls} \\
\toprule
Dataset type & Dataset & All & Forehand & Center & Backhand & Fast & Normal & Slow & Topspin & Nospin & Underspin & Lob \\
\midrule
Rallying & Initial & 2,585 & 1036 & 785 & 764 & 478 & 1,982 & 125 & 972 & 1250 & 363 & 150 \\
Rallying & Cycle 1 & 1,312 & 479 & 448 & 385 & 360 & 891 & 66 & 592 & 519 & 201 & 58 \\
Rallying & Cycle 2 & 1,409 & 587 & 192 & 630 & 18 & 1357 & 35 & 1019 & 379 & 15 & 12 \\
Rallying & Cycle 3 & 593 & 153 & 154 & 286 & 36 & 556 & 1 & 192 & 266 & 135 & 0 \\
Rallying & Cycle 4 & 756 & 417 & 58 & 280 & 2 & 697 & 57 & 303 & 379 & 75 & 5 \\
Rallying & Cycle 5 & 596 & 212 & 127 & 257 & 23 & 562 & 11 & 396 & 192 & 8 & 9 \\
Rallying & Cycle 6 & 5,792 & 1,831 & 2,192 & 1,759 & 149 & 4,933 & 710 & 1,540 & 3,872 & 380 & 551 \\
Rallying & Cycle 7 & 1,198 & 374 & 381 & 442 & 108 & 997 & 93 & 260 & 779 & 159 & 113 \\
Rallying & Final & 14,241 & 5,089 & 4,337 & 4,803 & 1,174 & 11,975 & 1,098 & 5,274 & 7,633 & 1,336 & 898 \\
Rallying & Final+reflection & 28,482 & 9,892 & 8,674 & 9,892 & 2,348 & 23,950 & 2,196 & 10,548 & 15,266 & 2,672 & 1,796 \\
\midrule
Serves & Initial & 858 & 350 & 175 & 318 & - & - & - & 5 & 796 & 57 & - \\
Serves & Cycle 1 & 1,999 & 344 & 1,050 & 570 & - & - & - & 4 & 1,940 & 55 & - \\
Serves & Cycle 2 & 512 & 161 & 253 & 90 & - & - & - & 1 & 453 & 58 & - \\
Serves & Final & 3,369 & 855 & 1,478 & 978 & - & - & - & 10 & 3,189 & 170 & - \\
\bottomrule
\end{tabular}
\caption{The task distribution dataset was developed over multiple training cycles. Notably, cycle 6 substantially improved the coverage of slow and lob balls, whilst the majority of fast balls came from the initial data collection and 1st cycle.}
\label{tab:dataset_summary}
\end{table*}

\begin{figure*}[!t]
    \centering
    \includegraphics[width=0.99\textwidth]{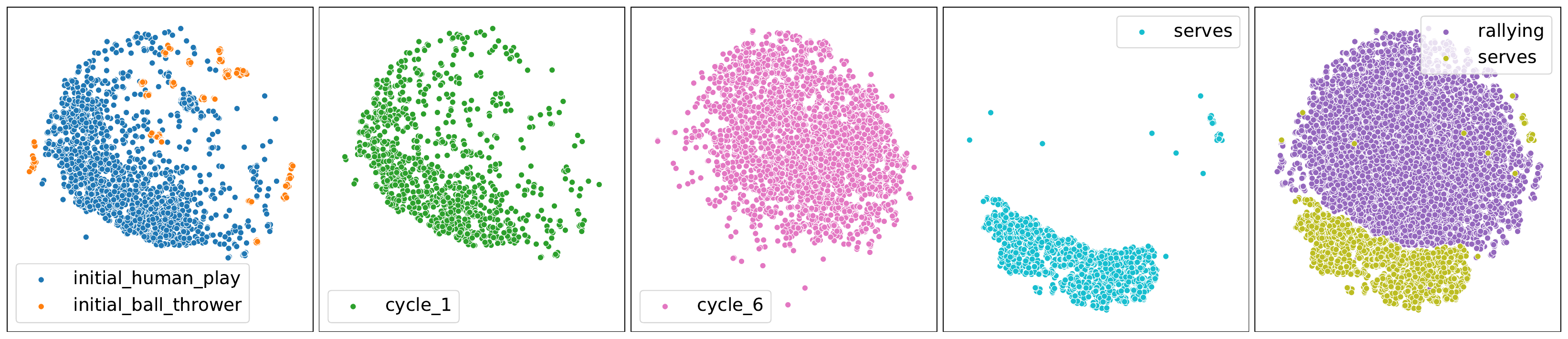}
    \caption{Visualization of the task distribution dataset. TSNE \cite{tsne} was used to project from 9-dimensional balls states to a 2-dimensional representation.}
    \label{fig:dataset_viz}
\end{figure*}

\subsubsection{\textbf{Spin ``correction" and sim-to-sim adapter layers}}
\label{topspin-correction}
As discussed above, the paddle rubber physical parameters in simulation are bimodal, depending on whether there is a topspin or underspin ball. Therefore, directly deploying LLCs revealed a large sim-to-real gap for topspin balls. We developed two solutions to mitigate this issue: topspin correction and sim-to-real adapter layers.

For topspin correction, we fine-tuned an LLC in simulation, switching to topspin-related paddle parameters when the incoming ball had topspin. We also incorporated a net height reward, requiring the returned ball to cross the net at a certain height, and a target joint angle for ball contact. This technique successfully closed the sim-to-real gap in many specialized skills, and also increased the speed of robot returns, adding a competitive edge. However, the sim-to-real gap observed in generalized skilled policies for higher topspin remained despite the topspin correction.

To address the remaining gap, we augmented the topspin-corrected policy with a thin FiLM layer \cite{DBLP:conf/aaai/PerezSVDC18} and trained the adapter using just the topspin balls. We learned a mapping $\text{FiLM}(A|\gamma, \beta = f(o_t)) = \gamma * A + \beta$, where $A \in \mathbb{R}^8$ is the original action. FiLM learns a function $f(o_t)$ that outputs $\gamma $ and $\beta$ $\in \mathbb{R}^8$. The FiLM layer consists of 2.8k parameters and we train the adapter for 5k steps. This closed the sim-to-real gap while preserving underspin return ability. Similar techniques could be applied to heavy underpsin, and side spins, but we leave this for future work.

\subsubsection{\textbf{Iteratively grounding the training task in the real world}}
\label{method-flywheel}

A seed dataset of 40 minutes of human vs. human play was collected along with 480 varied ball throws from a ball thrower. The perception system described in \cite{D_Ambrosio_2023} was used to extract ball positions at 125Hz. The sequence of ball positions was segmented into trajectories consisting of single ball hits where the first ball position of a trajectory is when the ball enters play or immediately after a hit. We then employed an offline optimization process to extract the initial ball state —-- position, velocity, and angular velocity —-- from each trajectory, similar to the approach in \cite{abeyruwan2022sim2real}, such that a simulated ball trajectory starting at that state matches the real ball trajectory as closely as possible. The output of this process is a dataset of initial ball states.

The initial data collection resulted in 2.6k initial ball states. Serves were excluded initially to simplify training. An independent initial serving dataset of 0.9k balls was gathered separately. We extracted ball states from the serving trajectories using optimization methods described in \cite{Triggs:1999:BAM:646271.685629}.

Policies were trained in simulation with the objective of returning all balls in the dataset. During simulated training, we sampled a ball state from the dataset, added small random perturbations, and validated the resulting trajectory. We then initialized the MuJoCo internal state with the ball state and started an episode.

Our non-parametric approach to generating initial ball states --- directly sampling them from the dataset --- is substantially more effective than prior work \cite{abeyruwan2022sim2real}, which used a uniform initial ball state distribution whose bounds were derived from real ball trajectories. The direct sampling approach improves sim-to-real transfer because it aligns the training distribution more closely with balls that are played by human players. \cite{abeyruwan2022sim2real} sampled values for each dimension of the ball state independently, whereas in reality position and velocity components are related. For example, a ball with high linear y velocity is unlikely to have high positive linear z velocity, nor is it likely to have high underspin. Thus independent sampling can create balls that are unrealistic in the real world or are not typically played by amateur humans. Our approach resolves this issue by preserving any empirical inter-relationships between different dimensions of the ball state. Additionally, since no training cycles were expended on unrealistic balls, model capacity was used more effectively, leading to faster training and higher return rates for the same model architecture and training algorithm.

The system was then deployed to the real-world and evaluated against human opponents. Following the same process outlined above, all evaluations were converted into another dataset of initial ball states, and automatically annotated by the system state machine with \textit{return} --- the ball was successfully returned, \textit{hit} -- the paddle made contact with the ball but it did not land on the opponent's side, or \textit{miss} ball --- the paddle did not touch the ball. This dataset was then added to the initial dataset. Balls that were not returned (with annotations \textit{hit} or \textit{miss}) can optionally be overweighted in subsequent training cycles.

This iterative cycle of training models in simulation on the latest dataset, evaluating it in the real world, and using the annotated evaluation data to extend the dataset, can be repeated as many times as needed. We completed 7 cycles for rally balls and 2 cycles for serving balls over the course of 3 months with over 50 different human opponents, leading to a final dataset size of 14.2k initial ball states for rallies and 3.4k for serves. A summary of the dataset evolution is presented in Table \ref{tab:dataset_summary} and Figure \ref{fig:dataset_viz}.

One advantage to this iterative approach to building the training task distribution is that if the policy is repeatedly evaluated against diverse opponents, gaps in capabilities are automatically identified and filled. As the agent's skills improved new weaknesses were revealed whilst simultaneously generating training data to address it. We observed that after 7 cycles performance had not plateaued and we think further cycles could have continued to yield performance improvements.

Two further modifications to the training data distribution were important for boosting performance. (1) Reflecting the data along the y axis. This helped to correct a bias towards forehand play and doubled the final dataset size to 28k ball states. (2) Manually segmenting the dataset into 7 non-mutually exclusive categories --- Fast, Normal speed, Slow, Topspin, No spin, Underspin, Lob. During training, balls were selected each episode by first sampling a category with a probability inversely proportional to the return rate of all balls within that category and then an initial ball state was sampled uniformly from within that category.  This approach allowed us to focus on weak categories while still maintaining performance on ``easier" balls within those categories and across all categories.

\subsubsection{\textbf{Deployment to hardware}}
\label{method-deploy}

Policies were trained in simulation to return individual incoming balls, substantially simplifying training.  To adapt these policies to play a full game of table tennis, we divide each point into sub-episodes \cite{SuttonMDDPWP11} that mimic the training against individual balls: they start when the opponent's paddle contacts the ball and end when the robot returns the ball or a point is scored by either player (i.e. the ball leaves play).  After a sub-episode the robot and  internal data structures of the real environment are reset, ensuring that the policy experienced the single episode semantics it saw in simulation. This step proved crucial for achieving high sim-to-real zero-shot transfer. To estimate paddle state, we employed a customized paddle equipped with motion capture capabilities.

\begin{figure*}[!t]
    \centering
    \includegraphics[width=0.24\textwidth]{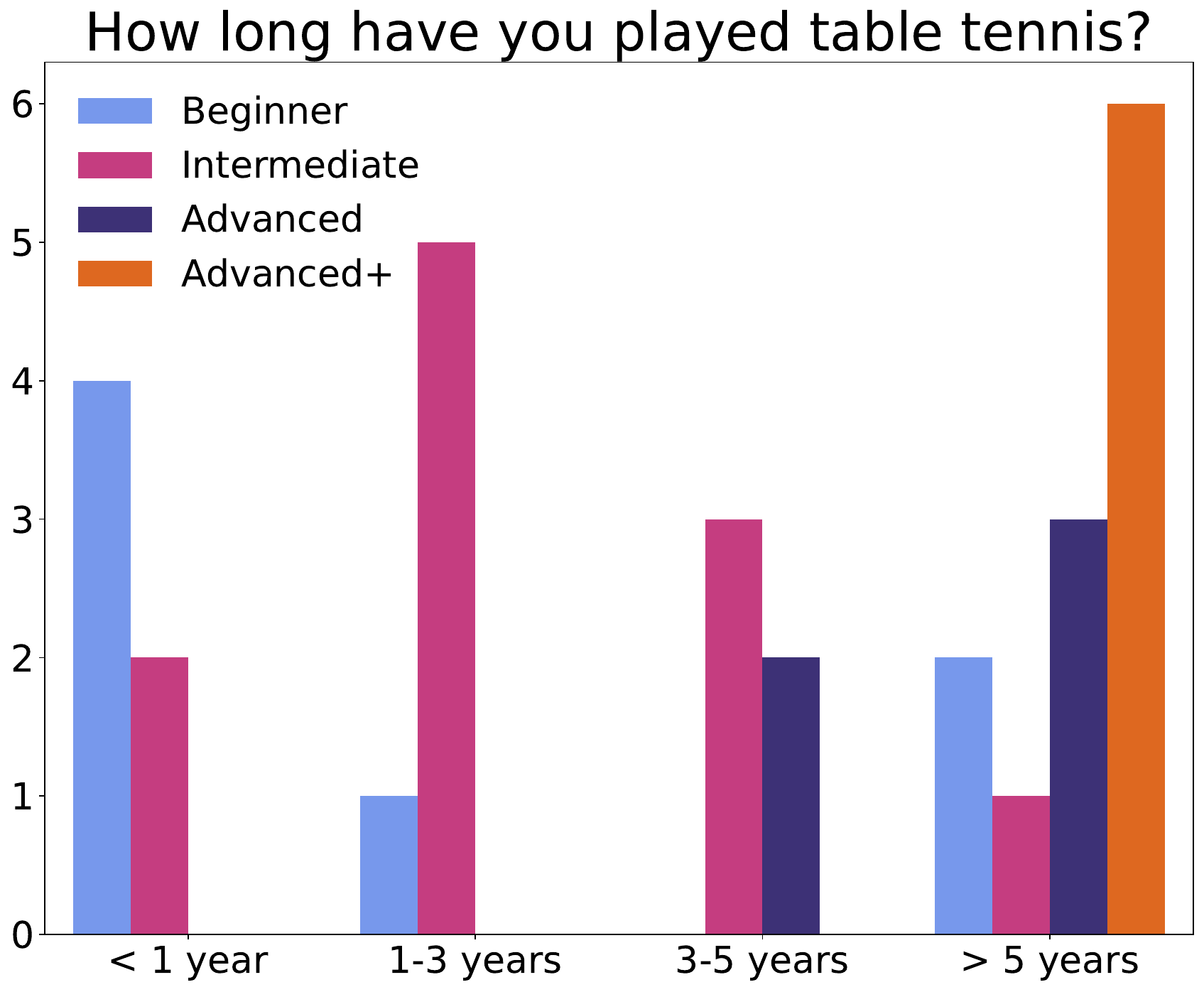}
    \includegraphics[width=0.255\textwidth]{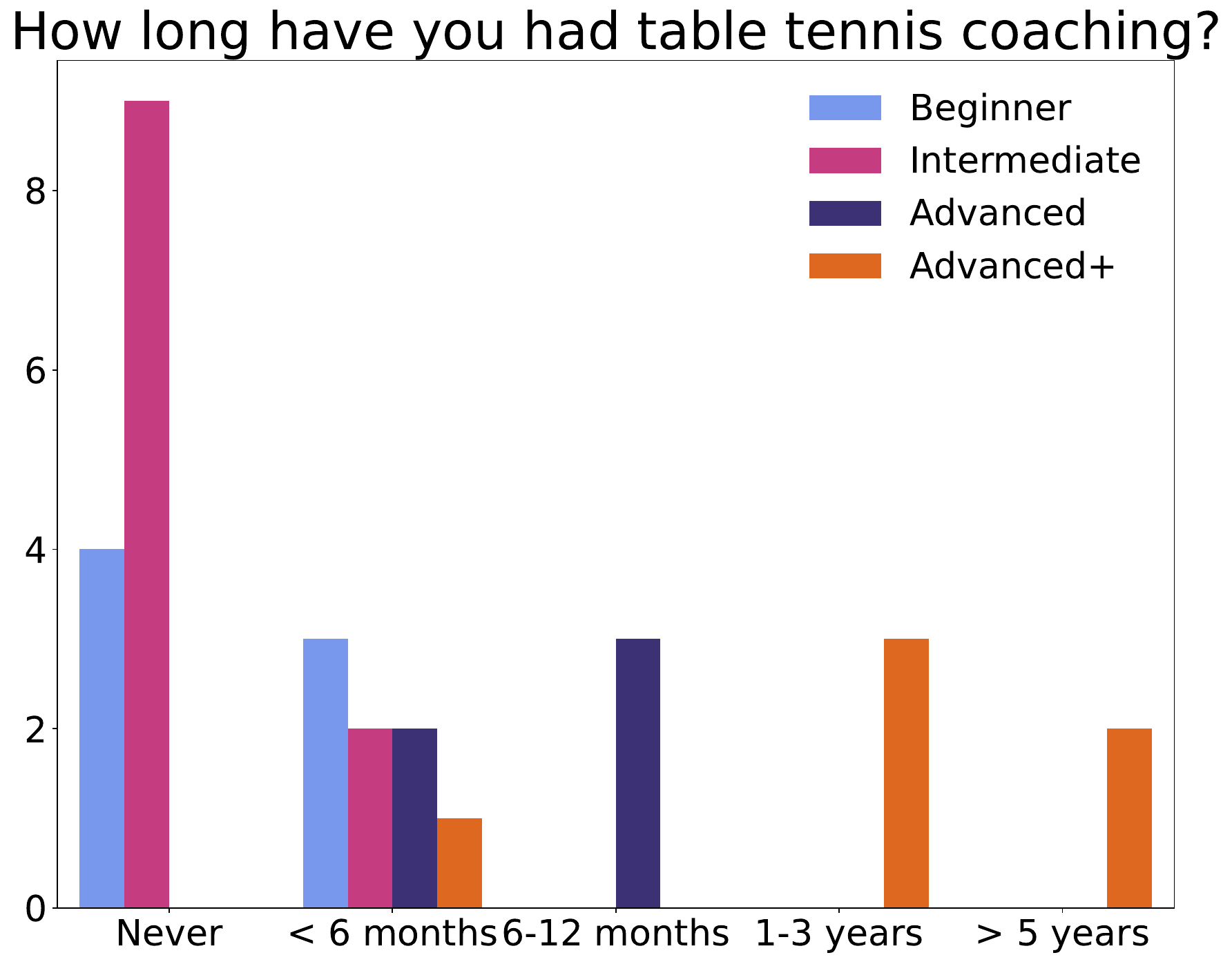}
    \includegraphics[width=0.24\textwidth]{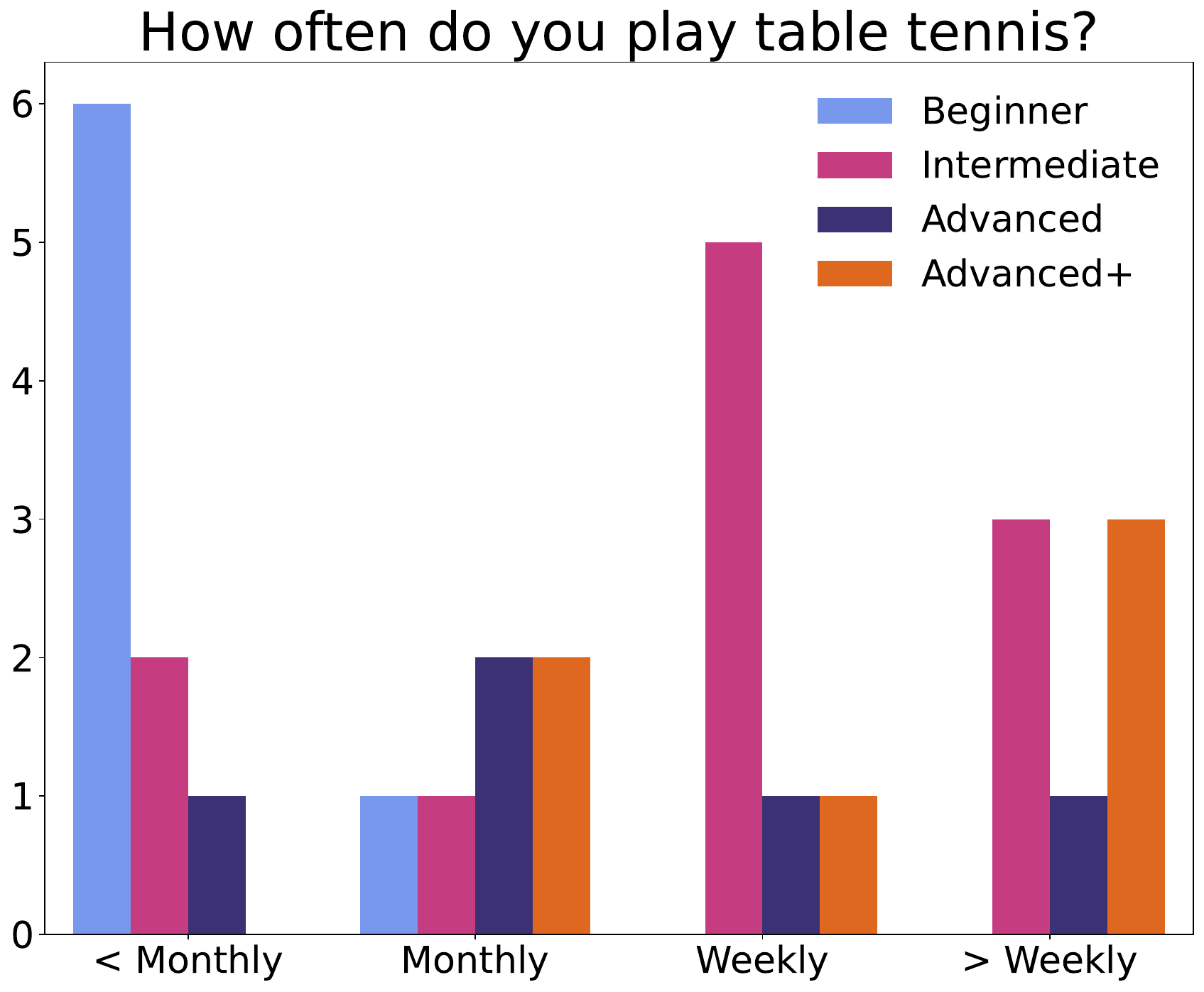}
    \includegraphics[width=0.24\textwidth]{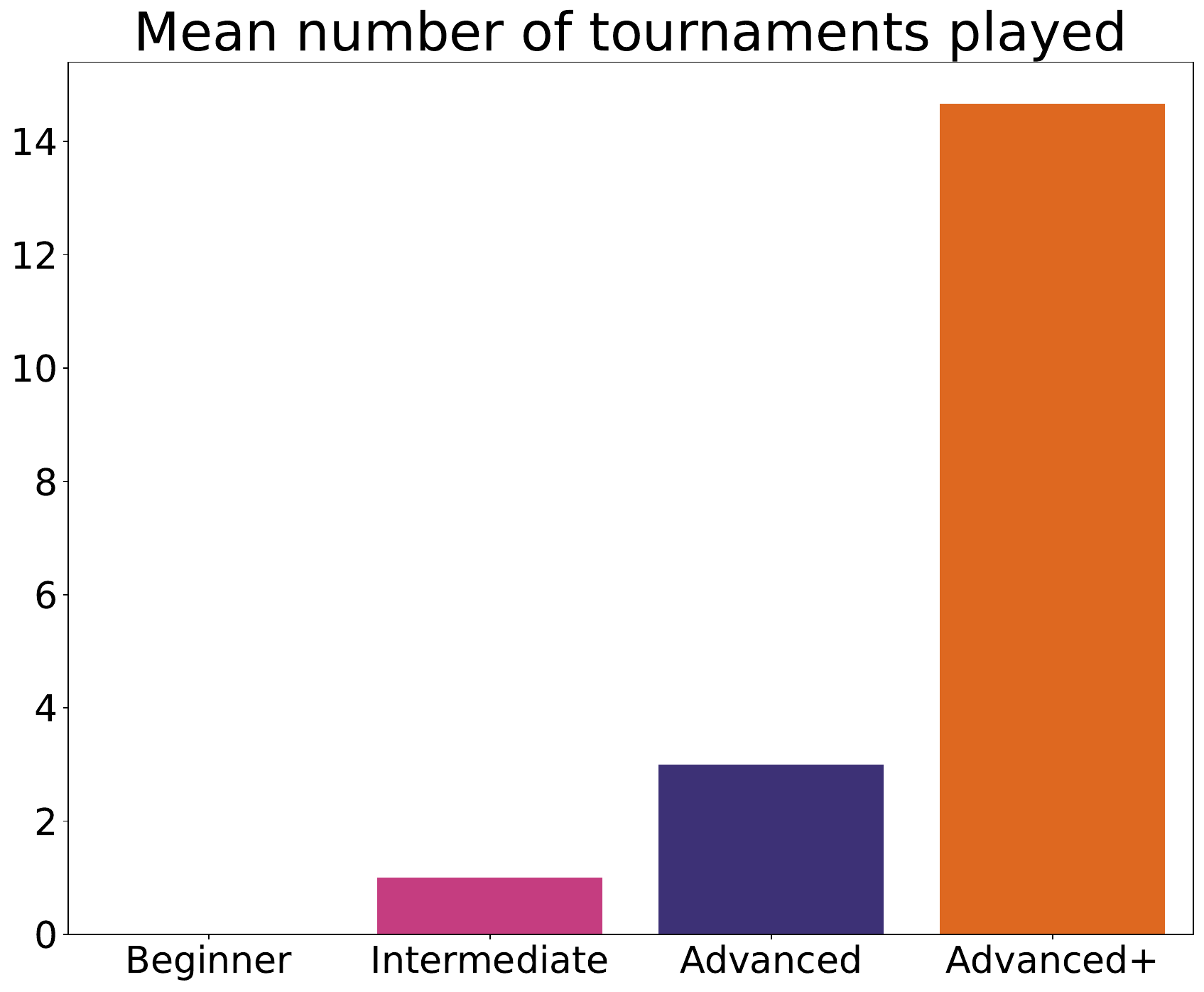}
    \caption{Participants' experience playing table tennis by group.}
    \label{fig:group_stats}
\end{figure*}

\section{Experiments and results}

\subsection{User study design}
To evaluate the skill level of our agent, we ran competitive matches against 29 table tennis players of varying skill levels -- beginner, intermediate, advanced, and advanced+. To ensure that the robot played against a range of human table tennis players, we first ran a pre-study to assess their skill levels.

\subsubsection{Pre-study}

We recruited human players from a local population including a table tennis club, excluding players who were members of our research team or were under 18 years of age. In the pre-study, 59 volunteers played against a professional table tennis coach (who is also an author on this paper, henceforth referred to as "the coach"), who rated each of them as a beginner, intermediate, advanced, and advanced+ player. Independently each person was asked to fill out a questionnaire containing a number of questions about their table tennis experience. The professional table tennis coach did not have access to the responses before rating the players, although the coach may have known some of the players from the local club.

Figure \ref{fig:group_stats} presents the results of the questionnaire and shows clear differences between each of the four groups, validating the coach's rating. Beginner and intermediate players have had little to no coaching and have played in almost no tournaments, however beginner players typically have less than a year's experience and play less than monthly whereas intermediate players tend to have been playing for more than a year and play weekly or multiple times per week. Advanced and advanced+ players have all been playing for more than three years and have had coaching. Compared to advanced players, advanced+ players have been playing for longer and have competed in more tournaments.

Based on their skill levels, we invited a subset of players at different skill levels to return to participate in the full user study.

\subsubsection{Full user study}

\begin{figure}[!b]
    \centering
    \includegraphics[width=0.30\textwidth]{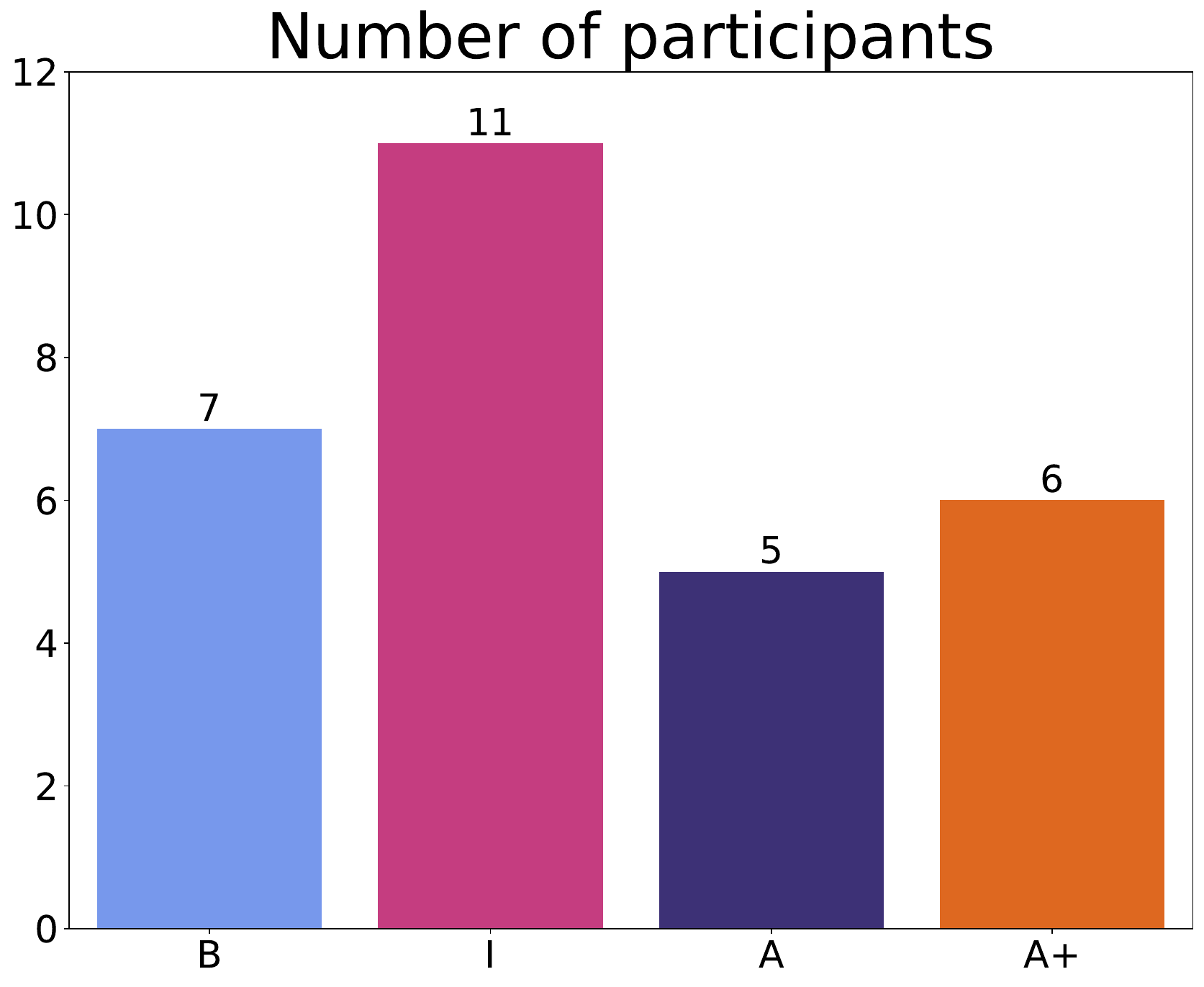}
    \caption{Number of participants by skill group. Human opponent skill level: B = Beginner, I = Intermediate, A = Advanced, A+ = Advanced+}
    \label{fig:num_participants}
\end{figure}

A total of 29 user study participants came to our lab to play table tennis against the robot. Figure \ref{fig:num_participants} gives the skill distribution. 

Each participant was given a safety briefing and a practice session with the special paddle\footnote{The paddle has motion capture LEDs for sensing paddle state and short pips rubber} used for playing against this robot\footnote{Due to human error, 1 beginner player was not given the option of a practice session.}. During the practice session, the participant played against the coach so they could get a feel for the paddle.

After the practice session with the coach, the participant walked to the table with the robot. The rules of the game were explained to the participant (details below). Members of the user study team stood by to run the robot and ensure the safety of the study participant.

The study participant then played three games against the robot and the coach served as the referee for these games. After completion of the three competitive games, we offered them an optional free play session with the robot (up to 5 minutes). Finally, we conducted a brief, semi-structured interview with the participant about their overall experience of playing table tennis against the robot. 

\subsubsection{User study metrics}
The primary measure of agent performance was how well the robot scored in the matches against the human players. We also video recorded the games played against each of the human players and made them available on our \textcolor{blue}{\href{https://sites.google.com/view/competitive-robot-table-tennis}{website}}.

Beyond agent performance, we also wanted to understand what it was like for the human players to play against the robot. We administered questionnaires after each game, asking participants about what it was like to play table tennis against the robot. We also asked whether they had used any particular strategies to play against the robot in the game. At the end of the third game, the final questionnaire also asked them about their level of interest in playing with the robot again.

To get a behavioral assessment of how interested they were in playing again with the robot, we recorded how long they chose to play against the robot during the free play time.

\begin{figure*}[!t]
    \centering
    \includegraphics[width=0.19\textwidth]{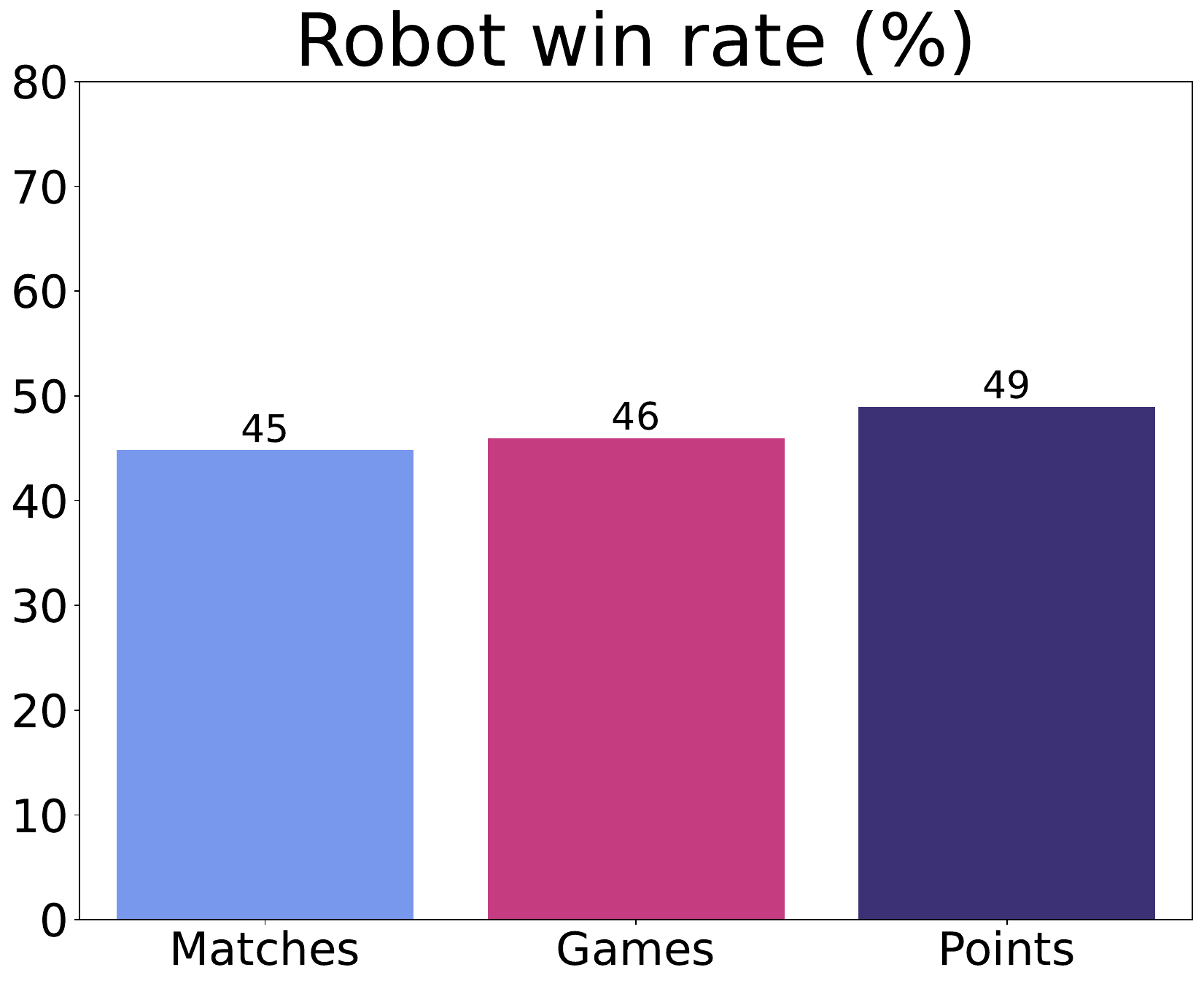}
    \includegraphics[width=0.19\textwidth]{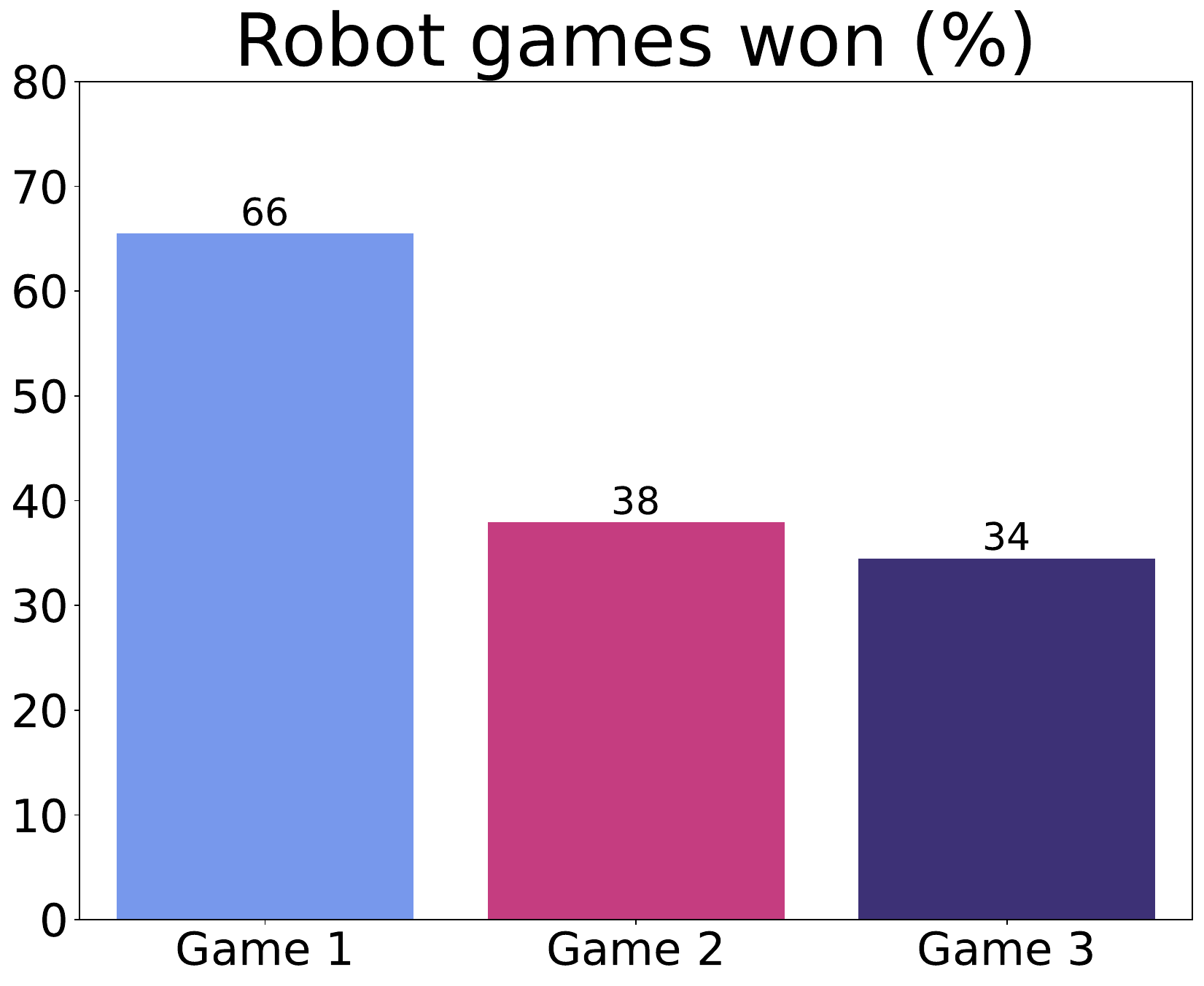}
    \includegraphics[width=0.19\textwidth]{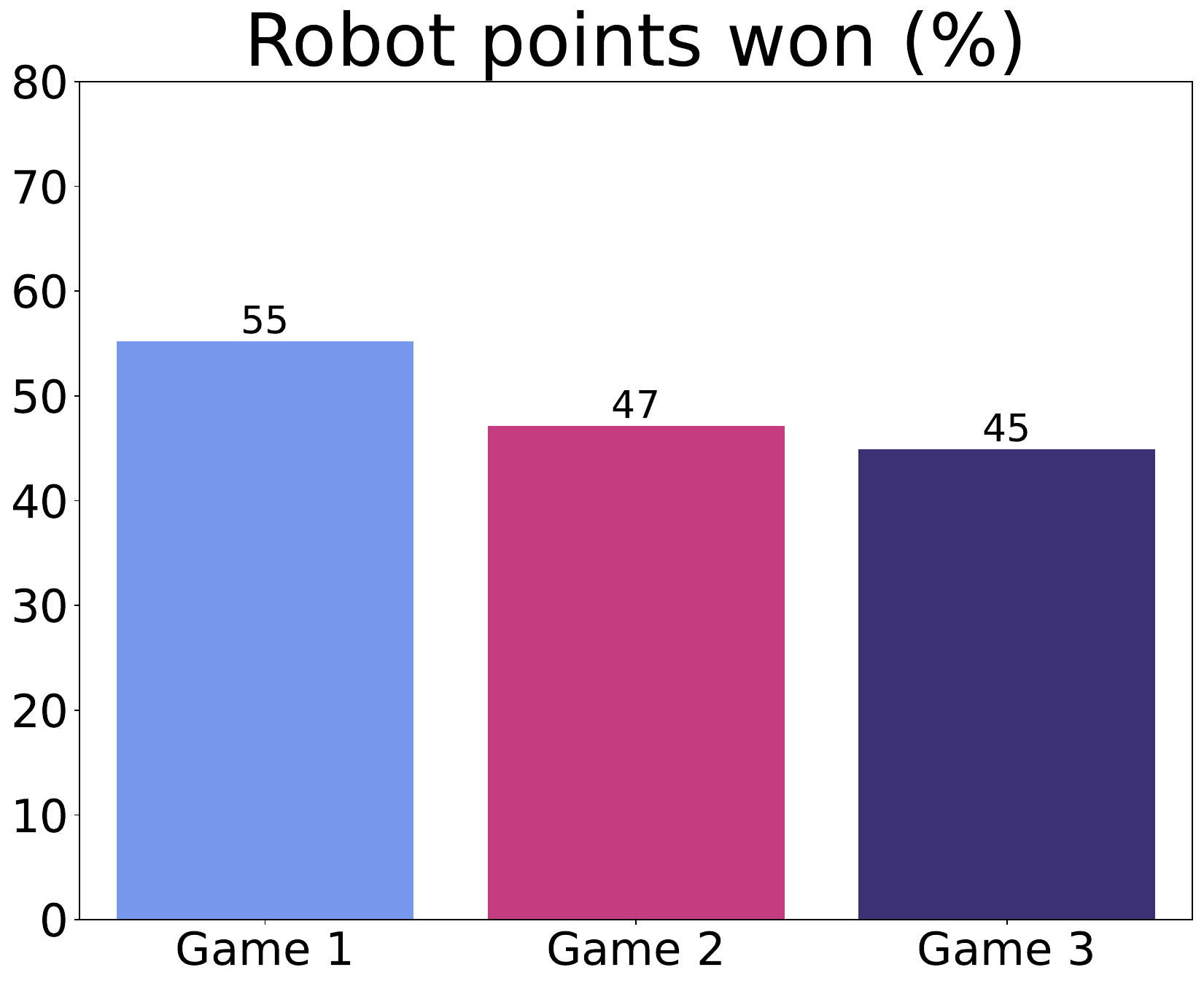}
    \includegraphics[width=0.19\textwidth]{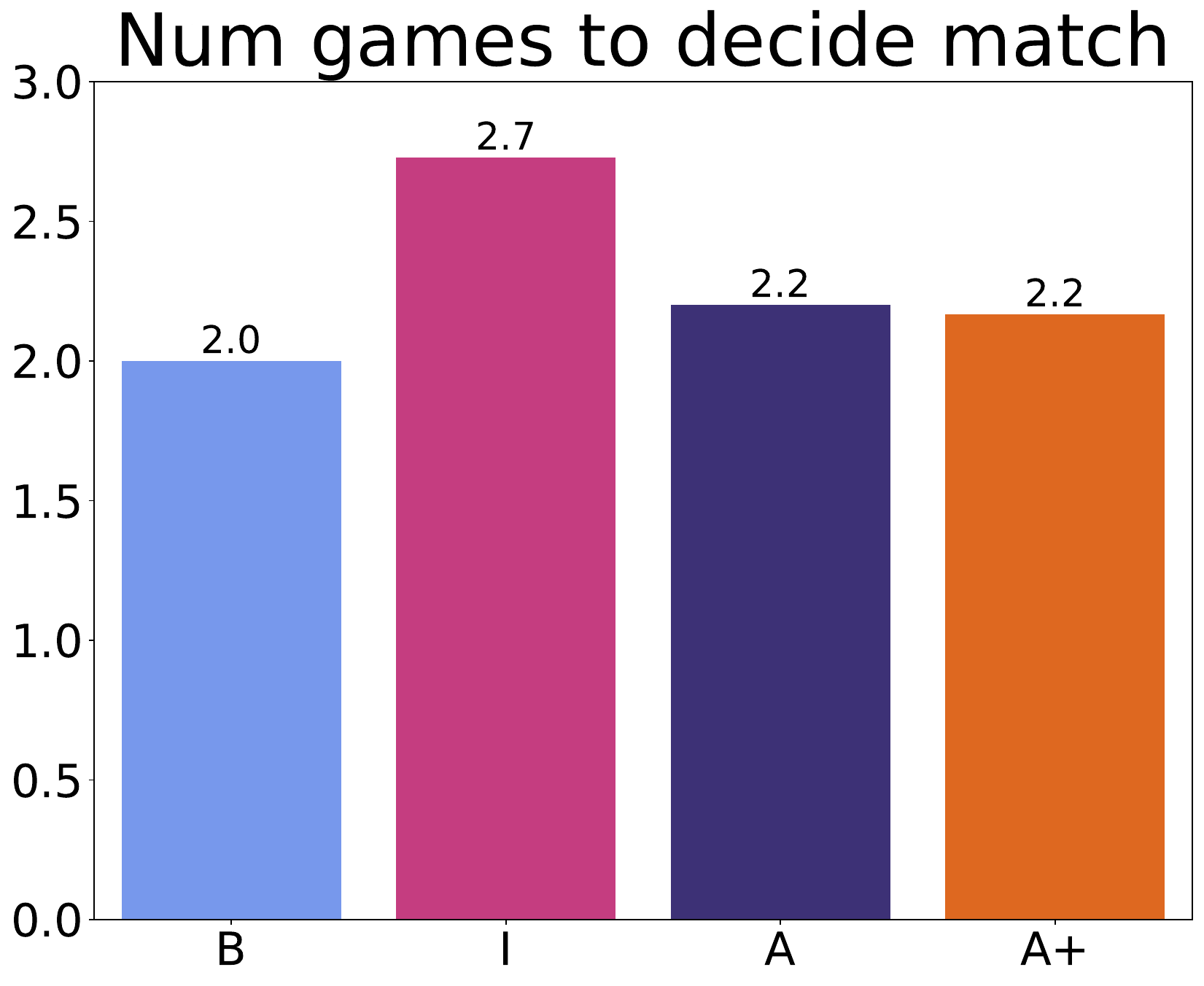}
    \includegraphics[width=0.19\textwidth]{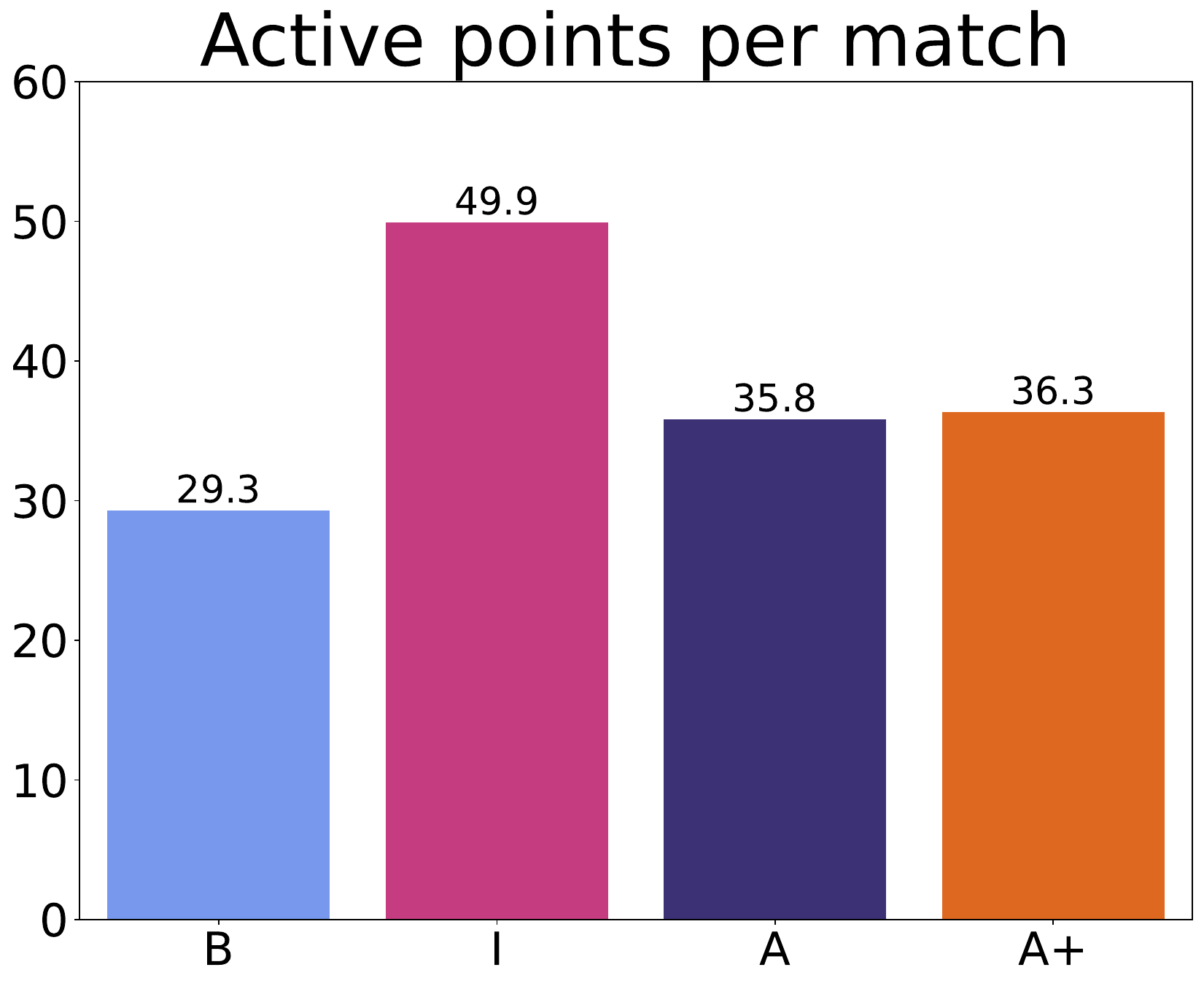}
    \includegraphics[width=0.265\textwidth]{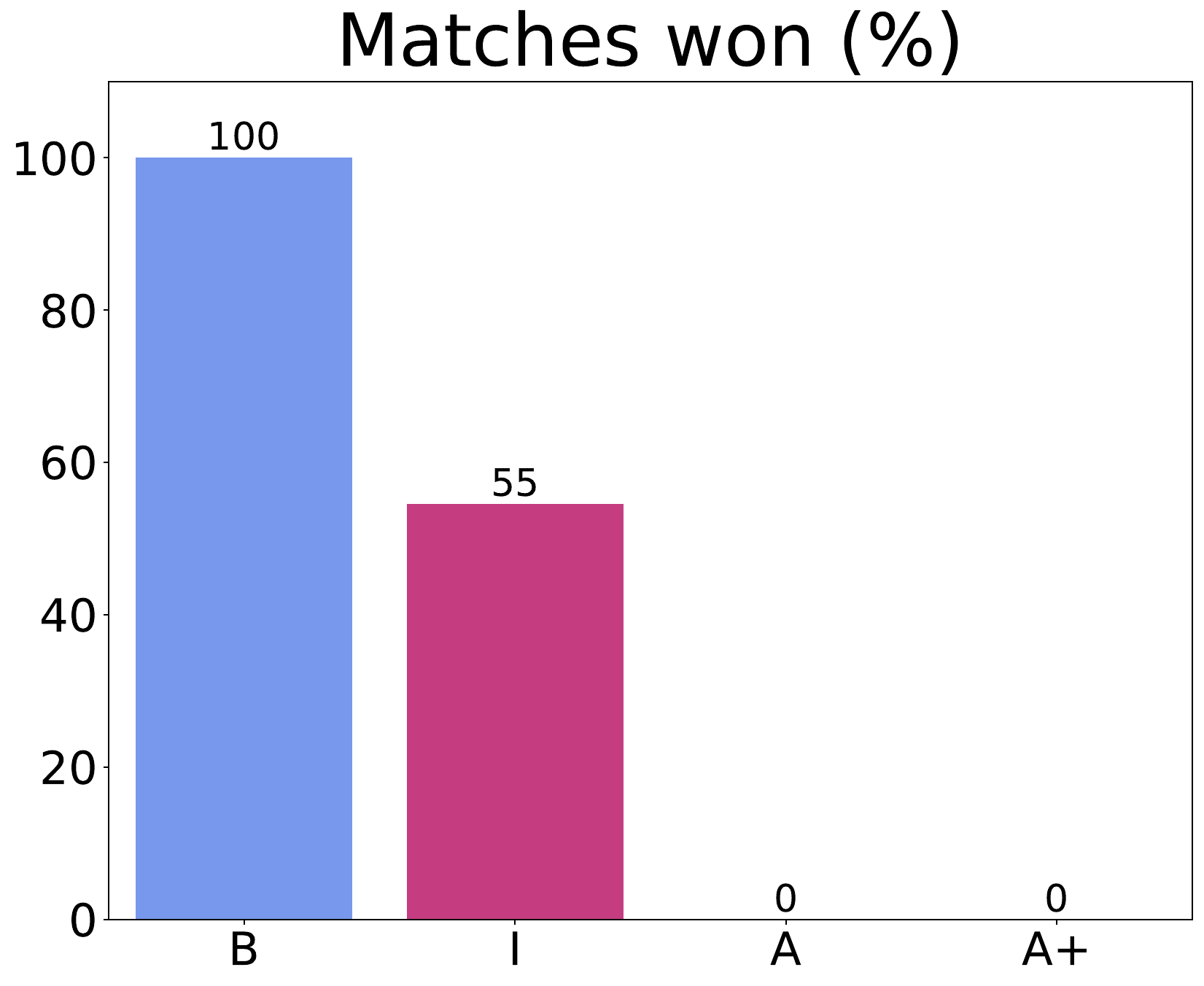}
    \includegraphics[width=0.36\textwidth]{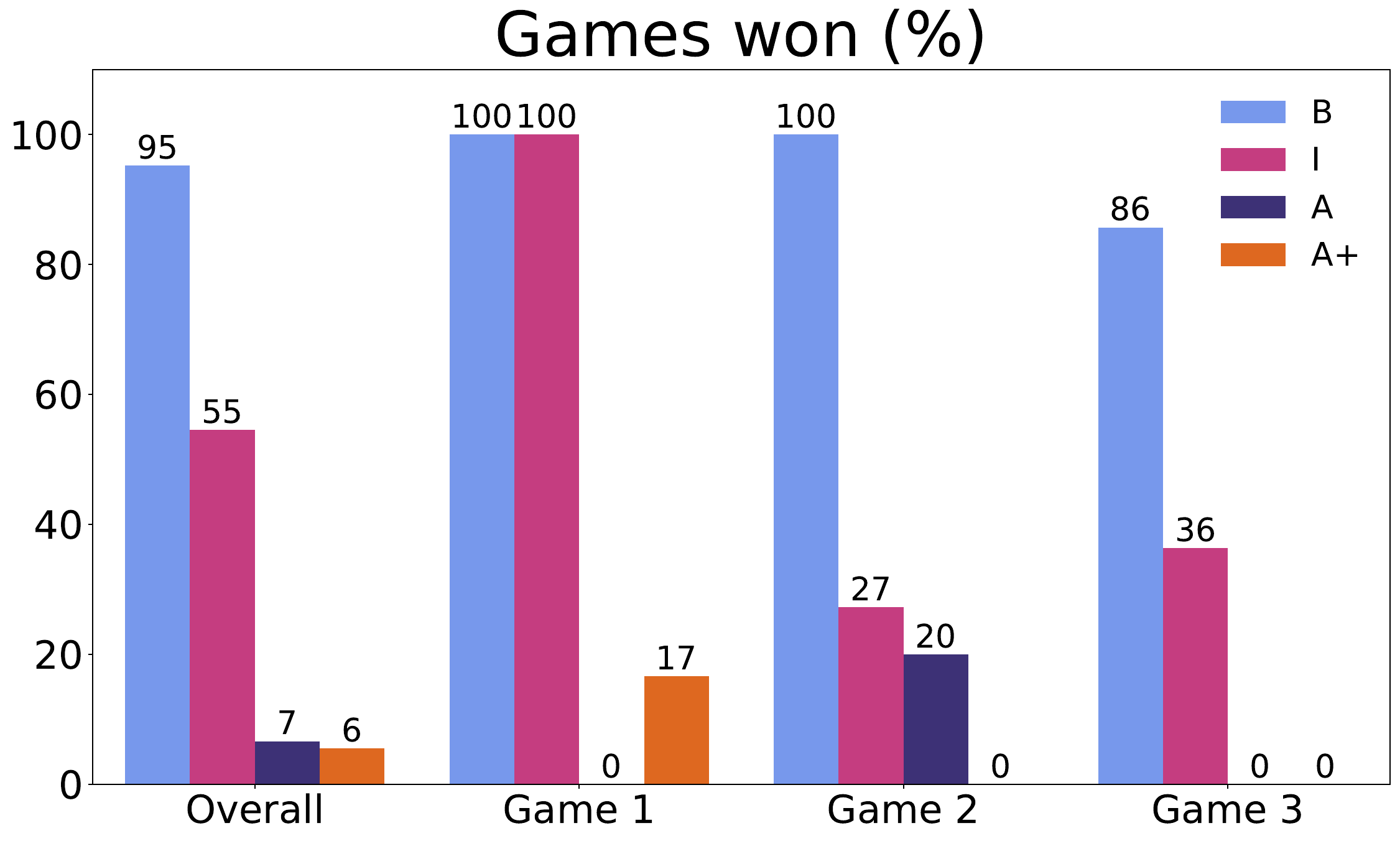}
    \includegraphics[width=0.36\textwidth]{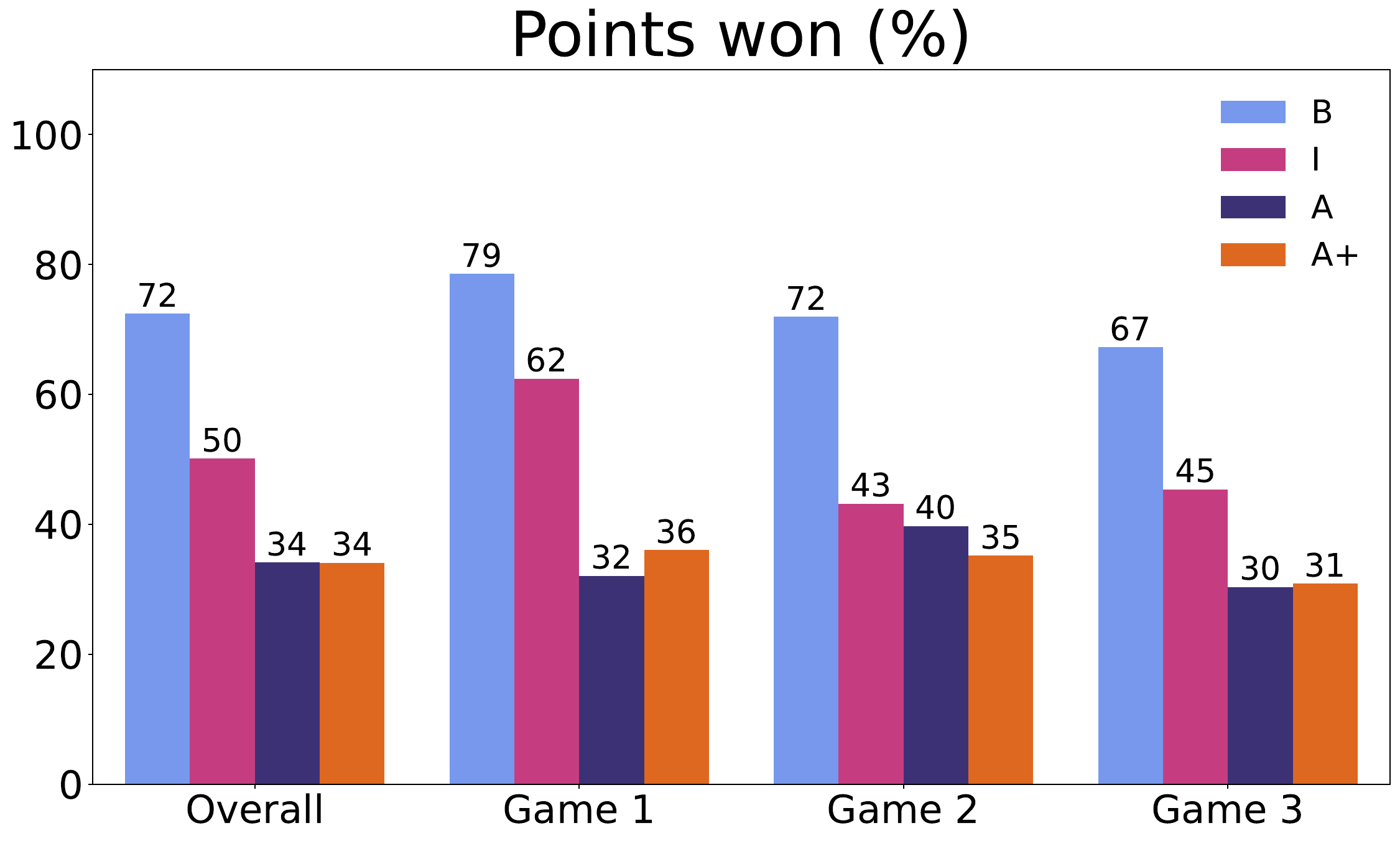}
    \caption{Match statistics. The robot won 45\% of all matches. 100\% against beginners, 55\% against intermediate players and no matches against advanced players. This suggests the robot performs at an amateur intermediate level. Human opponent skill level: B = Beginner, I = Intermediate, A = Advanced, A+ = Advanced+ }
    \label{fig:match_summary}
\end{figure*}
    
\subsection{Match rules}

The table tennis matches between the human players and the robot largely followed the rules set forth by the International Table Tennis Federation (ITTF) \cite{2024_ITTF_Statutes_clean_version}.  In summary, the robot and human play three games, in which the first player to reach eleven points by a margin of two points (or a score of twenty points) wins the game. The player that wins the majority of games wins the match. Unlike a real ``best-of-three" match, all three games were played to ensure consistent data among participants. The coach acted as a referee to determine scoring and rules violations. Human players were given a minimum two-minute break between matches to rest and fill out a short survey.

In table tennis, one player serves the ball to another by tossing the ball up and striking it so that it bounces on their side of the table and then the opponent's.  While the specific requirements of the serve limit the types of shots possible, the serving player nevertheless has a strong advantage \cite{katsikadelis2013interaction} as they can control the initial conditions of the ball e.g.\ to exploit weaknesses in their opponent or to set themselves up for a strong shot. For this reason, standard table tennis rules rotate the serving player every two points. However, the robot in this study (as designed) cannot serve and therefore the human player always serves the ball to the robot. To compensate for this limitation the human cannot score points or lose points on the serve; the robot must return the ball and then points may be scored. This rule did result in some more skilled players repeatedly attempting risky serves near the edges of the robot's capabilities to maximize their score, however we felt this was a necessary compromise to accommodate players of lower skill who may not be used to official serving rules. Overall we felt this method of serving struck the best balance between the human's serving advantage and robot's capabilities. Additionally, these rules ensure that matches progress beyond services and into the rallying component of the game, which involves a much broader set of skills than the service component. A smaller five player study was conducted with alternate serving rules and is discussed in Section \ref{results:serves}.

Two other limitations of the robot were accounted for. If the robot entered a protective stop state, then the point was considered a ``let" (no one scores). Similarly if the ball was hit very high (roughly 2 meters above the table) the point would also be a let due to the limited field of view of the cameras. Applications of all rules were up to the referee's discretion. Other limitations of the robot such as it's inability to reach all the way to the net were deemed too complicated to accommodate for and thus no additional rules were applied.

\subsection{Match results}

Figure \ref{fig:match_summary} breaks down various statistics of the matches between the humans and the robot. Overall the robot was solidly in the middle of the participants, winning 45\% of matches, 46\% of games, and 49\% of points. When we break down matches by skill level, a clear pattern emerges. The robot won 100\% of matches against beginner opponents, 55\% of matches against intermediate opponents, and no matches against more skilled opponents. The implication is that the robot's skill level is intermediate; it can easily beat the previous skill level, is unable to win against higher skill levels and has roughly even odds to win against this skill level. That is not to say that the robot completely dominates or is dominated by other skills levels. Looking at the breakdown of points scored, the robot won 72\% of points against beginners, 50\% of points against intermediate players, and 34\% of points against advanced and advanced+ players. Thus, the robot can still provide an interesting game to all levels of skills.

\begin{figure}[!b]
    \centering
    \includegraphics[width=0.48\textwidth]{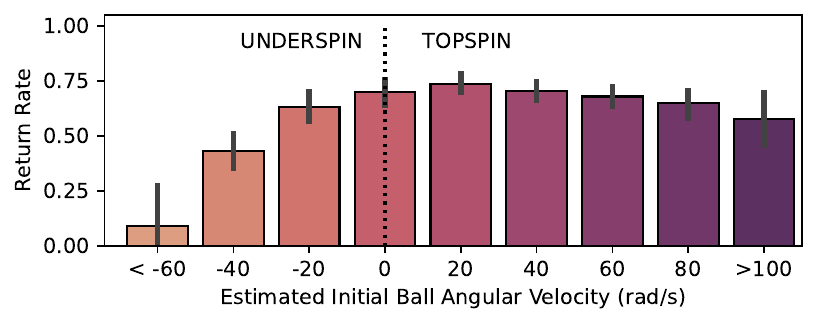}
    \caption{Robot's return rate vs different estimated spin levels during rally shows that the robot is better at returning incoming balls with topspin compared to balls with underspin. Data is aggregated across all matches.}
    \label{fig:spin}
\end{figure}

When playing against beginner and intermediate skill levels, there was an intriguing trend --- the robot always won the first game (see Fig. \ref{fig:match_summary}, Games Won (\%)). We hypothesize that during the first game, the human is getting used to the novel situation they find themselves in: playing a sport against a robot, using an unfamiliar paddle, pressure from the competitive setting, and so on. In post-game surveys players commented that they were still getting used to the robot, with several players also suggesting that the robot was intimidating and loud, which could be a lesson for balancing high-speed performance and human comfort in HRI scenarios. Focusing on the intermediate players, by the second game, they appeared to have become more comfortable with the situation and the more skilled players had identified weak points in the robot's policy they can exploit. Additionally, because the player may not have been playing optimally in the first game, the robot's adaptation may not have keyed in on the correct features leading to lower game 2 performance by the robot. However, by the third game the robot was able to learn from the opponent's play style and improved its win rate (also see Fig. \ref{fig:qvalue_changes}).

\begin{figure*}[!t]
    \centering
    \includegraphics[width=0.24\textwidth]{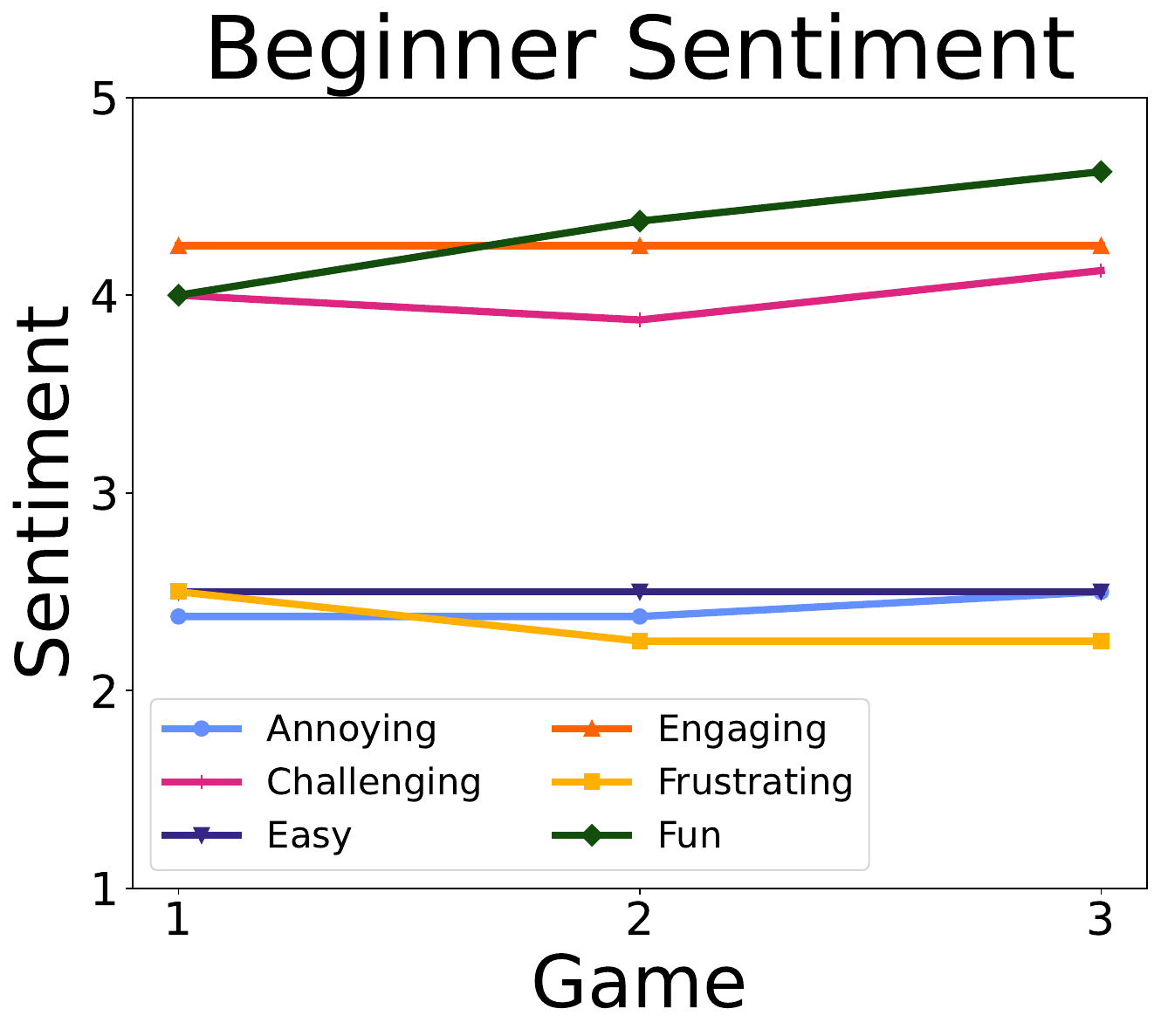}
    \includegraphics[width=0.24\textwidth]{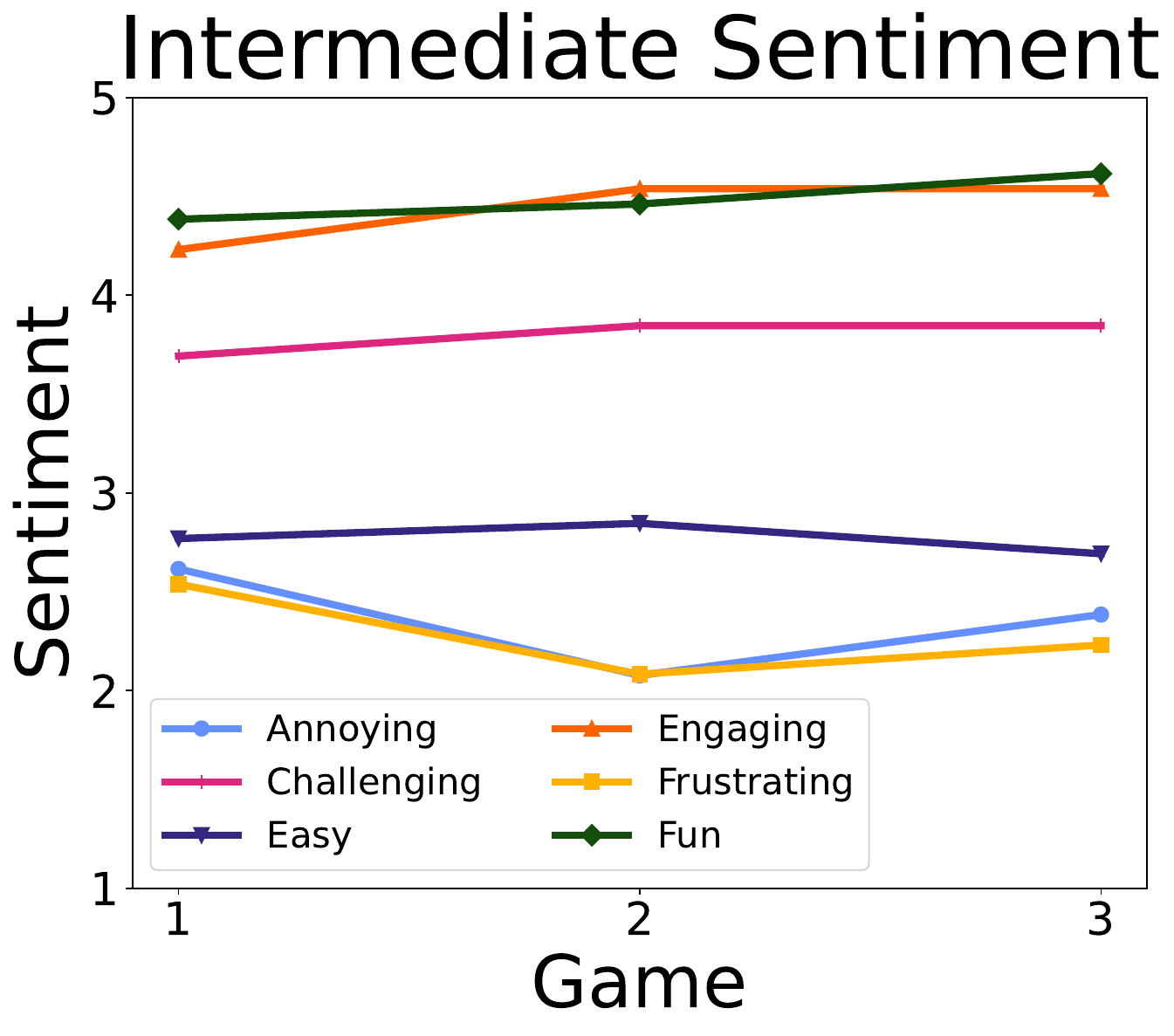}
    \includegraphics[width=0.24\textwidth]{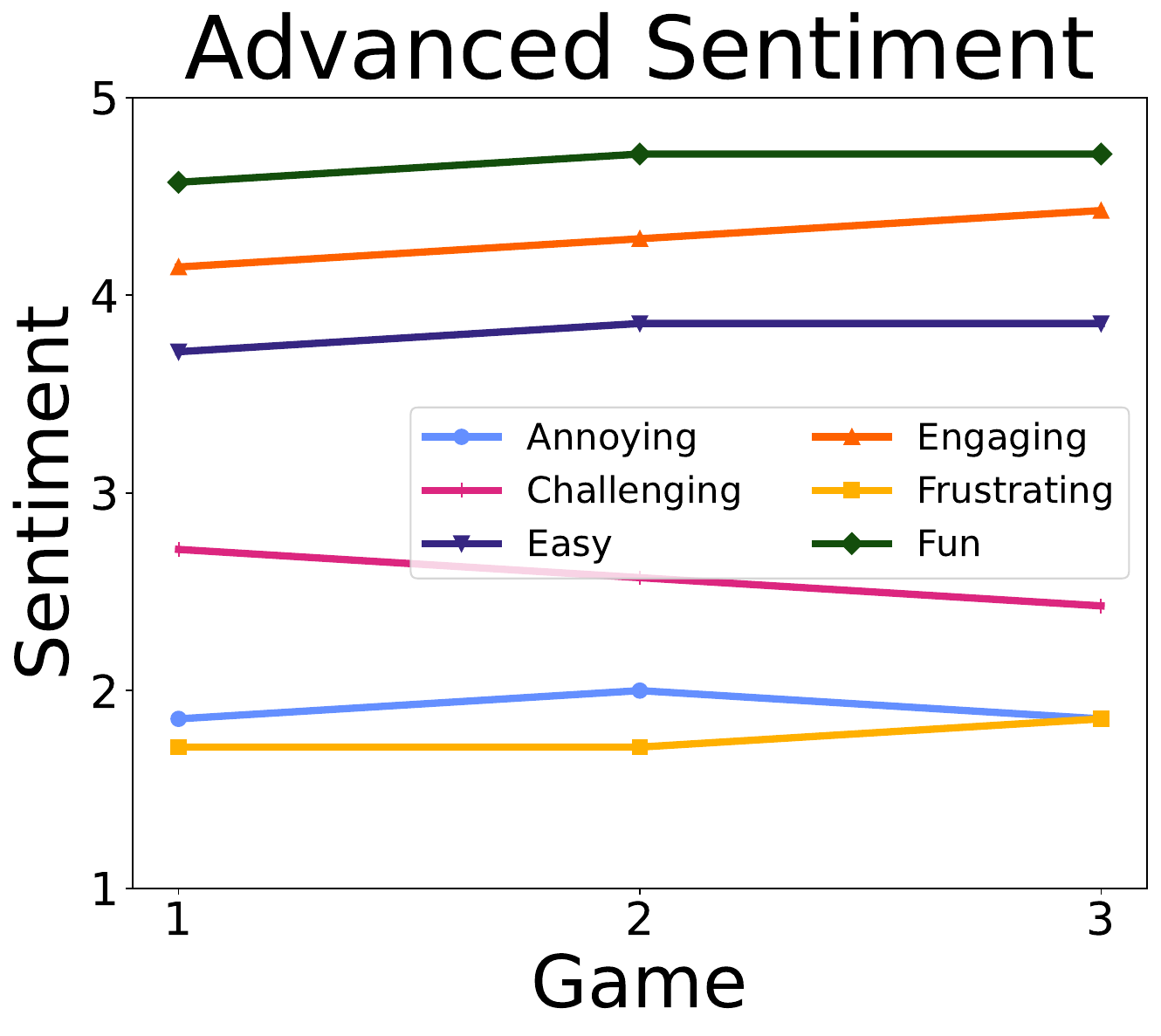}
    \includegraphics[width=0.24\textwidth]{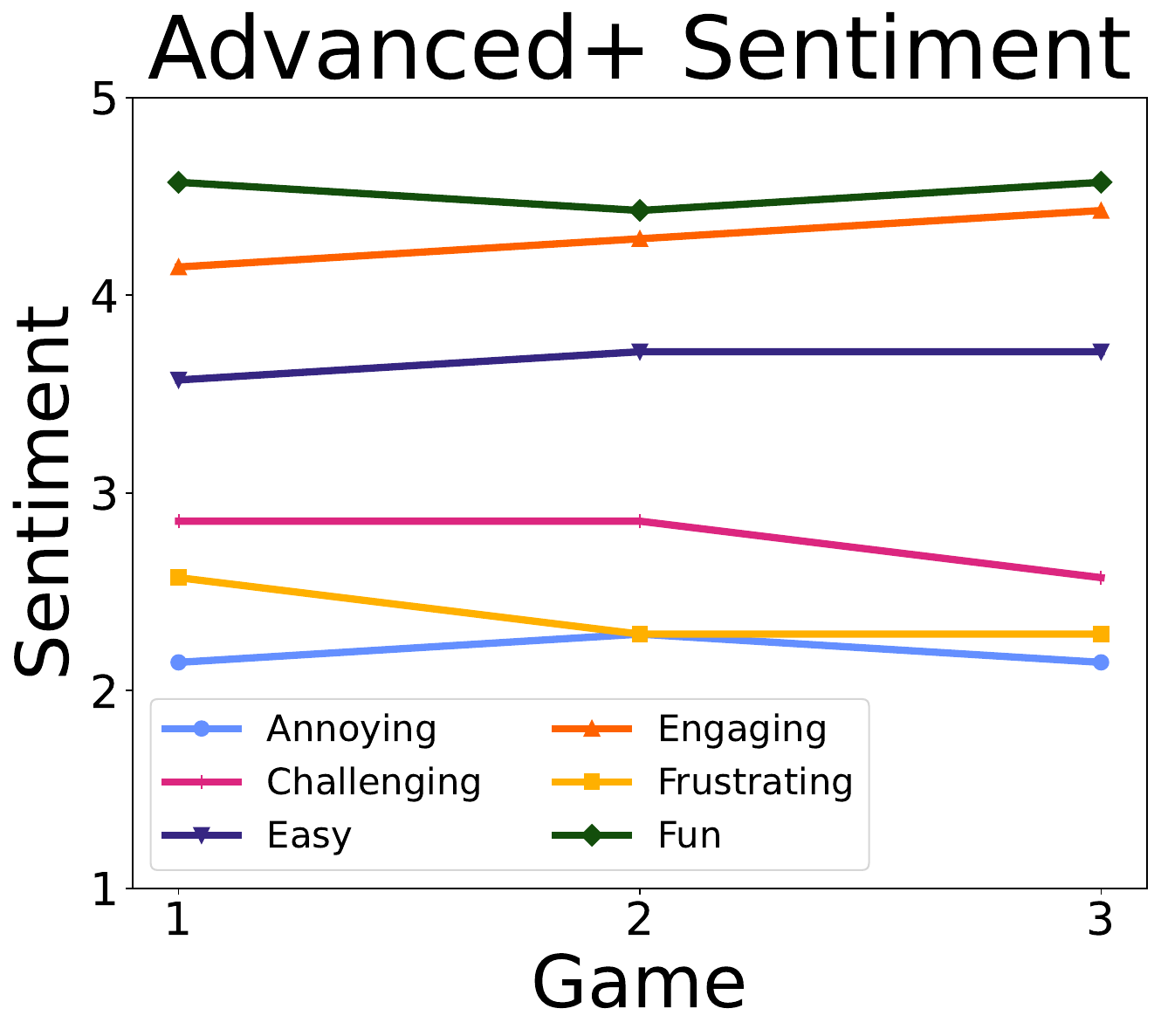}
    \includegraphics[width=0.24\textwidth]{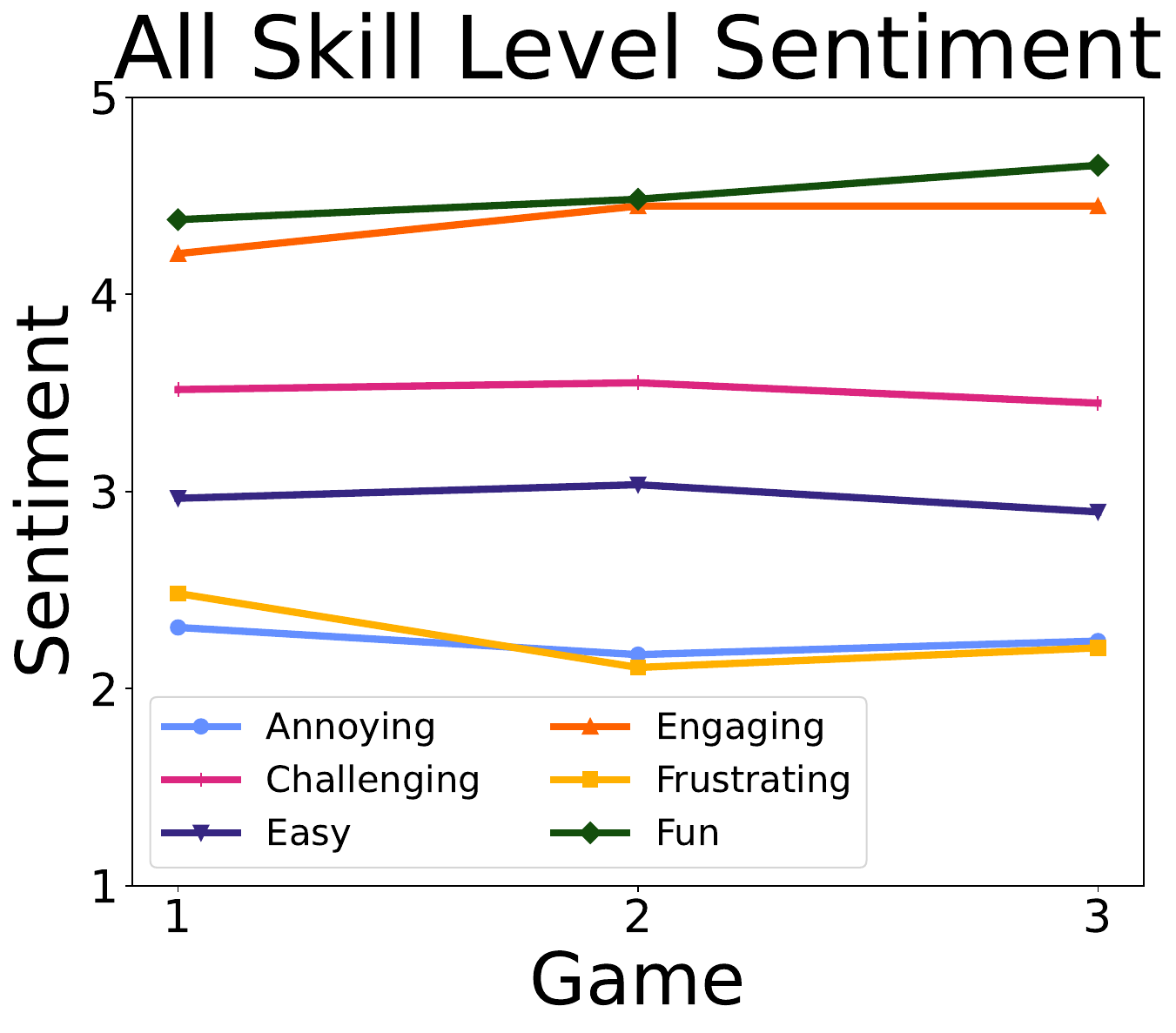}
    \includegraphics[width=0.24\textwidth]{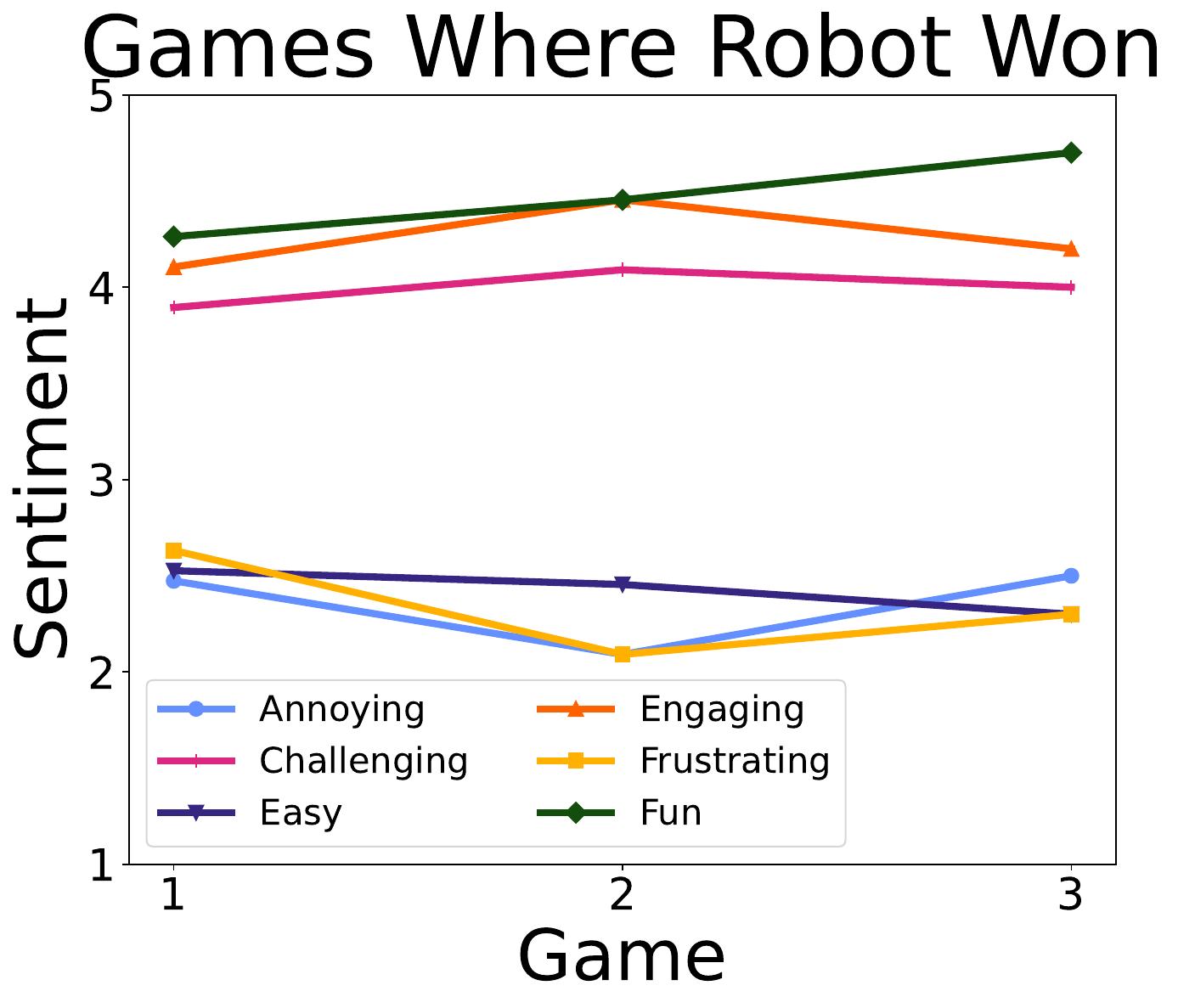}
    \includegraphics[width=0.24\textwidth]{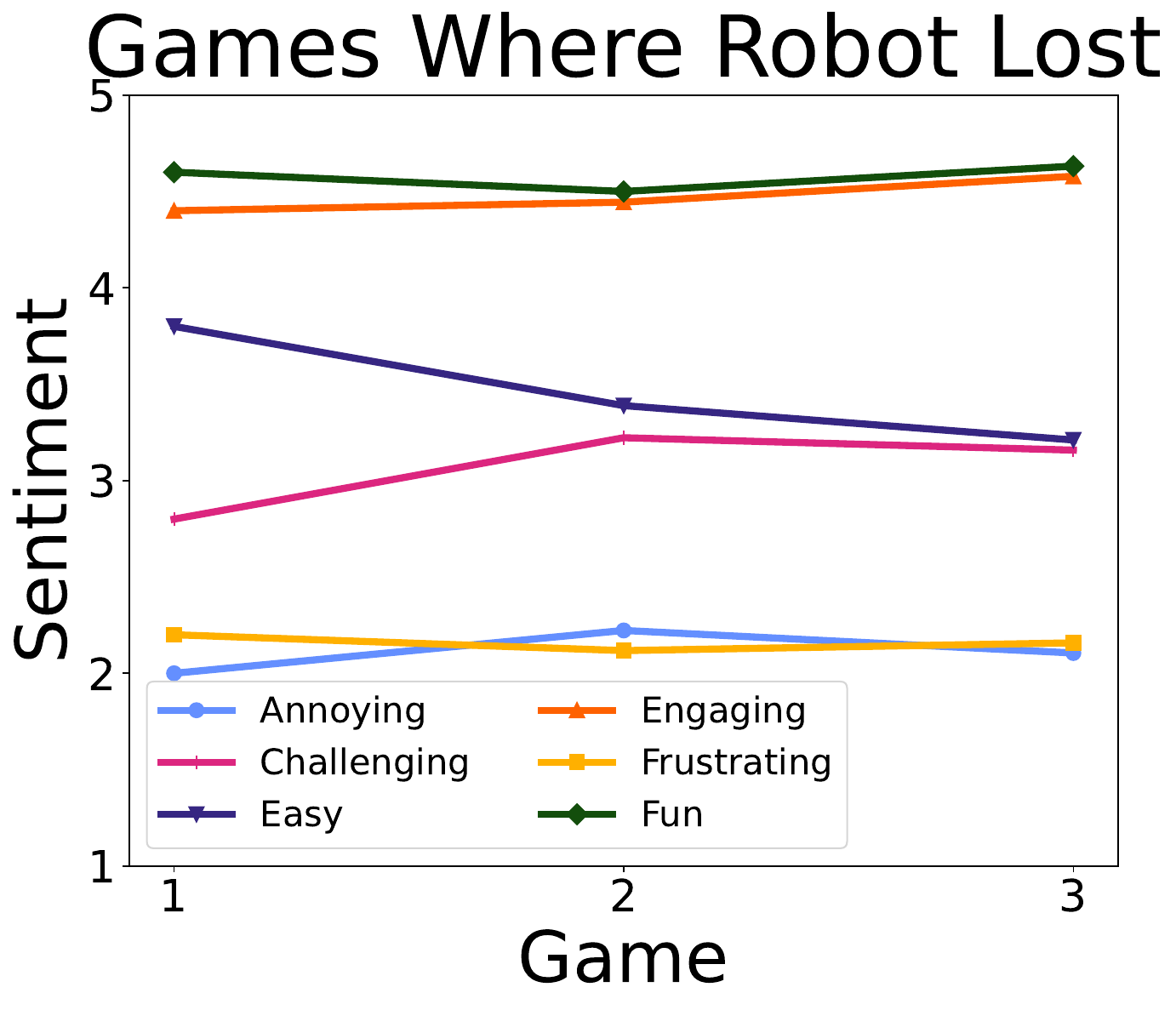}
    \includegraphics[width=0.24\textwidth]{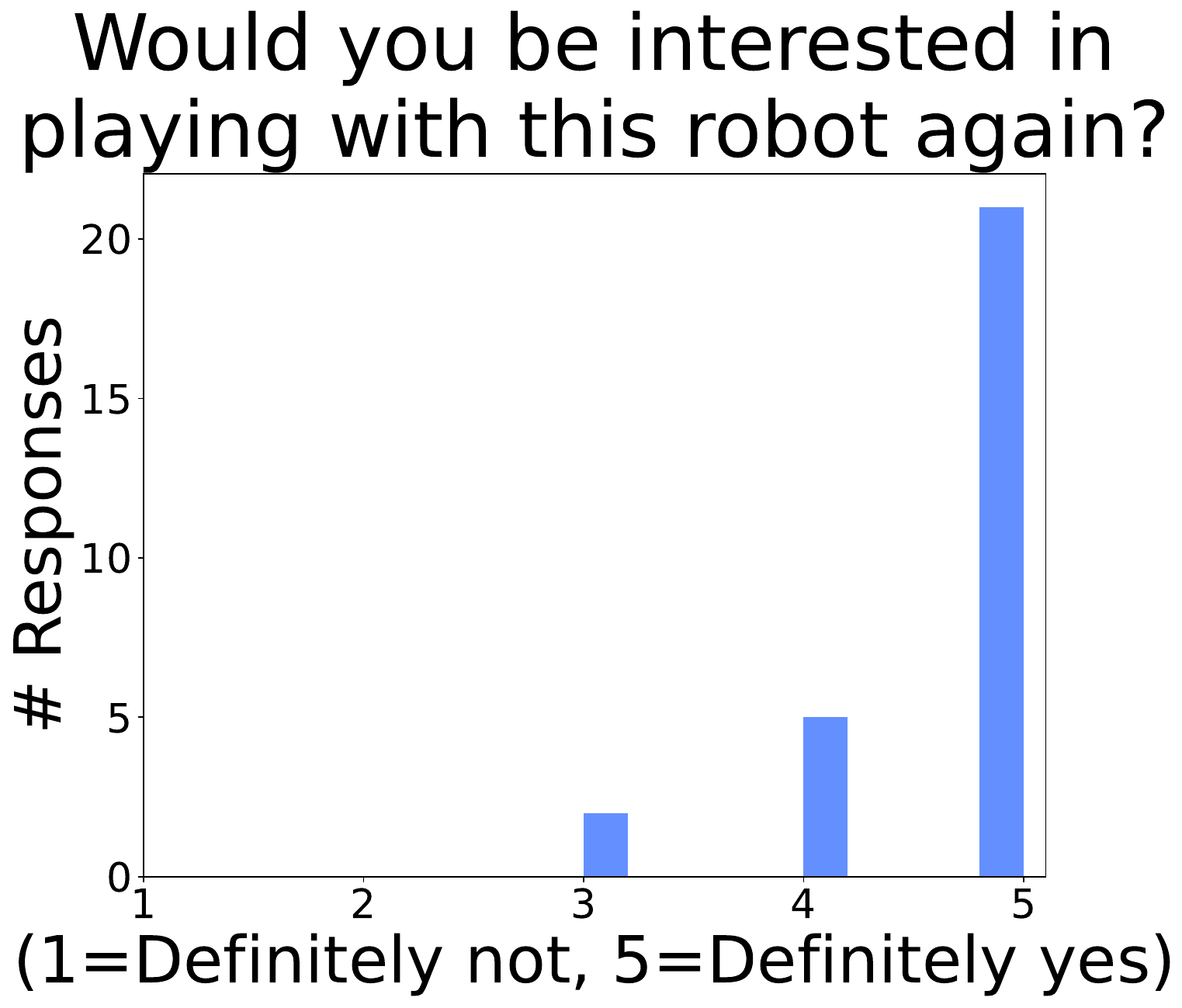}
    \caption{Players were asked ``To what degree do these words describe your experience with playing table tennis with this robot?".  These graphs show the mean player sentiment for each word across games out of a five point Likert scale with where 1 is ``Strongly Disagree" and 5 is ``Strongly Agree". The final graph is histogram of responses to the question ``Would you be interested in playing with this robot again?"}
    \label{fig:player_sentiment}
\end{figure*}

\subsection{Participant strategies and experience playing with robot}

In addition to the quantitative match results, we also wanted to understand the qualitative side of this study; what was it actually like to play against a robot?  Table tennis already has many so-called ``robots" to aid in training, but these are essentially ball launchers whereas our system has the potential to be more dynamic, is better able to mimic the playstyles of a human, and carry on a full game.

After each game, players were asked to describe any strategies they employed against the robot. Before the study, we identified several weaknesses in the robot's capabilities, most notably (1) difficulty dealing with large amounts of underspin, (2) very fast balls, (3) very low balls due to a hard-coded constraint that prevented the paddle from getting too close to the table, and (4) that the robot was physically unable to reach balls that landed very close to the net. Analyzing the post-game surveys we see that most players did not employ a specific strategy in game 1 or were mostly focused on probing the robot's capabilities.  During the second and third games, skilled players were able to identify gaps in the robot's capabilities, which correlated with higher win rates: players that mentioned ``downspin", ``backspin",``chops", or ``underspin" (synonyms for weakness 1) in their game 2 and 3 comments were significantly more likely to have won their match ($p < 0.05$) and also to be of a higher skill level ($p<0.001$). Only a few players mentioned ball speed (i.e.\ ``fast") or the robot's reach (``net").  Beginner players were less able to identify holes in the policies, typically describing simple strategies like alternating forehand and backhand or hitting directly at the robot. Some intermediate  level players were able to identify the gaps but were unable to exploit them.

To validate the players' observation that the robot is weaker at underspin balls we measured the return rate of the robot against the players by estimated spin amounts. Figure \ref{fig:spin}, shows that the robot is more robust to balls with topspin compared to underspin. The robot's return rate is reduced to 50\% with some underspin, and reaches zero quickly against higher underspin values. On the other hand, the robot can handle topspin better, still returning 60\% of the balls with topspin with estimated 80 (rad/s).

\begin{table*}[!t]
\centering
\begin{tabular}{cccccccccccccc}
\toprule
{LLC} & {Serving} & {Style} & {Policy Type} & {Skill} & {Ball Count} & \multicolumn{2}{c}{Land Rate} & \multicolumn{2}{c}{Hit Vel (y)} & \multicolumn{2}{c}{Landing (x)} & \multicolumn{2}{c}{Landing (y)} \\ 
 &  &  &  &  & Study & \multicolumn{1}{c}{Study} & Sim & \multicolumn{1}{c}{Study} & Sim & \multicolumn{1}{c}{Study} & Sim & \multicolumn{1}{c}{Study} & Sim \\ \midrule
0 & N & FH & B+A & generalist & 27 & \multicolumn{1}{c}{0.41} & 0.66 & \multicolumn{1}{c}{6.38} & 6.94 & \multicolumn{1}{c}{-0.29} & -0.09 & \multicolumn{1}{c}{0.35} & 0.88 \\ 
1 & N & FH & B+A & generalist & 69 & \multicolumn{1}{c}{0.58} & 0.65 & \multicolumn{1}{c}{5.88} & 6.79 & \multicolumn{1}{c}{-0.12} & -0.07 & \multicolumn{1}{c}{0.44} & 0.77 \\ 
2 & N & FH & B+A & generalist & 69 & \multicolumn{1}{c}{0.59} & 0.58 & \multicolumn{1}{c}{5.67} & 6.05 & \multicolumn{1}{c}{-0.21} & -0.13 & \multicolumn{1}{c}{0.45} & 0.82 \\ 
3 & N & FH & B & hit right & 4 & \multicolumn{1}{c}{0.75} & 0.63 & \multicolumn{1}{c}{5.29} & 6.81 & \multicolumn{1}{c}{-0.20} & -0.39 & \multicolumn{1}{c}{0.42} & 0.80 \\ 
4 & N & FH & B & hit left & 118 & \multicolumn{1}{c}{0.62} & 0.66 & \multicolumn{1}{c}{6.31} & 6.68 & \multicolumn{1}{c}{0.12} & 0.26 & \multicolumn{1}{c}{0.50} & 0.89 \\ 
5 & N & FH & B & hit left & 69 & \multicolumn{1}{c}{0.51} & 0.66 & \multicolumn{1}{c}{5.85} & 6.79 & \multicolumn{1}{c}{0.05} & 0.20 & \multicolumn{1}{c}{0.43} & 0.86 \\ 
6 & N & FH & B & hit left & 39 & \multicolumn{1}{c}{0.23} & 0.63 & \multicolumn{1}{c}{6.23} & 6.99 & \multicolumn{1}{c}{0.04} & 0.18 & \multicolumn{1}{c}{0.38} & 0.84 \\ 
7 & N & FH & B & hit right, fast & 87 & \multicolumn{1}{c}{0.33} & 0.73 & \multicolumn{1}{c}{6.83} & 7.69 & \multicolumn{1}{c}{-0.25} & -0.29 & \multicolumn{1}{c}{0.34} & 0.80 \\ 
8 & N & FH & B & hit left, fast & 25 & \multicolumn{1}{c}{0.44} & 0.60 & \multicolumn{1}{c}{6.76} & 7.14 & \multicolumn{1}{c}{0.11} & 0.27 & \multicolumn{1}{c}{0.72} & 0.72 \\
9 & N & BH & B+A & generalist & 93 & \multicolumn{1}{c}{0.34} & 0.67 & \multicolumn{1}{c}{5.34} & 6.66 & \multicolumn{1}{c}{0.18} & 0.11 & \multicolumn{1}{c}{0.41} & 0.82 \\ 
10 & N & BH & B & hit fast & 349 & \multicolumn{1}{c}{0.41} & 0.70 & \multicolumn{1}{c}{5.79} & 7.31 & \multicolumn{1}{c}{0.10} & -0.03 & \multicolumn{1}{c}{0.42} & 0.88 \\ 
11 & N & BH & B & hit right & 146 & \multicolumn{1}{c}{0.32} & 0.69 & \multicolumn{1}{c}{5.59} & 7.01 & \multicolumn{1}{c}{-0.10} & -0.22 & \multicolumn{1}{c}{0.37} & 0.86 \\ 
12 & N & BH & B & hit left & 91 & \multicolumn{1}{c}{0.35} & 0.61 & \multicolumn{1}{c}{5.26} & 6.99 & \multicolumn{1}{c}{0.35} & 0.36 & \multicolumn{1}{c}{0.36} & 0.75 \\ \midrule
13 & Y & FH & B & topspin & 696 & \multicolumn{1}{c}{0.75} & N/A & \multicolumn{1}{c}{5.89} & N/A & \multicolumn{1}{c}{-0.09} & N/A & \multicolumn{1}{c}{0.53} & N/A \\ 
14 & Y & BH & B & topspin & 862 & \multicolumn{1}{c}{0.60} & N/A & \multicolumn{1}{c}{5.12} & N/A & \multicolumn{1}{c}{0.14} & N/A & \multicolumn{1}{c}{0.36} & N/A \\ 
15 & Y & FH & B & underspin & 134 & \multicolumn{1}{c}{0.48} & N/A & \multicolumn{1}{c}{6.72} & N/A & \multicolumn{1}{c}{-0.37} & N/A & \multicolumn{1}{c}{0.64} & N/A \\ 
16 & Y & BH & B & underspin & 211 & \multicolumn{1}{c}{0.66} & N/A & \multicolumn{1}{c}{6.24} & N/A & \multicolumn{1}{c}{0.32} & N/A & \multicolumn{1}{c}{0.50} & N/A \\ \bottomrule
\end{tabular}
\caption{Summary of low level skill policies (LLCs). Policy types: B = base policy, A = FiLM adapter layer. Ball Count refers to the number of real world ball throws per policy used to calculate average the land rate, hit velocity in y direction and ball land position. 280k simulated ball throws per LLC were used to calculate the sim metrics.}
\label{tab:llc_summary}
\end{table*}

We also wanted to ensure that playing with the robot was actually something people would want to do. It is easy to imagine scenarios where a robot smashes every ball or repeats the same actions to not be very engaging, and so many of our design decisions (like sampling and learning from mistakes) were chosen not only to improve performance but also to create an enjoyable experience for the participants. Based on player feedback, we think this goal was achieved.  Figure \ref{fig:player_sentiment} shows that across all skill groups and win rates players agreed that playing with the robot was ``fun" and ``engaging" based on a five point Likert scale. Novelty may play some role in this assessment, but the score tends to increase slightly over games and when players were offered additional time to freely play with the robot, 26/29 of them accepted and played for a mean of 4:06 and median of 5:00 out of a maximum of five minutes, implying that there is some lasting appeal to playing with the robot. Additionally, when asked ``Would you be interested in playing with this robot again?", on a scale of one to five, the average response was 4.87 and the median response was 5.  Similarly players of all levels disagreed that playing with the robot was ``annoying" or ``frustrating" (frustrating also scored lower over time as players presumably adapted to the robot). Players gave middling scores to the robot being ``easy" and ``challenging".  When broken down by skill level, lower skilled players found the robot more challenging, but other sentiment metrics were largely consistent among groups. Post-study interviews supported these findings, with players calling the robot ``dynamic", ``fun", and ``exciting."

\subsection{LLC Skills}

LLCs provide diverse set of skills such as consistency, faster returns, targeting to specific part of the table or returning different types of balls. The LLCs used and their evaluations on hardware is represented in Table \ref{tab:llc_summary}. Based on the hardware evaluations, the land rates of LLCs vary between 0.23 and 0.75 per hit, and average hit velocities vary between 5.1 and 6.8 m/s. We can also see the lateral and horizontal diversity of the returned balls by looking at the x and y coordinates of the landing point for the returned balls. The difference between simulations and study comes from both sim-to-real gap as well as the difference in incoming balls. In simulations, we use a diverse set of incoming balls, but during the study, the players are competitive, sending more challenging balls and higher percentage of underspin hits to win the game. The increase in underspin balls during the study also explains the lower hit velocity and lower landing points in the study compared to the simulations.

\begin{figure}[!t]
    \centering
    \includegraphics[width=0.48\textwidth]{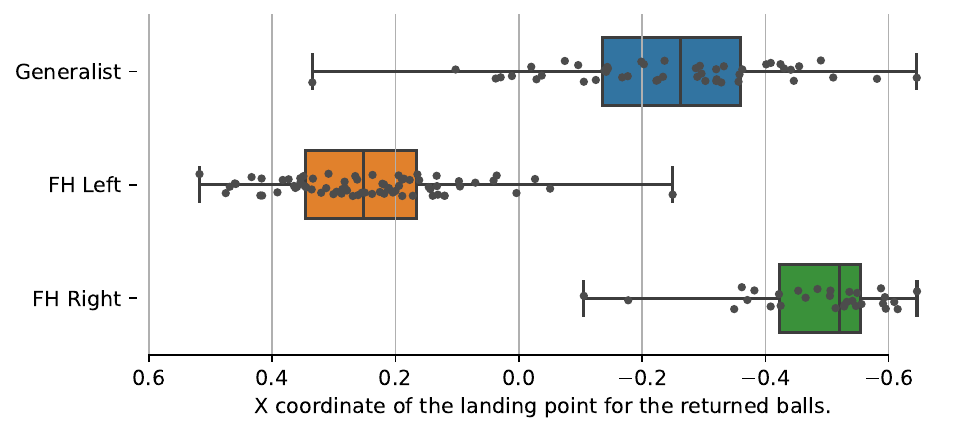}
    \caption{Distributions of robot's returned balls when using different LLCs that are trained for targeting to different edges of the table.}
    \label{fig:targeting}
\end{figure}

\begin{figure*}[!t]
    \centering
    \includegraphics[width=0.24\textwidth]{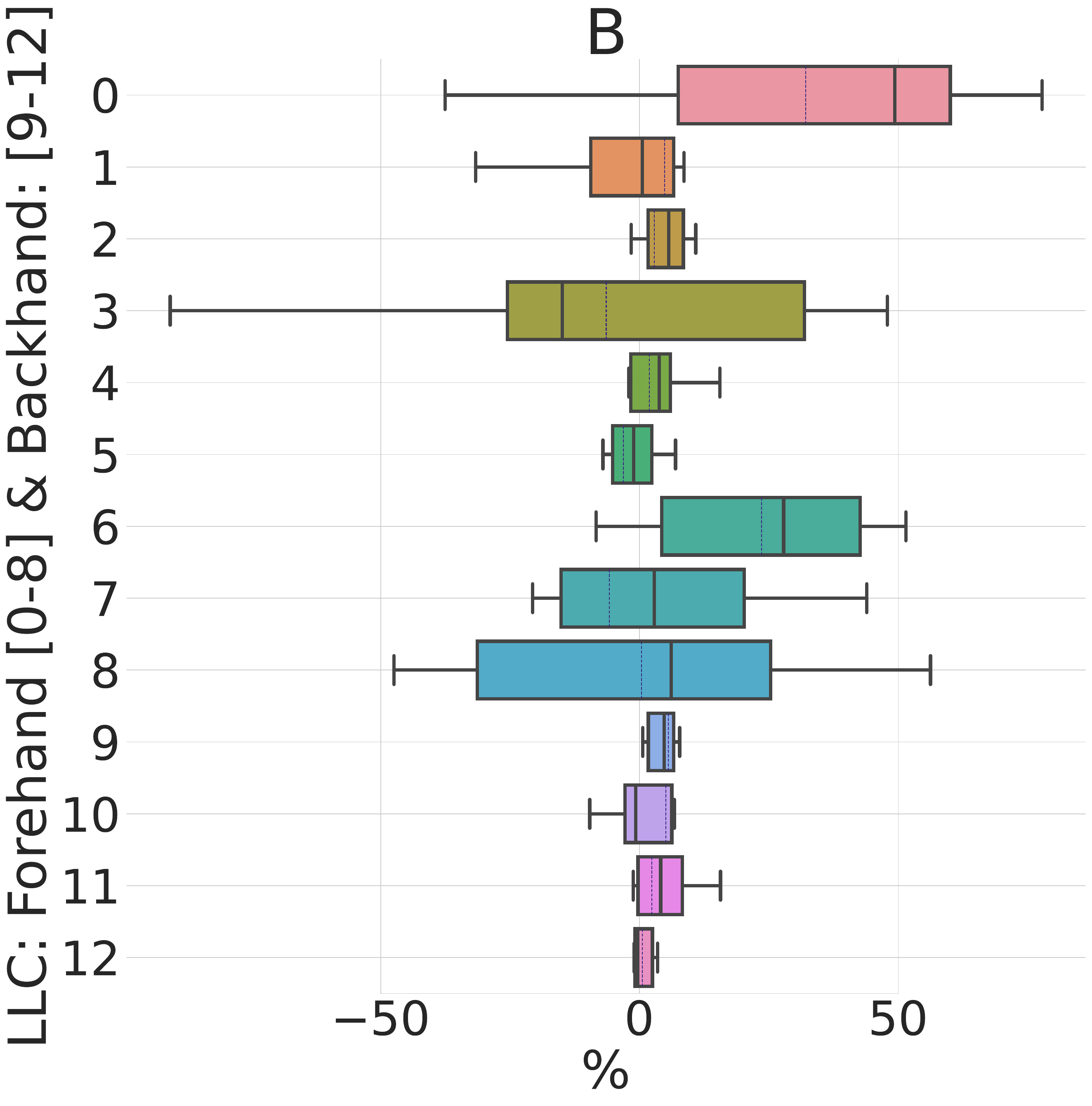}
    \includegraphics[width=0.24\textwidth]{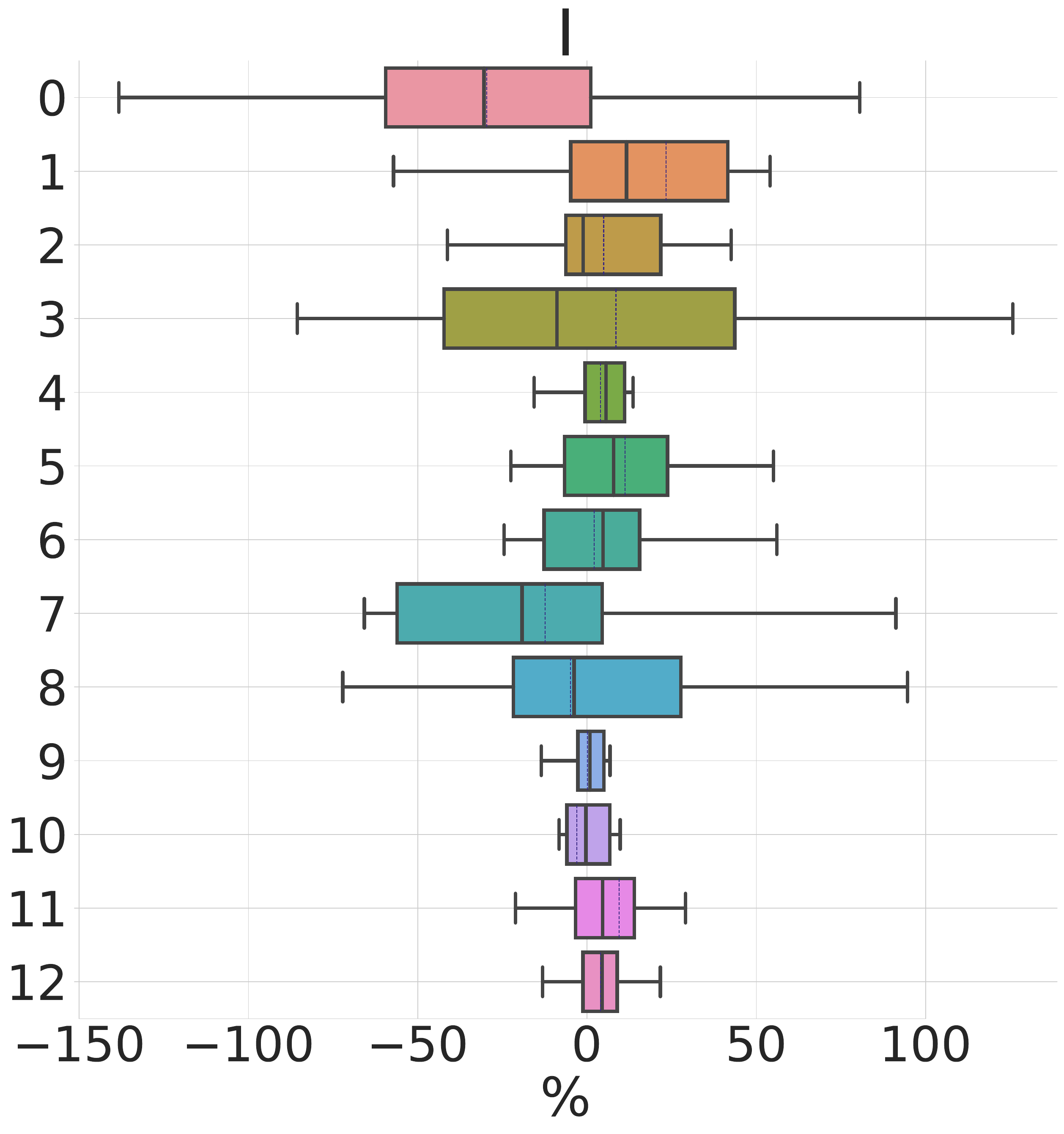}
    \includegraphics[width=0.24\textwidth]{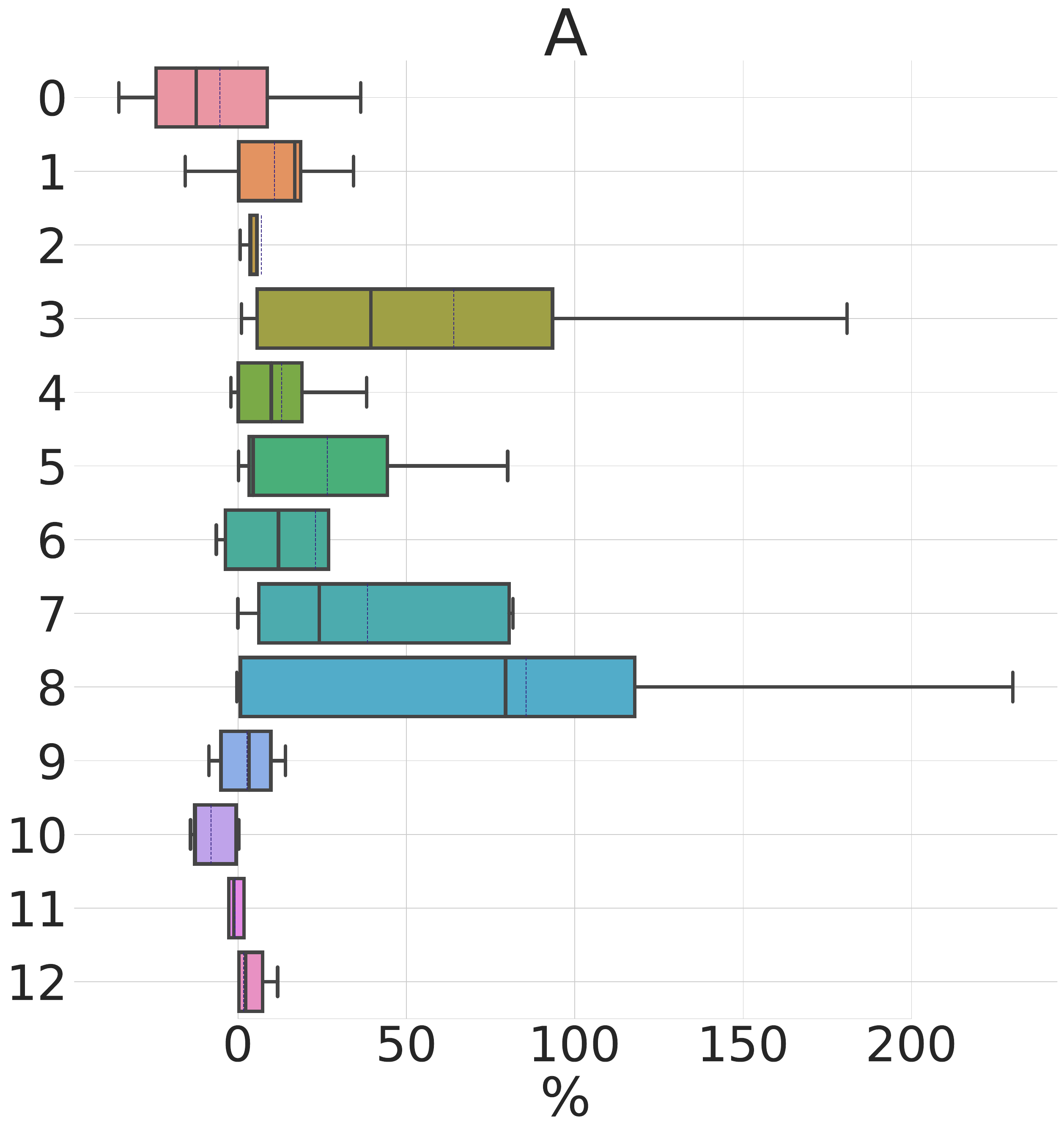}
    \includegraphics[width=0.24\textwidth]{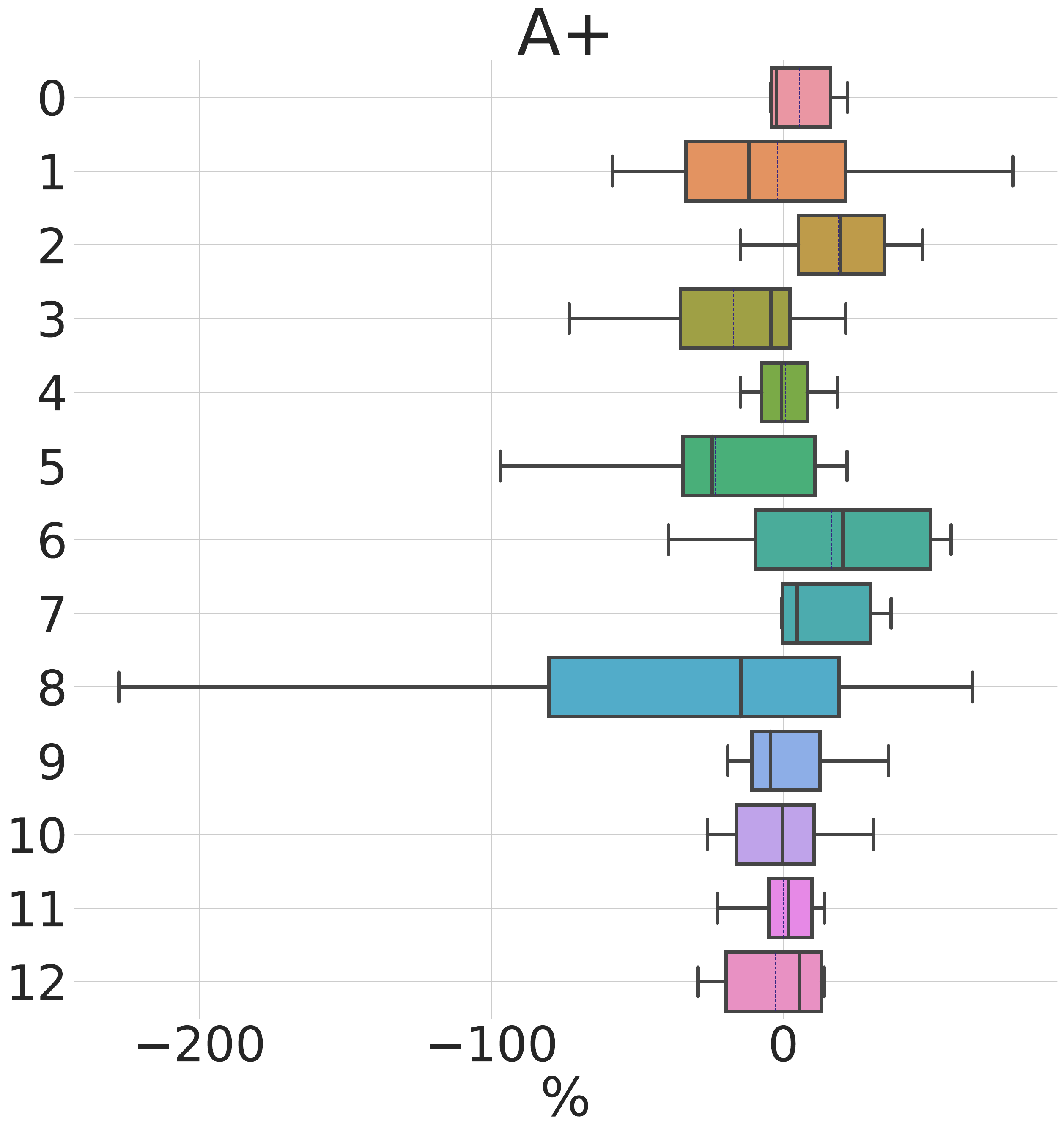}
    \caption{The percentage change in preferences (H-values) after three games for each LLC per skill levels: B = Beginner, I = Intermediate, A = Advanced, and A+ = Advanced+ skill levels.}
    \label{fig:qvalue_changes}
\end{figure*}

A subset of LLCs are specialized for targeting their returns to different parts of the table. Figure \ref{fig:targeting}, shows a comparison of one of the non-targeting LLCs with targeting ones tested on the hardware against human players. The non-targeting forehand policy returns towards the center of the table, with a slight preference on the right side using diagonal returns. The targeting policies that are obtained by further training with a different reward function that shapes their returns closer to left or right side of the table. Overall, these LLCs diversify the portfolio of the choices for the HLC and afford strategic options such as exploiting a player's weaker side or forcing the player to move from where they hit the ball.

\subsection{HLC strategy analysis}

\begin{table*}[!ht]
    \centering
    \begin{tabular}{lccccccccccccc}
    \toprule
        \textbf{LLC} & \textbf{0} & \textbf{1} & \textbf{2} & \textbf{3} & \textbf{4} & \textbf{5} & \textbf{6} & \textbf{7} & \textbf{8} & \textbf{9} & \textbf{10} & \textbf{11} & \textbf{12} \\ \midrule
        Beginner & \textbf{4.82} & \textbf{4.82} & 3.61 & 0.0 & \textbf{13.25} & \textbf{4.82} & 2.41 & \textbf{7.23} & 2.41 & \textbf{10.84} & \textbf{27.71} & \textbf{10.84} & 7.23 \\ 
        Intermediate & 3.07 & 6.36 & \textbf{7.02} & 0.0 & \textbf{11.62} & \textbf{7.46} & 2.63 & \textbf{7.46} & 1.1 & 6.58 & \textbf{28.07} & \textbf{9.87} & 8.77 \\ 
        Advanced & 4.28 & 3.74 & \textbf{6.95} & 1.07 & \textbf{12.3} & \textbf{8.02} & 2.14 & \textbf{8.02} & 1.6 & 5.88 & \textbf{26.74} & \textbf{12.83} & 6.42 \\ 
        Advanced+ & 1.98 & 5.14 & \textbf{7.91} & 0.4 & \textbf{10.28} & 3.95 & 3.16 & \textbf{6.72} & 2.77 & 9.09 & \textbf{30.43} & \textbf{11.07} & 7.11 \\ \bottomrule
    \end{tabular}
    \caption{LLC preferences for each skill level at the end of three games. Bold = top 3 forehand and top 2 backhand LLCs by skill level, including ties.}
\label{results:llc_preferences}
\end{table*}

During each match, the HLC adapts to each opponent by learning numerical \textit{preferences} (H-values) for the LLCs based on their online performance. We were interested in two questions about the adaptation component of the system; (1) how much adaptation occurred during the matches? (2) Did the agent develop different strategies depending on the skill level of the opponent?

\textbf{(1) How much adaptation occurred during the matches?} The change in H-values during a match measures the extent of adaptation. The more the H-values change, the greater the adaptation. Figure \ref{fig:qvalue_changes} shows the aggregated percentage change in H values after three games by player skill. For the forehand LLCs (ID 0 - 8) we consistently observe large changes in H-values of +/- 50\% or more, and this trend holds across skill levels. In particular, the H-values for LLCs 0 (a generalist), 3 (a right targeting policy), and 8 (a fast hitting left targeting policy) changed the most. However for the backhand LLCs (ID 9 - 12) the change in H-values was much smaller and often just a few percentage points. This indicates the HLC adapted when it played a forehand style but not the backhand. Qualitatively this is consistent with the observation from the coach that the backhand play was not at the level of the forehand during the matches.

\textbf{(2) Did the agent develop different strategies depending on the skill level of the opponent?} The H-values indicate a relative preference for a particular LLC. Since all matches began with the same initial H-values, the final H-values can be compared across skill groups to assess if the strategy differed. Table \ref{results:llc_preferences} presents these values and there was clearly some variation in strategy. For example, the H-values for LLCs 1, 2, 5, 8 and 9 all differ by a factor of $\approx2$ comparing the smallest and largest values across the player skill groups. Looking at the top LLCs per skill group (bolded), we can see that whilst there are a number of commonly favored LLCs (4, 7, 10 and 11), there are also some differences in strategy, most noticeably for beginners compared with the rest. LLCs 0, 1 and 9 were preferred for beginners whilst LLC 2 was favored for intermediate and advanced players. We can also see that the beginner skill group has the most LLCs with relatively high scores. This indicates that many LLCs are effective choices to use for beginners which is intuitive since this was the least skilled group.

\subsection{Serves}
\label{results:serves}

\begin{table}[!b]
\centering
\begin{tabular}{cccccc}
\toprule
\multicolumn{3}{c}{} & \multicolumn{3}{c}{Score (Human-Robot)} \\
Skill Level & Winner & RPLS$^{\star}$ (\%) & Game 1 & Game 2 & Game 3 \\ 
\midrule
Beginner & Robot & 43 & 7-11 & 10-12 & 4-11 \\ 
Intermediate & Human & 23 & 11-4 & 11-5 & 9-11 \\ 
Intermediate & Human & 42 & 11-4 & 11-8 & 11-6 \\ 
Advanced & Human & 36 & 14-16 & 11-2 & 11-7 \\ 
Advanced+ & Human & 36 & 11-9 & 11-5 & 11-6 \\ \bottomrule
\hline
\end{tabular}
\caption{Match details for games with points allowed on services. $\star$ RPLS = \% of points lost by robot that were lost on services.}
\label{results:rule2_scores_per_match}
\end{table}

In order to assess the service component of the game, we conducted a smaller 5-person study with a different set of rules. In these games the serve alternated every two points between the human participant and the robot. During the participant's serve, the participant could win or lose points as per normal ITTF rules. The only constraint was that serves should follow the Paralympic service rules \cite{2024_ITTF_Statutes_clean_version}. During the robot's ``serve", the participant physically served the ball but, as per the main study, could not win or lose points on the service. Only once the robot returned the ball and landed on the opponent side was the ball considered ``in play".

The match results are summarized in Table \ref{results:rule2_scores_per_match}. The robot won only 20\% matches (1/5), 33\% (5/15) games, and 43\% (117/271) points, indicating that the ability of the agent to return serves is at a lower level than its rallying skills, around that of a beginner. 

We hypothesize this is due to two factors. First, we had strict collision avoidance protocols built into the agent to avoid the paddle colliding into the table. While this was necessary to protect the equipment, it posed a significant challenge in handling serves that stay very close to the table after the bounce, such as serves with high underspin. Thus the agent missed many underspin serves. 

Additionally, we were not able to train a single LLC that could handle a broad range of spin, and in a serve, there is a higher chance of getting high spin compared to rallying. Thus, for serves we had to rely on a spin classifier along with specialist underspin and topspin LLCs to deal with the high spin serves. 

Our spin classifier had high precision of 1.0 but low recall of 0.4 for underspin serves based on testing prior to the user study, meaning that numerous underspin serves were mis-identified as topspin. We also observed this to be the case during the user study. One contributing factor to this low accuracy was the motion capture system that tracks the paddle pose was unfortunately subject to frequent failures (for example two full matches had to be discarded due to persistent failures in this component of the agent). Additionally, even when functioning, the human player can temporarily obscure the paddle leading to transient interruptions in the availability of paddle state data.

Given the underlying capabilities of the specialized topspin and underspin serving LLCs, we conjecture that enabling the paddle to move closer to the table and improving spin classification of serves in real time would close most of the gap between the serving and rallying performance of the system. In hindsight, more domain randomization of contact parameters and observation noise specific to service data might also help close the gap in rallying and serve based games.
    
\section{RELATED WORK}
\subsection{Robot sports}

Since the early days of Artificial Intelligence, Chess and other competitive games have played a critical role in the development of new algorithms and technologies.
Examples of highly-skilled AI agents can be found for backgammon \cite{td-gammon}, chess \cite{deep-blue, alphazero}, Go \cite{alphago}, Dota 2 \cite{openaifive}, poker \cite{pluribus}, and a large number of other games. In many of the these domains, agents have already achieved expert or super human-level performance by playing in a simulated or virtual environment against themselves or a human opponent. A more recent development are competitive games played between people and robotic agents in the real, physical world. In particular, robot sports games have gained substantial attention within the research community since they require fast-paced action generation in challenging, non-deterministic and partially-observable environments. A prominent example is the RoboCup competition \cite{robocup} -- a grand robotics challenge with the goal of fielding a fully autonomous robot team that can win a soccer match against the human World Cup champion team by 2050. RoboCup provides a comprehensive research domain that address real-world complexities like sensor fusion, reactive behavior learning, strategic acquisition, planning, multi-agent systems, context recognition, vision, strategic decision-making, motor control, and intelligent robot control in both simulated and hardware environments: e.g.\  \cite{10.1007/978-3-031-28469-4_24}, \cite{stone2005reinforcement}, \cite{behnke2006see}, \cite{10.1007/978-3-031-55015-7_1}, \cite{10.1007/978-3-031-55015-7_3}. Table tennis presents a research field of comparable intricacy, prompting us to devise strategies enabling direct competition with human opponents. Recently \cite{2304.13653} demonstrated that RL-trained policies could learn a number of core soccer skills such as tackling, kick and chase, getting up from the ground, and blocking opponent shots in 1 vs. 1 robot vs. robot humanoid soccer. Other examples for robot sports include fencing~\cite{yang2021competitive, yang2022motivating}, skiing~\cite{petric2012}, tennis~\cite{Zaidi_2023}, catching~\cite{abeyruwan2023agile} and badminton~\cite{mori2019}. The majority of works and competitions in robot sports focus on either sub-aspects of the game or on playing in substantially simplified settings, rather than realistic and competitive matches with humans. There is one notable exception \cite{champion-drone}, which demonstrated impressive champion level performance in drone racing on a fixed track known in advance in a head-to-head race vs. human experts. By contrast, this work focuses on an interactive sport in which the robot must continuously respond to an \textit{unseen} human's actions instead of a head-to-head time trial.

In this paper, we focus playing a competitive game of table tennis against human players of different skill-levels and under realistic conditions. Early work on robot table tennis can be traced back to  1983 with Billingsley's challenge -- a competition based on a set of simplified rules that account for technological limitations of the time~\cite{Billingsley83}. Since then, a variety of increasingly skilled table tennis agents have been proposed~\cite{andersson1988robot, Hashimoto1987DevelopmentOP, Knight1986PingpongplayingRC, schweitzer1994neural, mulling2013learning, Tebbe2018ATT, TableTennisMuscular, D_Ambrosio_2023}. A number of works have focused on aspects of action and motion generation, such as returning the ball \cite{Muelling2010LearningTTMOMP, Huang2015LearningOS, Ko2018OnlineOT, Zhu2018TowardsHL, Tebbe2020}, aiming to a specific target \cite{ding2022learning, Liu2013}, rallying \cite{abeyruwan2022sim2real, Matsushima2003LearningTT, Sun2011BalanceMG}, or even smashing \cite{LFSD-GT, TableTennisMuscular, guist2024safeaccuratespeed}. Conversely, other works have focused on aspects of state estimation inherent to the table tennis game, including, for instance, trajectory prediction \cite{Nakashima2011RoboticTT}, racket pose detection~\cite{Gao2019MarkerlessRP}, spin detection~\cite{BlankGE17, gossard2024tabletennisballspin} or the identification of human strategies~\cite{Muelling2014LearningIRL, Wang2013, Wang2017AnticipatoryAS}. The architecture and framework proposed in this paper leverages a combination of these skills in order to tackle the full competitive game for the first time.

To date, the Omron Forpheus robot \cite{Kyohei2019ThePP, Liu2013} has the closest capabilities to the agent presented in this work, demonstrating sustained rallies on a variety of stroke styles with skilled human players. A key point of difference is that our agent learns the control policies and perception system, whereas the Forpheus agent uses a  model-based approach. More specifically, Forpheus leverages rebound and aerodynamics models in order to identify the optimal configuration of the robot so as to return the ball to a target position. The Omron system represents a highly engineered system that cannot easily be customized to new players, environments, or paddles. Its main objective is to enable extended rallying with humans and, in turn, provide feedback to human player regarding their performance. In contrast to that, we aim to develop a trainable, adaptive robot agent that can play an engaging and skilled game of table tennis against a human player. To this end, we also perform a human-subject study to gauge the level of skill of the robot and its ability to engage the human partner in an interested and satisfying game. While there have been many demonstrations of robots playing table tennis against human players in the past, we believe this research is one of the first human-robot interaction studies to be conducted with full competitive matches against such a wide range of player skill levels.

\subsection{Sim-to-real}

Learning control policies in simulation using the reinforcement learning (RL) paradigm provides two main advantages. First, RL does not require any demonstrations of desired behavior, which is often time consuming and expensive to gather, and is sometimes infeasible. Second, learning through trial and error in simulation avoids any potential damage to the robot or environment that could occur as a result of random exploration. To be effective, simulated training needs to faithfully model both the real world physical dynamics and the task distribution. In table tennis, the relevant physical dynamics are the robot, paddle, table, and ball and the task is the expected distribution of incoming balls. Differences between the simulator and the real world are known as the sim-to-real gap~\cite{zhao2020}.

Minimizing the sim-to-real gap for physical dynamics is widely recognized to be a challenging problem and an extensive body of research has been dedicated to closing it in different domains. Multiple techniques have been developed, including system identification~\cite{sontakke2023}, learning hybrid simulators~\cite{jiang2021}, dynamics randomization~\cite{peng2018sim} and learning from privileged information~\cite{lee2018spigan}. The agent presented in this paper utilises many of these techniques; system ID, dynamics randomization, simulated latency, automatic resets and a policy architecture designed to encourage smooth control. For more details see \cite{D_Ambrosio_2023}. Likewise, advances in physics engine development, MuJoCo's \cite{conf/iros/TodorovET12} solid state fluid dynamics model for example, also contribute substantially to closing the sim to real gap. In addition, our agent utilizes task-conditioned physics parameters, in which some physical parameters are changed depending on the state of the environment, as well as an additional stage of simulated training to ``spin correct" and our use of sim-to-real adapter layers.

The second contribution of this work in the sim-to-real domain relates to modeling the expected human behavior, i.e. the ball distribution that humans play. We improve upon the iterative approach from \cite{abeyruwan2022sim2real} in three main ways. First we create the initial bootstrap dataset from human vs. human play instead of single hits across the table, leading to an initial starting point at a much higher level of performance. Second, zero-shot policy transfer was also essential for preventing forgetting and preserving the ball distribution knowledge from simulation, since one human playing with the robot at any particular moment in the real world typically generates a much narrower ball distribution compared to the simulated distribution. This overcomes a key weakness of \cite{abeyruwan2022sim2real} which requires fine tuning in real which not only lead to forgetting but is very time consuming. Third, the human behavior model was also changed. A spin estimate was added to the initial ball state modeling (\cite{abeyruwan2022sim2real} only uses ball position and velocity). Then, given a set of initial ball states derived from real world data, instead of transforming the states into a 6-D hyper-cube from which simulated ball states were randomly sampled (as in \cite{abeyruwan2022sim2real}), we use a non-parametric dataset-based ball distribution. This trajectory generation approach led to much better alignment between simulated ball throws and real world human play as well as more efficient training.

\subsection{Playing games in the physical world with humans}

Being able to play with and adapt to a range of \textit{diverse} human players remains an open problem. A line of research has focused on tackling this challenge in cooperative games, with the goal of building AI agents that can assist any human to perform a task. This is a long-standing problem in AI, and is known as ad-hoc team-play \cite{stone2010ad}, or zero-shot coordination \cite{hu2020other}. 
Initial attempts to first collect human data and use it to train agents that can coordinate well with humans showed a significant gap between the performance achieved with a simulated human proxy and real humans \cite{carroll2019utility}, in part because of the vast diversity of human styles and skill levels. Subsequent research has focused on using large simulated populations of self-play agents to train a policy that could coordinate well with a broad swath of players  \cite{strouse2021collaborating,charakorn2022generating,cui2022adversarial,sarkar2023diverse}. Much of this work has focused on techniques for increasing the diversity of the simulated population \cite{yu2022learning, zhao2023maximum,tang2020discovering,pugh2016quality, tjanaka2022approximating, wu2022quality,lupu2021trajectory,zhao2023maximum}. However, these works have focused on simulated games, and often do not even test with real humans. In contrast, our approach conducts extensive evaluation with real humans of diverse skill levels, in a highly dynamic, real-world setting.

\subsection{Hierarchical robot policies}
Control architectures and hierarchies play a critical role in robotics. The introduction of behavior-based robotics, for instance, is considered to be a seminal moment in robotics~\cite{brooks1986, arkin1998}. In these frameworks, a set of low-level policies form the basic behavioral building blocks which are switched or combined in order to synthesize complex robot control patterns. An arbitration module -- the arbiter -- determines which individual behavior or sub-set of behaviors to activate at any given moment, based on environmental conditions. Traditionally, both the low-level policies, as well as the arbiter are carefully engineered by an human expert to achieve the intended effect. However, foreseeing the potential interplay between behaviors under a variety of conditions is challenging and can lead to undesirable local minima~\cite{reynolds1999steering}. Accordingly, machine learning approaches have been proposed to automatically learn policies and arbitration modules. In~\cite{Rosenblatt1997} a hierarchical architecture was proposed in which each low-level policy \emph{votes} for one or multiple possible next robot actions. The set of all votes collected from the policies is then combined by the higher-level arbiter to yield a final control signal which best satisfies the objectives of all policies. This architecture already used neural network controllers for low-level policies. A brain-inspired hierarchical neural network controller was proposed in~\cite{kawato1987hierarchical} which leverages a combination of forward and inverse dynamics models for control. More recent approaches, such as HiREPS in~\cite{daniel2016}, can learn both the arbitration module and the low-level policies jointly. To this end, HiREPS first estimates the probability that an action has been created by an individual policy. Thereafter the parameters of each lower-level policy are updated based on these probabilities.

A much earlier version of our system \cite{mahjourian2019hierarchicalpolicydesignsampleefficient} demonstrated the effectiveness of a hierarchical approach to segment strategy, targeting, and execution for simulated robot table tennis, with a major difference being that the lower levels were model-based and were thus difficult to translate from simulation to the real world.

Our current work is closest in spirit to the work in~\cite{mulling2013learning} in which a gating network is learned to create mixtures of existing low-level policies. The gating network generates probabilities indicating the likelihood that a policy is the right one given the current context. The approach builds on an imitation learning framework and requires expert human demonstrations for each of the lower-level behaviors. In a similar vein, SayCan~\cite{saycan2022arxiv} uses learned value functions in order to determine if a low-level policy will be successful if executed from a given state. However, training SayCan is performed using both supervised and reinforcement learning. The overall objective of SayCan is to use a large language model for high-level planning which is the grounded via value functions and low-level policies in the real world. 
Our approach deviates from previous methodologies in that it builds upon instance based learning. Skill descriptors of the low-level policies are represented as KD-trees. Hence, new information can incrementally be added without retraining the HLC. We leverage this ability to enable rapid sim to real transfer, i.e., we collect real-world data to improve the accuracy of skill descriptors. In addition, the HLC also continuously collects information about the current opponent, without the need for retraining. As a result, our approach enables real-time learning and is highly adaptive to the current environment and opponent.
    
\section{Limitations} \label{sec:limitations}
Our research demonstrates progress in training robots to play table tennis against human opponents. However, several limitations warrant further investigation:
(1) \textbf{Fast Balls}: The robot struggles to react effectively to very fast balls due to multiple factors; lack of data, high system latency, and between-shot reset. Even with a high control frequency (50hz), the inherent latency in the system (${\sim}100ms$) significantly restricts the number of actionable decisions the robot can make within a given timeframe. This problem was exacerbated by the fact that the robot was reset to the same initial pose --- close to the table, paddle facing forward --- in between each ball hit to account for the single ball training. While increasing the control frequency could mitigate this issue, it reduces the available time per action for computation. (2) \textbf{High Balls}: The robot is unable to handle balls higher than ${\sim}6ft$ above the table, as it goes outside the camera FOV. In practice, it was seldom a problem. (3) \textbf{Low Balls}: The robot's inability to consistently handle very low balls is primarily attributed to collision avoidance protocols built into the system. These protocols, while essential for protecting the robot's paddle, pose a significant challenge in handling balls that stay close to the table. This limitation was frequently exploited by skilled players by sending underspin balls that stay very low. (4) \textbf{Short Pips Paddle}: In this work, we chose a paddle that can impart spin, while still being possible to model within the simulator. The paddle doesn't have the smooth rubber found on advanced table tennis paddles \cite{Globe889_2024}. We found accurately modeling such advanced paddles extremely challenging given the highly nonlinear and multimodal nature of the paddle. This limitation was frequently noted by human participants and highlights the need for more advanced paddle modeling techniques. (5) \textbf{Spin Detection}: The policy is unable to accurately read extreme spins on an incoming ball. This limited the amount of spin that could be effectively handled, a fact that was exploited by a number of opponents. It was possible to train specialist LLC policies to handle different types of spin, so improvements in spin detection would likely directly translate to improved overall performance. (6) \textbf{No Multi-ball Strategic Game Play}: The robot's gameplay is predominantly "one ball at a time", lacking the multi-ball strategic play observed in skilled human players. This limitation is attributed to the single-ball training approach. (7) \textbf{Motion Capture Unreliability}: The reliance on motion capture technology introduces a degree of unreliability, as occasional glitches or inaccuracies in motion tracking can negatively impact the robot's performance. (8) \textbf{Backhand} The policy had weaker backhand play compared with the forehand. (9) \textbf{Predictability} The policy did vary its placement however this could be improved further. Being less predictable in placement of certain balls was one of the top three areas for improvement as identified by the match referee (in addition to reading spin better and improving the backhand).

\section{Conclusions and Future Work}

In this work we presented a learned robot agent that achieves amateur human-level performance in competitive table tennis, a sport renowned for its dynamism and demanding skill requirements. Performance was assessed through 29 competitive matches vs. unseen human players with varied table tennis skills and who all reported to enjoy playing with the robot.

The limitations identified in Section \ref{sec:limitations} suggest many directions for future research in the area of robot table tennis. To address the latency constraints that hinder the robot's reaction time to fast balls, we propose investigating advanced control algorithms and hardware optimizations. These could include exploring predictive models to anticipate ball trajectories or implementing faster communication protocols between the robot's sensors and actuators. Additionally, training the policy to choose the reset pose, or simply removing between-shot resets altogether, would give the policy more time and flexibility to react. The challenge of low balls could be tackled by developing more sophisticated collision detection and avoidance algorithms. These algorithms could classify different potential collisions with the table and in certain cases allow for the robot's to move closer to the table whilst ensuring the safety of the paddle. The robot's strategic capabilities can be improved by training on scenarios that last an entire rally to better capture the game state, potentially even exploring self-play techniques. Additionally, advanced and advanced+ players were able to find and exploit holes in the robot's capabilities and mentioned this during interviews; we are hopeful that with our iterative learning method we could fill the gaps and adapt to these players with more training rounds, at least within the physical capabilities of the robot.  

We also hope this research makes a useful contribution beyond robot table tennis. Four aspects have broader implications.

(1) \textbf{Hierarchical policy architecture} A crucial component of performing well on complex real world tasks is having a good model of an agent's capabilities. Our LLC skill descriptors are a novel approach to building such a model both in the level of context-specific detail they provide to a higher level policy, but also because they can be continuously updated online based on real world experience. Exploring ways to incorporate such a model into the increasingly popular hierarchical robot control systems is a fruitful area for future investigation.

(2) \textbf{Enabling zero-shot sim-to-real via iterative real world data collection} Our agent used real world data to define the training tasks, whilst leveraging simulation to learn control policies. As a result it learned to solve real-world tasks whilst remaining data efficient, using only 17.5k examples. There are many challenges to using simulation and RL to train generalist robot controllers that work in the real world; namely the difficulty in scaling simulator design to many tasks, the sim-to-real gap, and the difficulty in scaling RL training compared with state-of-the art supervised learning techniques. This work offers an approach to bridging the sim-to-real gap from a task distribution perspective, and physical dynamics modeling continues to improve thanks to active research in this area. Given this, it is worth considering how to better leverage simulators. Can they be used to train general purpose skill libraries for particular embodiments?

(3) \textbf{Real time adaptation:} We enable rapid adaptation to the opponent by tracking in real-time the match statistics representing the robot and opponent’s strengths and weaknesses. This online adaptation helps the controller to adapt to novel opponents and allows the robot to learn and refine its decision-making process as the game evolves, leading to improved robustness. It is applicable to any situation in which the deployment distribution differs from the training distribution and where a policy is choosing to execute one of multiple skills whilst receiving online feedback about its success rate.

(4) \textbf{System design:} This work demonstrates that a (perhaps surprisingly) high level of performance in table tennis can be achieved using relatively simple neural network architectures and training algorithms. The policy architectures in our agent are well-known, have existed for years, and have few parameters. This suggests the importance of system design in developing highly performant learned robotic controllers. Every aspect of the system went through multiple rounds of optimization and redesign. This played a central role in the robustness and sim-to-real performance of the controller sustained over hours of gameplay. Looking forward, in order to develop both highly capable and robust robot controllers for complex real world tasks, system design may be as important as the algorithms, policy architectures, and datasets.

This is the first robot agent capable of playing an interactive competitive sport with humans at human level and represents a milestone in robot learning and control. However, it is also only a small step towards a long-standing goal in robotics of achieving human-level performance on many useful real world skills. A lot of work remains in order to consistently achieve human-level performance on single tasks, and then beyond, in building generalist robots that are capable of performing many useful tasks, skillfully and safely interacting with humans in the real world.

\section*{ACKNOWLEDGMENT}
We are very thankful to Jon Abelian, Justin Boyd, Omar Escareno, Gus Kouretas, Khem Holden, Utsav Malla, Sphurti More and Thinh Huu Nguyen for their tireless efforts helping us to maintain the system. We are so grateful to Jon Abelian, Omar Escareno, Tomas Jackson, Sphurti More, Thinh Huu Nguyen, Diego Reyes and April Zitkovich for all of their help running the user study. Thank you to Tianli Ding and Pierre Sermanet for their earlier work on the system in targeting. Finally, we are grateful to Ken Caluwaerts, Nicolas Heess, Kanishka Rao, and Martin Riedmiller for their valuable feedback on an early draft of this paper and to the entire Google DeepMind Robotics team for their help and support.

\bibliographystyle{IEEEtran}
\bibliography{bibtex/bib/IEEEabrv.bib,bibtex/bib/references.bib}{}

\clearpage
\newpage

\section*{APPENDIX}
\subsection{Contribution Statements}
\label{app:contribution_statements}

\textbf{David B.\ D'Ambrosio}$^{1,*}$: Worked on all parts of the system over the course of many years.  Developed the policy architecture and training approach.  Conceived, wrote, and edited this paper.  Helped run and analyze the user study.  

\textbf{Saminda	Abeyruwan}$^{1,*}$: Worked on all parts of the system over the course of many years.  Developed the policy architecture and training approach.  Conceived, wrote, and edited this paper.  Helped run and analyze the user study.

\textbf{Laura Graesser}$^{1,*}$: Worked on all parts of the system over the course of many years.  Developed the policy architecture and training approach.  Conceived, wrote, and edited this paper.  Helped run and analyze the user study.

\textbf{Atil Iscen}$^{1}$: Worked on targeting policies, sim-to-real, simulation modeling, results analysis.  Helped run and analyze the user study.

\textbf{Heni Ben Amor}$^{2}$: Investigated initial adapter and attenuation models, evaluated LLCs and spin classifiers, data collection, general advisor, paper writing.

\textbf{Alex Bewley}$^{2}$: Led the design and implementation for the vision system. Built components for data infrastructure and model training. Camera configuration analysis.

\textbf{Barney J.\ Reed}$^{2,\dagger},$: Expert table tennis advisor, coaching engineers, human data collection. Evaluator and referee for user study. Feedback on robot progress and skill level.

\textbf{Krista Reymann}$^{2}$: Program Manager: organized operational support for the project; initiated and managed the user study and organized operational support during the study.

\textbf{Leila Takayama}$^{2,\S}$: Experiment design, metrics development, wrote methods section, wrote HRI related work.

\textbf{Yuval Tassa}$^{2}$: Contributed to modeling and system identification of both robot and ball-flight dynamics.

\textbf{Krzysztof Choromanski}: Developed Blackbox optimization library and ES algorithms used for training table tennis policies over the course of many years.

\textbf{Erwin Coumans}: Helped setting up the initial physics simulations. Created initial virtual reality setup that allows to play tennis. Advise during research of simulation setup, URDF files, constraints setup.

\textbf{Deepali	Jain}: Developed Blackbox optimization library and ES algorithms used for training table tennis policies over the course of many years. 

\textbf{Navdeep	Jaitly}: Conceived, designed, and led the initial stages of the project, built and sourced prototypes, foundational designs of systems like control and vision.  Created initial vision pipeline and supervised algorithm development.

\textbf{Natasha	Jaques}: Initial brainstorming and ideation for policy architecture design and opponent model and paper writing.

\textbf{Satoshi	Kataoka}: Developed and maintained the custom robotics module orchestration system. Initial consultation on cameras and other infrastructure-related components.

\textbf{Yuheng Kuang}: Develops and advises on data infrastructure.

\textbf{Nevena Lazic}: Implemented the initial simulator and training algorithms.

\textbf{Reza Mahjourian}: Developed early hierarchical RL policy, defined the ball target skill, and built an initial version of the simulator.

\textbf{Sherry Moore}: Contributed to a software or hardware component that is in use today.

\textbf{Kenneth	Oslund}: Wrote and integrated the control middleware layer, which provides a unified interface for the different types of robot hardware and connects the python environment layer to the manufacturer's C++ driver.

\textbf{Anish Shankar}: Worked on earlier version of the system’s hardware and software infrastructure and system performance.

\textbf{Vikas Sindhwani}: Initiated and developed ES research agenda for table tennis and catching. Supported and advised on an ongoing basis.

\textbf{Vincent	Vanhoucke}: Executive support and research direction.

\textbf{Grace Vesom}: Built the camera driver and an early version of the ball detection pipeline. Built camera calibration software and hardened camera hardware stability.

\textbf{Peng Xu}: Robot cell design and build, prototypes for generating data.

\textbf{Pannag R.\ Sanketi}$^{1}$: Overall lead on the project (Managed the project + team). Set the vision and the research direction. Coded technical components for the project and wrote parts of the paper.
\\
\\
\textbf{Corresponding Authors}: Saminda Abeyruwan, David B. D'Ambrosio, Laura Graesser\\\texttt{\{saminda, ddambro, lauragraesser\}}@google.com
\\
\\
$^{1}$Primary contributors\\
$^{*}$Corresponding authors (order randomized, equal contributions)\\
$^{2}$Core contributors (Alphabetized)\\
$^{\dagger}$Work done at Google DeepMind via Stickman Skills Center LLC\\
$^{\S}$Work done at Google DeepMind via Hoku Labs.\\

\newpage

\begin{table*}[!ht]
    \centering
    \begin{tabular}{|l|l|l|}
    \hline
        \textbf{Category} & \textbf{Property} & \textbf{Value} \\ \hline
        Physics & Integrator & implicitfast \\ \hline
        ~ & Timestep (s) & 0.001 \\ \hline
        Fluid dynamics & density & 1.225 \\ \hline
        ~ & viscosity & 1.8e-5 \\ \hline
        ~ & wind & (0, 0, 0) \\ \hline
        ~ & fluidshape & ellipsoid \\ \hline
        ~ & Blunt drag coefficient & 0.235 \\ \hline
        ~ & Slender drag coefficient & 0.25 \\ \hline
        ~ & Angular drag coefficient & 0.0 \\ \hline
        ~ & Kutta lift coefficient & 1.0 \\ \hline
        ~ & Magnus lift coefficient & 1.0 \\ \hline
        Contact (Ball-Table) & solref & "-1000000 -17" \\ \hline
        ~ & friction & "0.1 0.1 0.005 0.0001 0.0001" \\ \hline
        ~ & solimp & "0.98 0.99 0.001 0.5 2" \\ \hline
        ~ & solreffriction & "-0.0 -200.0" \\ \hline
        Contact (Ball-Paddle) & solref/ topspin & "-268072 -103"/ "-268072 -0" \\ \hline
        ~ & friction & "1.5 1.5 0.005 0.0001 0.0001" \\ \hline
        ~ & solimp & "1.0 0.95 0.001 1.0 6" \\ \hline
        ~ & solreffriction & "-0.0 -488.0" \\ \hline
        Rubber & stiffness & 2e3 \\ \hline
        ~ & damping & 1e0 \\ \hline
        ~ & armature & 1e-3 \\ \hline
        ~ & mass & 1e-6 \\ \hline
        ~ & damping & 1e0 \\ \hline
        Actuator & $<$intvelocity/$>$ &  \\ \hline

    \end{tabular}
    \caption{MuJoCo simulator parameters.}
\label{tab:mj-si-params}
\end{table*}

\begin{table*}[!ht]
    \centering
    \begin{tabular}{|l|l|l|l|l|l|l|l|l|}
    \hline
        \textbf{Joint} & \textbf{ctrlrange} & \textbf{actrange} & \textbf{forcerange} & \textbf{kp} & \textbf{kv} & \textbf{frictionloss} & \textbf{damping} & \textbf{armature} \\ \hline
        SlideY & -2.0 2.0 & -1.8861 0.174 & -4102 4102 & 1279178.0 & 164636.0 & 0.0 & 30.924 & 0.002 \\ \hline
        SlideX & -2.0 2.0 & -1.8937 1.88 & -1499 1499 & 248791.0 & 13157.0 & 0.0 & 255.683 & 27.097 \\ \hline
        Axis1 & -4.36 4.36 & -2.0 2.0 & -100 100 & 85874.0 & 1599 & 0.1 & 25 & 0.01 \\ \hline
        Axis2 & -4.36 4.36 & 0.2 1.22 & -100 100 & 10669.0 & 265 & 0.1 & 25 & 0.01 \\ \hline
        Axis3 & -4.36 4.36 & 0.0 0.9 & -60 60 & 180694.0 & 2735 & 0.1 & 25 & 0.01 \\ \hline
        Axis4 & -5.585 5.585 & -1.5708 1.5708 & -20 20 & 21905.0 & 331 & 0.1 & 5 & 0.01 \\ \hline
        Axis5 & -5.585 5.585 & -1.75 -0.65 & -25 25 & 9000.0 & 211 & 0.1 & 3 & 0.01 \\ \hline
        Axis6 & -7.33 7.33 & -0.7 2.3 & -100 100 & 5101.0 & 66 & 0.1 & 3 & 0.01 \\ \hline
    \end{tabular}
    \caption{MuJoCo actuator parameters.}
    \label{tab:mj-actuator-params}
\end{table*}

\subsection{Simulation Details}
\label{app:simulator}

Table \ref{tab:mj-si-params} contains the MuJoCo simulator parameters and Table \ref{tab:mj-actuator-params} contains the actuator parameters.

\subsection{Details on dataset creation}
\label{app:dataset}

\begin{figure*}[!t]
    \centering
    \includegraphics[width=0.99\textwidth]{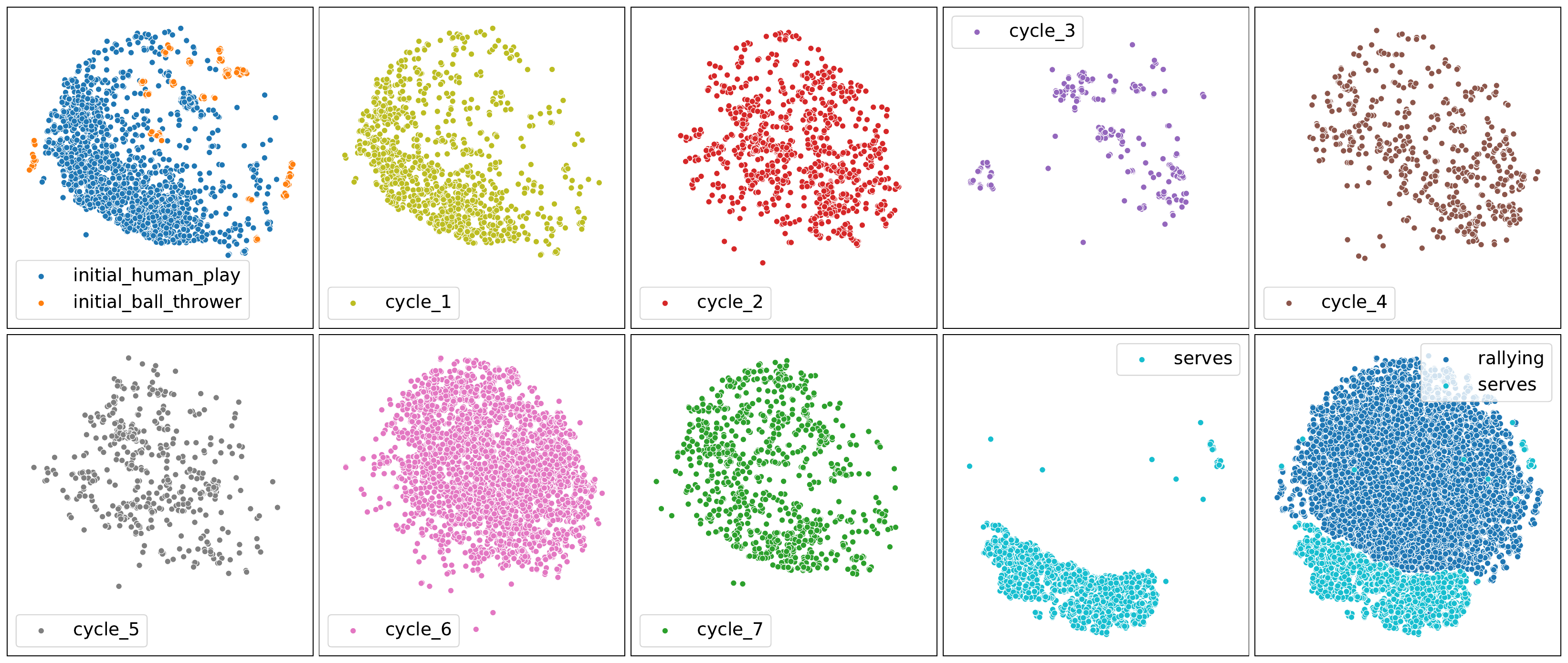}
    \includegraphics[width=0.59\textwidth]{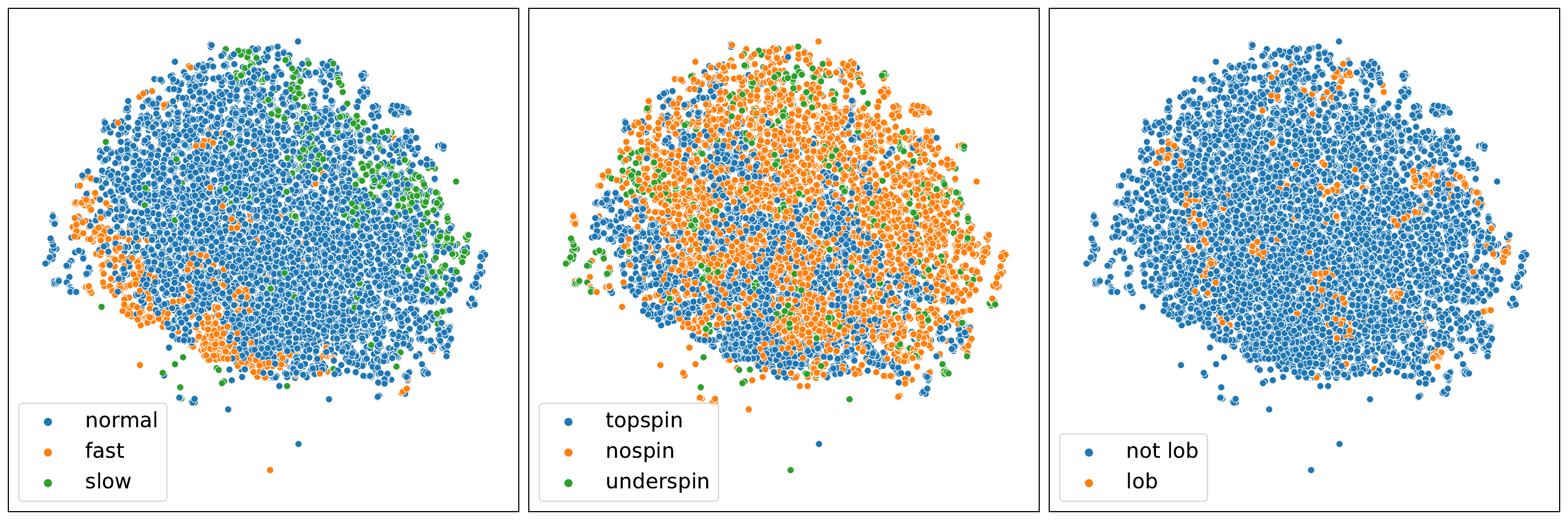}
    \caption{\textbf{Rows 1-2:} Evolution of the rallying task distribution dataset plus the serving task dataset. There are three broad clusters --- the initial dataset together with cycle 1 (r I,7), cycles 2 - 7 (r 2-7), and serves. \textbf{Rows 3: Dataset clustered by different ball types.} TSNE \cite{tsne} was used to project from 9-dimensional balls states to a 2-dimensional representation}
    \label{fig:detailed-dataset_viz}
\end{figure*}

Figure \ref{fig:detailed-dataset_viz} shows the evolution of the rallying task distribution dataset cycle by cycle. The 6th and 7th rally ball cycles were designed to close specific gaps in the agent's capabilities;
\begin{itemize}
    \item Cycle 6: The initial cycles of data collection and evaluation had focused on intermediate level play. As a result, the system had overfit to intermediate players and did not perform well vs. beginners. In this cycle we targeted evaluations with 20+ beginner players and gathered in 5.8k initial ball states of beginner level play.
    \item Cycle 7: This was the first cycle which incorporated serves. All previous cycles were focused on rallying. Evaluation were conducted with ~20 players of varying levels resulting in 1.2k rallying initial ball states and 2k services. This not only expanded our serving data but we also observed that the typical first return from a human player after the service is meaningfully different from the first ball that a player hits if they start off rallying (i.e. hit the ball across the table without an initial bounce on their side). This gap in the dataset was only revealed once a system that could support serves was deployed.
\end{itemize}

The definitions of the 7 manual sub-categories of the dataset --- Fast, Normal speed, Slow, Topspin, No spin, Underspin, Lob --- are given below:

\begin{itemize}
    \item \textbf{Fast}: Forward velocity larger than 7m/s.
    \item \textbf{Normal Speed}: Forward velocity between 7m/s and 3.5m/s.
    \item \textbf{Fast}: Forward velocity less than 3.5m/s.
    \item \textbf{Topspin}: Angular velocity ($\omega_x$) larger than 50 rad/s.
    \item \textbf{No Spin}: Angular velocity ($\omega_x$) between 50 rad/s and -25 rad/s.
    \item \textbf{Backspin}: Angular velocity ($\omega_x$) less than -25 rad/s.
    \item \textbf{Fast}: Forward velocity less than 5.1m/s and vertical velocity larger than 2.5m/s.
\end{itemize}

\subsection{LLC training details}
\label{app:llc_training}

Table \ref{table:rewards_sim_base} describes the rewards used in simulation to train base LLCs. In addition reward the robot for successfully hitting the ball (1 and 2), additional rewards are provided to encourge the robot to move safely (3-8) and to reach a particular play style pose depending on which style is being trained (9 and 10).  During the topspin correction phase, we employ two additional rewards. One of them is the net height reward (NHR), which requires the returned ball to cross the net at a specific height,

\[
    NHR(z)= 
\begin{cases}
    e^{-10 * |z-0.173|},& \text{if } 0.173 \leq z < 0.3\\
    -1.1,              & \text{otherwise}
\end{cases}
\]

and  a reward is given when the paddle reaches a target joint angle at the moment of ball contact. For the last ABB axis, we define a target joint angle. This angle is set to -0.12 for forehands and 2.0 for backhands. The reward at the moment of contact is max(1.0 - min\_dist\_to\_target, 0). The specialists LLCs are fine tuned from the topspin corrected generalist policies to implicitly target a position on the opponent side with a tolerance radius of 0.1m.

\begin{table*}[!ht]
\centering
\begin{tabular}{|l|l|l|l|}
\hline
\textbf{Reward} &  \textbf{Range}&  \textbf{Weight} &  \textbf{Weighted max score} \\
\hline
(1) State transition plus bonus for landing the ball& [0, 2] & 1 & 2  \\
\hline
(2) Bonus for hitting the ball and landing it on the opponent side of the table& [0, 1]  & 0.1  & 0.1  \\
\hline
(3) Episodic jerk reward (proxy for faulting in real)& [0, 1]  & 0.3 & 0.3\\
\hline
(4) Episodic acceleration reward (proxy for faulting in real)& [0, 1]  & 0.3  & 0.3 \\
\hline
(5) Episodic velocity reward (proxy for faulting in real)& [0, 1]   & 0.4 & 0.4\\
\hline
(6) Episodic joint angle reward (safety reward, aimed to prevent faulting in real )&[0, 1]    &1  & 1\\
\hline
(7) Safety reward, penalty for robot colliding with itself or table & [-1 * timesteps, 0]   &  1& 0 \\
\hline
(8) Paddle height reward & [-1 * timesteps, 0]   &  0.5& 0 \\
\hline
(9) Style (initial pose) reward (for forehand) &  max(1-min($\left \lVert pose_i, pose \right \rVert)$, 0)  &  1& 1  \\
\hline
(10) Style (initial pose) reward (for backhand) &  max(2-min($\left \lVert pose_i, pose \right \rVert)$, 0)  &  1& 2  \\
\hline
\textbf{Total} &    &  & [5.1 - 6.1]\\
\hline
\end{tabular}
\caption{Rewards used in simulation to train base LLCs.}
\label{table:rewards_sim_base}
\end{table*}

We represent our policy using a three layer 1D fully convolutional gated dilated CNN with 10676 parameters. Details are given in \autoref{table:model_arch}. The observation space is 2-dimensional (timesteps x [ball position, ball velocity, robot joint position, style]) which is an (8 x 16) matrix. The networks outputs a vector (8,) representing joint velocities.

\begin{table}[htbp]
\centering
\begin{tabular}{l|ccc|}
\cline{2-4}
                       & \multicolumn{3}{c|}{Layer} \\ \hline
\multicolumn{1}{|l|}{Parameter} & 1       & 2       & 3      \\ \hline
\multicolumn{1}{|l|}{Convolution dimension} &   1D      &    1D     &    1D    \\
\multicolumn{1}{|l|}{Number of filters} &   76      &    96     &    8    \\
\multicolumn{1}{|l|}{Stride} &      1   &    1     &  1      \\
\multicolumn{1}{|l|}{Dilation} &    1     &     2    &      4  \\
\multicolumn{1}{|l|}{Activation function} & tanh        &   tanh      &    tanh    \\
\multicolumn{1}{|l|}{Padding} & valid        & valid        & valid       \\ \hline
\end{tabular}
\vspace{3mm}
\caption{CNN model architecture.}
\label{table:model_arch}
\end{table}

\autoref{table:es_hypers} presents the ES hyper-parameters used for both simulated and real world training. 

\begin{table}[htbp]
\centering
\begin{tabular}{|lcc|}
\hline
 Parameter & Simulation  & Sim-to-Sim  \\ \hline
 Step size & 0.00375 & 0.00125 \\
 Perturbation standard deviation & 0.025 & 0.025 \\
 Number of perturbations & 200 & 5 \\
 Number of rollouts per perturbation & 15 & 3 \\
 Percentage to keep (top x\% rollouts) & 30\% & 60\% \\
 Maximum environment steps per rollout & 200 & 200  \\
 Use orthogonal perturbations & True & True \\
 Use observation normalization & True & True \\ \hline
\end{tabular}
\vspace{3mm}
\caption{ES hyperparameters.}
\label{table:es_hypers}
\end{table}

Table \ref{table:sensor-latency-params} details the parameters used in the simulated sensor latency model described above.

Table \ref{table:domain_radomization} details the variable used in domain randomization and the ranges.

\begin{table}[htbp]
\centering
\begin{tabular}{lcc|}
\cline{2-3}
\multicolumn{1}{l|}{}  & \multicolumn{2}{c|}{Latencies (ms)} \\ \hline
\multicolumn{1}{|l|}{Component} &   $\mu$ &         $\sigma^2$         \\ \hline
\multicolumn{1}{|l|}{Ball observation}  &      40            &  8.2                \\
\multicolumn{1}{|l|}{ABB observation}  &        29          &        8.2          \\
\multicolumn{1}{|l|}{Festo observation}  &       33           &       9           \\
\multicolumn{1}{|l|}{ABB action}  &              71    &          5.7        \\
\multicolumn{1}{|l|}{Festo action x-axis}  &            56.1      &          12.3        \\
\multicolumn{1}{|l|}{Festo action y-axis}  &            84.0      &          12.3        \\ \hline
\end{tabular}
\vspace{3mm}
\caption{Sensor latency model parameters per component.}
\label{table:sensor-latency-params}
\end{table}

\begin{table}[]
\centering
\begin{tabular}{|l|l|}
 \hline
 Variable & Range   \\ \hline
 Table damping& $\mathcal{U}(-1.0, 5.0)$    \\
 Paddle damping& $\mathcal{U}(-5.0, -1.0)$    \\
 Paddle friction & $\mathcal{U}(-0.29, 0.29)$\\
 Table friction & $\mathcal{U}(-0.05, 0.05)$ \\ \hline
\end{tabular}
\caption{Variables and ranges for domain randomization.}
\label{table:domain_radomization}
\end{table}

\subsection{HLC details}
\label{app:hlc}

Algorithm \ref{algo:q_update} presents the pseudocode for the online (i.e. during a match) LLC preference update. Table \ref{tab:wait1v3} compares waiting for 1 and 3 timesteps after the opponent hit to make a decision, and Table \ref{tab:decisivevindecisive} compares a decisive HLC (i.e. select an LLC once and stick to it for the remainder of the shot) with an HLC that has the option to re-decide every k steps. 

\begin{algorithm}
\caption{Pseudocode for LLC preference update}
\label{algo:q_update}
{\small
\begin{algorithmic}[1]
\ttfamily
\Procedure{update\_H}{$M, H, \bar{R}, Z, A, \alpha, t$}
\State \hspace{0.3cm}
\State \textbf{Input:}
\State \hspace{0.3cm} $M$: latest batch of match statistics
\State \hspace{0.3cm} $H$: current preferences, one per LLC
\State \hspace{0.3cm} $\bar{R}$: average reward
\State \hspace{0.3cm} $Z$: mask, zero-array
\State \hspace{0.3cm} $A$: action counts per LLC
\State \hspace{0.3cm} $\alpha$: step size
\State \hspace{0.3cm} $t$: timestep
\State \textbf{Output:}
\State \hspace{0.3cm} $H^*$: updated preferences, one per LLC
\State \hspace{0.3cm}
\State \Comment{$P$ = prob. each LLC is selected}
\State $P = \text{softmax}(H)$
\State \Comment{Update the H values for each shot}
\State $H^* = H$
\For{$m\; \in\; M$}
\State \Comment{$R$ = 1 if robot returned ball else 0}
\State $R = \text{get\_reward}(m)$
\State $\lambda = \text{get\_llc}(m)$
\State \Comment{Update timestep}
\State $t \mathrel{+}= 1$
\State \Comment{Update action count}
\State $A[\lambda] \mathrel{+}= 1$
\State \Comment{Update average reward}
\State $\bar{R} \mathrel{+}= (R - \bar{R}) / t$
\State $Z[\lambda] = 1$
\State \Comment{Update preferences}
\State $H^* \mathrel{+}= \alpha * (R - \bar{R}) * (Z - P)$
\EndFor
\State \hspace{0.3cm}
\State \Return $H^*$, $A$, $\bar{R}$, $t$
\State \hspace{0.3cm}
\EndProcedure
\end{algorithmic}
}
\end{algorithm}

\begin{table}[H] 
\centering 
\begin{tabular}{|l|c|c|} 
\hline 
Metric & Wait 1 Step & Wait 3 Steps \\ 
\hline 
Num balls & 89 & 89 \\ 
Hit  & 81\% & 69\% \\ 
Hit (cleared net) & 66\% & 34\% \\ 
Hit (didn't clear net) & 15\% & 35\% \\ 
Land & 39\% & 25\% \\ 
Miss & 19\% & 31\% \\ 
\hline 
\end{tabular}
\caption{HLC ablation: wait 1 vs wait 3 steps before making a decision}
\label{tab:wait1v3}
\end{table}

\begin{table}[H] 
\centering 
\begin{tabular}{|l|c|c|} 
\hline 
Metric & Decisive choice & Choose every k steps  \\ 
\hline 
Num balls & 393 & 273 \\ 
Hit  & 89\% & 75\% \\ 
HLC wrong choice & 5\% & 8\% \\ 
Indecisiveness & 0\% & 12\% \\ 
Land & 64\% & 56\% \\ 
Miss & 11\% & 12\% \\ 
\hline 
\end{tabular}
\caption{HLC ablation: decisive (commit to first decision) first select LLC every k time steps. For these experiments, k = 1.}
\label{tab:decisivevindecisive}
\end{table}

\subsubsection{Spin classifier details}

The features extracted at each timestamp for training the classifier are as follows
\begin{itemize}
\item $paddle_z[t] - paddle_z[t-3]$. (1 dimensional)
\item $paddle_{normal}[t] - paddle_{normal}[t-3]$  (3 dimensional)
\item $paddle_z[t] - ball_z[t]$ (1 dimensional)
\item $dist(paddle, ball), i.e. norm(paddle_{xyz}, ball_{xyz})$ (1 dimensional)
\end{itemize}

The feature at each timestamp is 6-d, and we stack features at 3 timestamps preceding the hit, thus making the input size $6 * 3 = 18$ for the model. We experimented with different features involving paddle and ball states, different history sizes and different subsampling strategies and found this feature vector to be the best performing one. The ground truth label for the feature is calculated using post-processing and optimizing for the spin coefficients given the full trajectory \cite{hansen2016evolution}, \cite{Triggs:1999:BAM:646271.685629} of the ball. For data augmentation, we create samples from all the timestamps within a past window of $100ms$ of the actual hit and label it with the same label as the actual hit. The augmented data size is around 7500 samples.

The policy is a 2-layer MLP of hidden sizes (128, 64). 

\subsection{User Study: Post match interview questions}

In the final interview, we asked each participant: What are your first impressions of this robot? What was it like to play table tennis against this robot? How does playing against this robot compare to playing against other non-humans, e.g., ball throwers, walls? If you could use this robot in the future, what would you use it for? Did your experience playing with the robot differ between the match and free play at the end?

\newpage
\subsection{Match Results Details}

\subsubsection{Details on skill assessment}

\begin{figure*}
    \centering
    \includegraphics[width=0.32\textwidth]{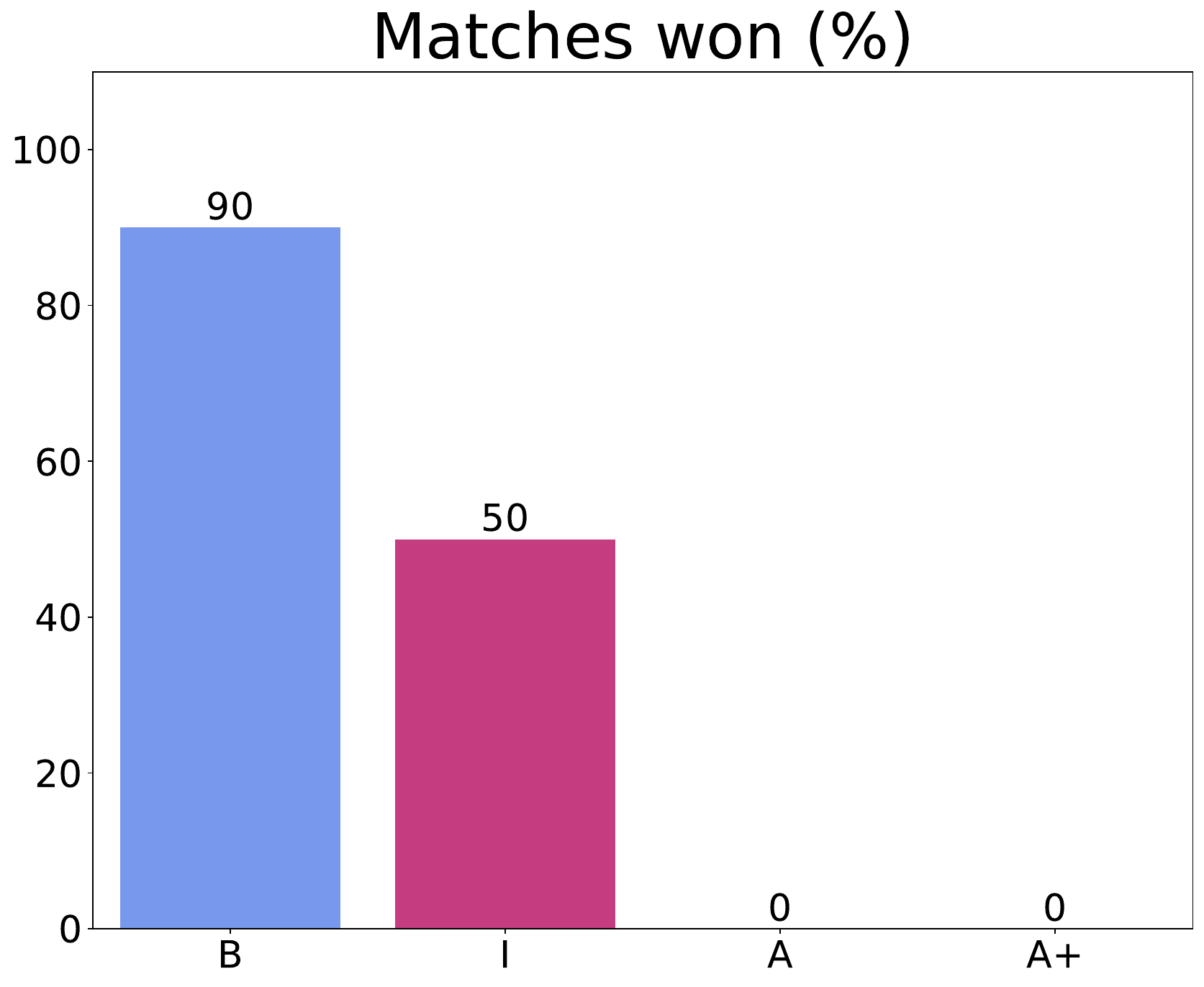}
    \includegraphics[width=0.32\textwidth]{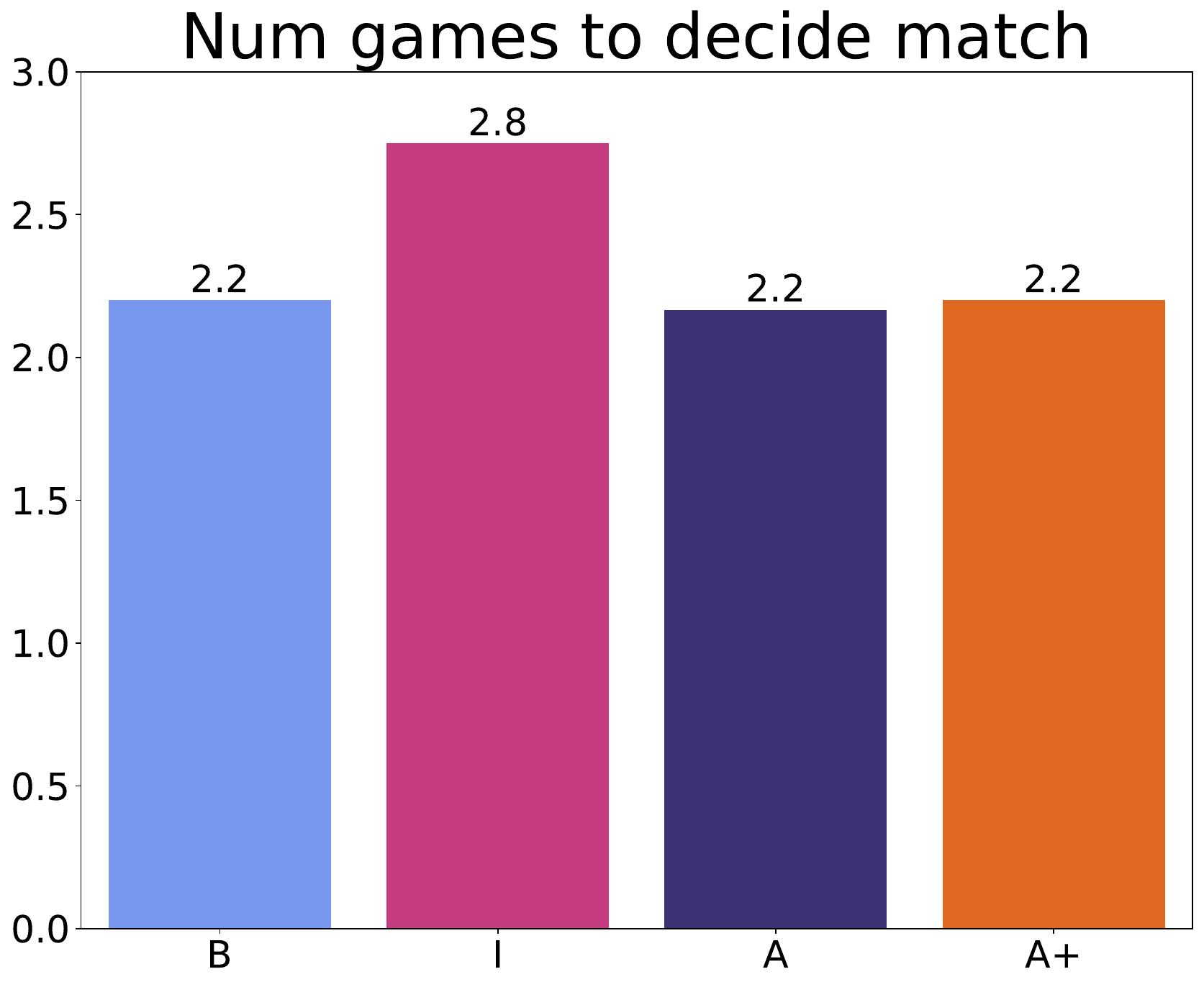}
    \includegraphics[width=0.32\textwidth]{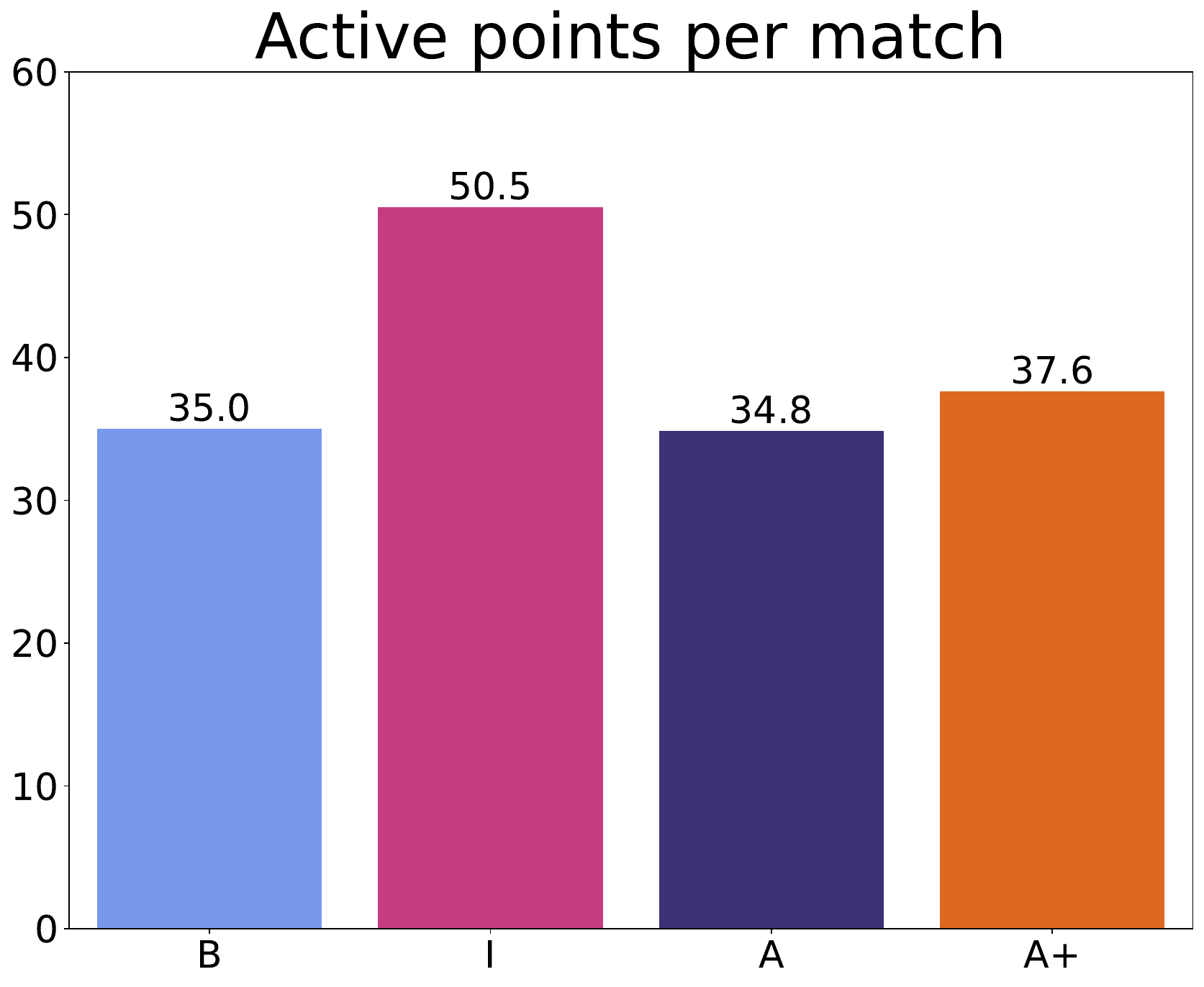}
    \includegraphics[width=0.48\textwidth]{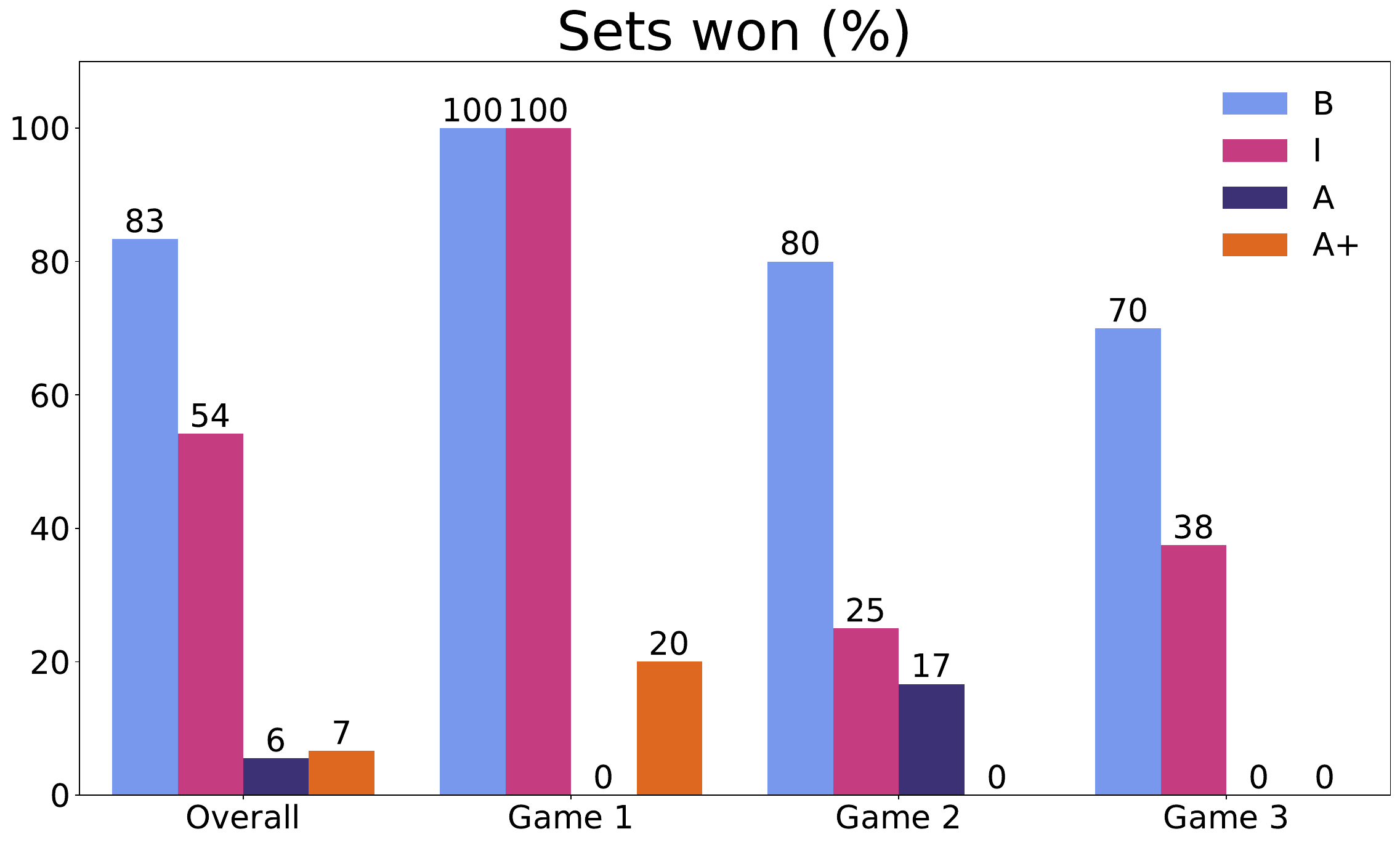}
    \includegraphics[width=0.48\textwidth]{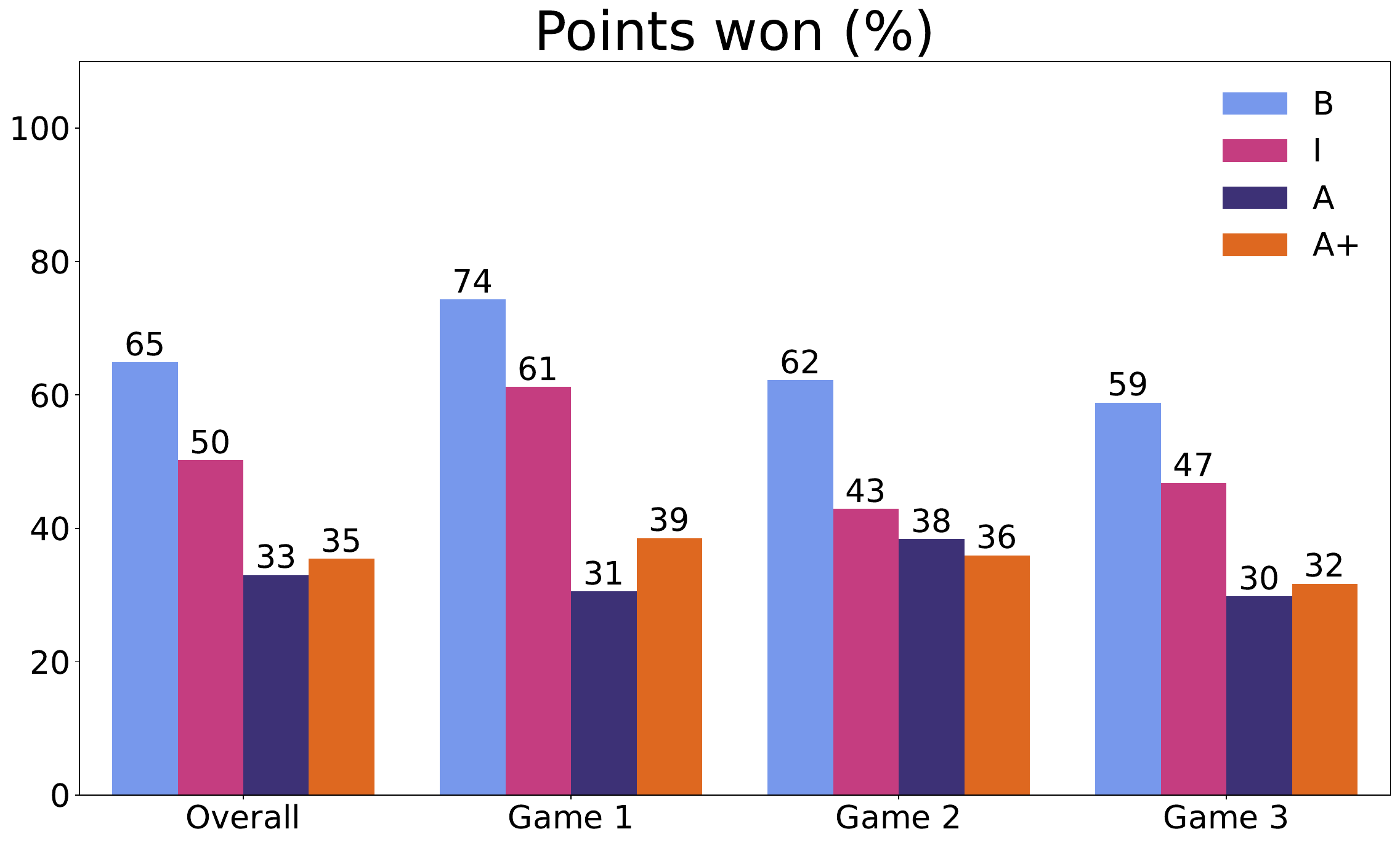}
    \caption{Match statistics using the pre-study grouping. Human opponent skill level: B = Beginner, I = Intermediate, A = Advanced, A+ = Advanced+ }
    \label{fig:match_summary_pre_study}
\end{figure*}
\newpage

\begin{figure*}[!t]
    \centering
    \includegraphics[width=0.24\textwidth]{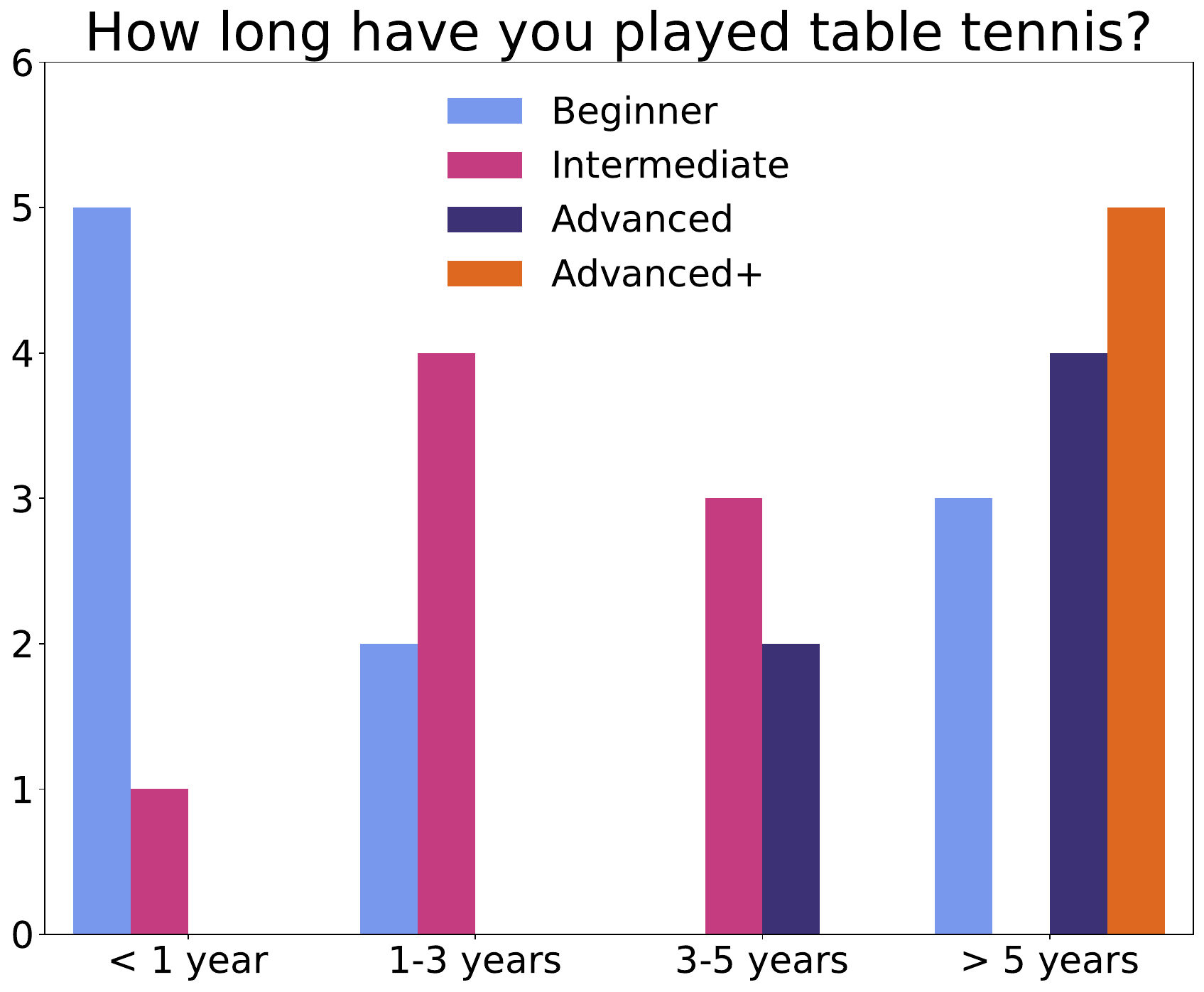}
    \includegraphics[width=0.255\textwidth]{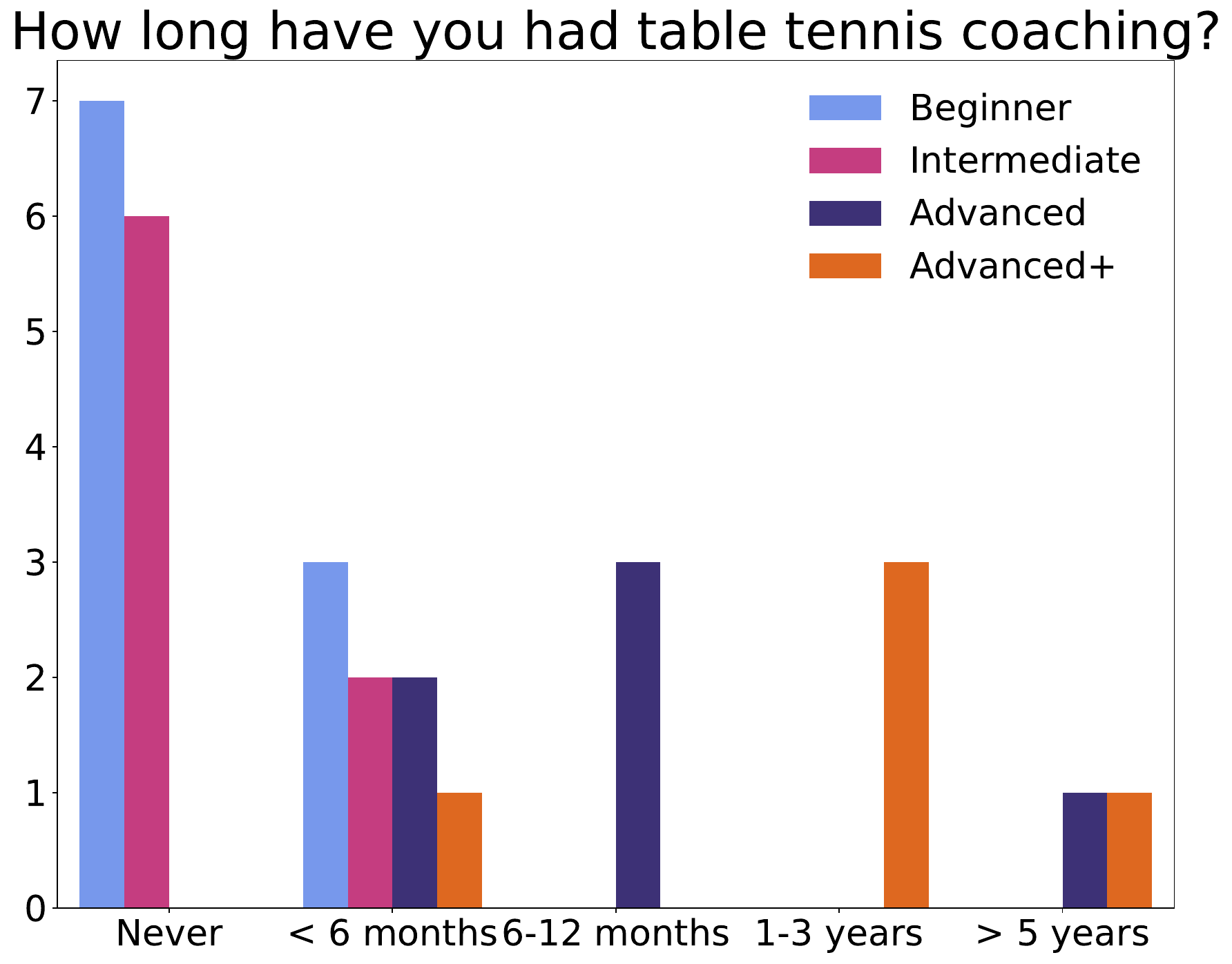}
    \includegraphics[width=0.24\textwidth]{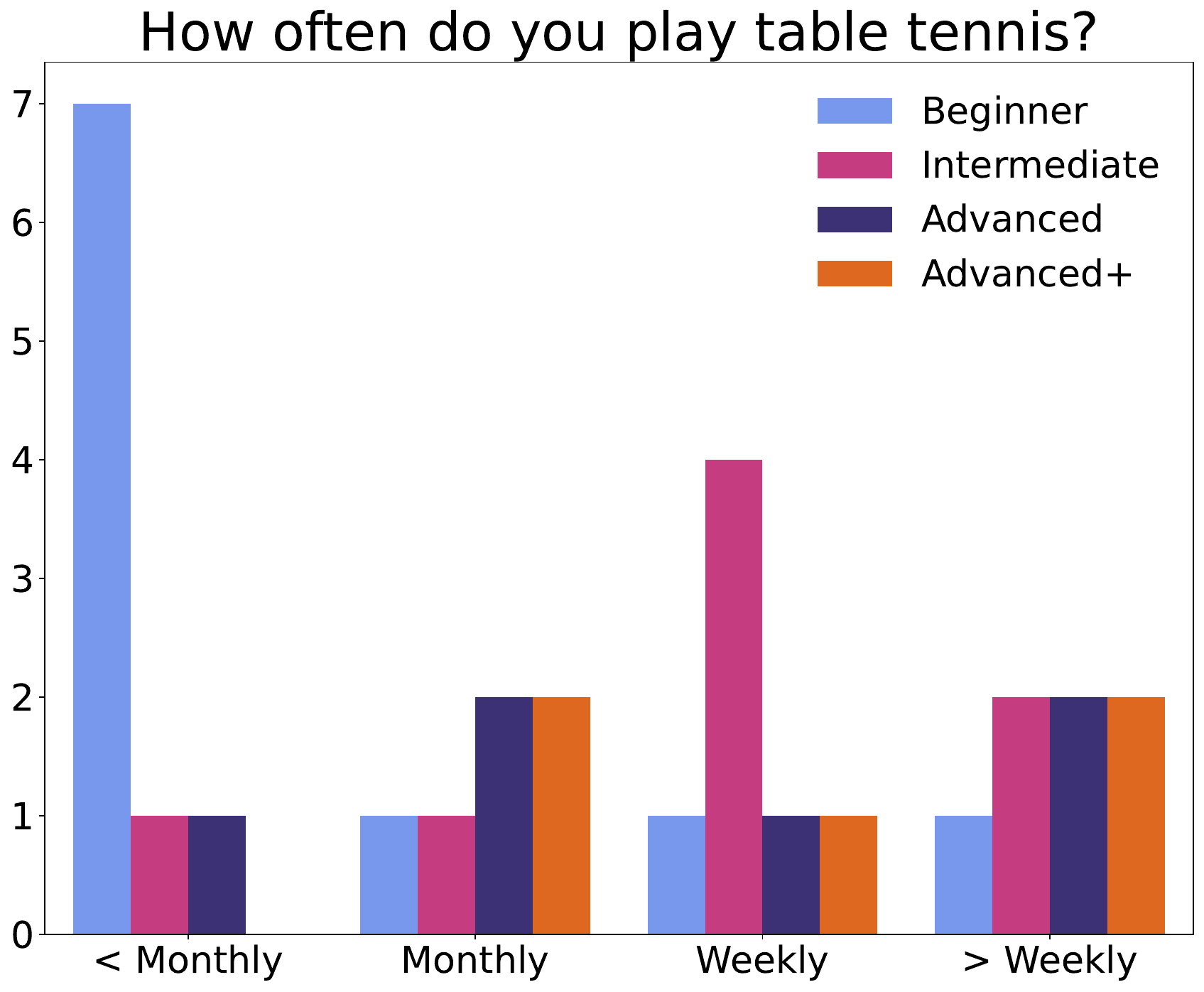}
    \includegraphics[width=0.24\textwidth]{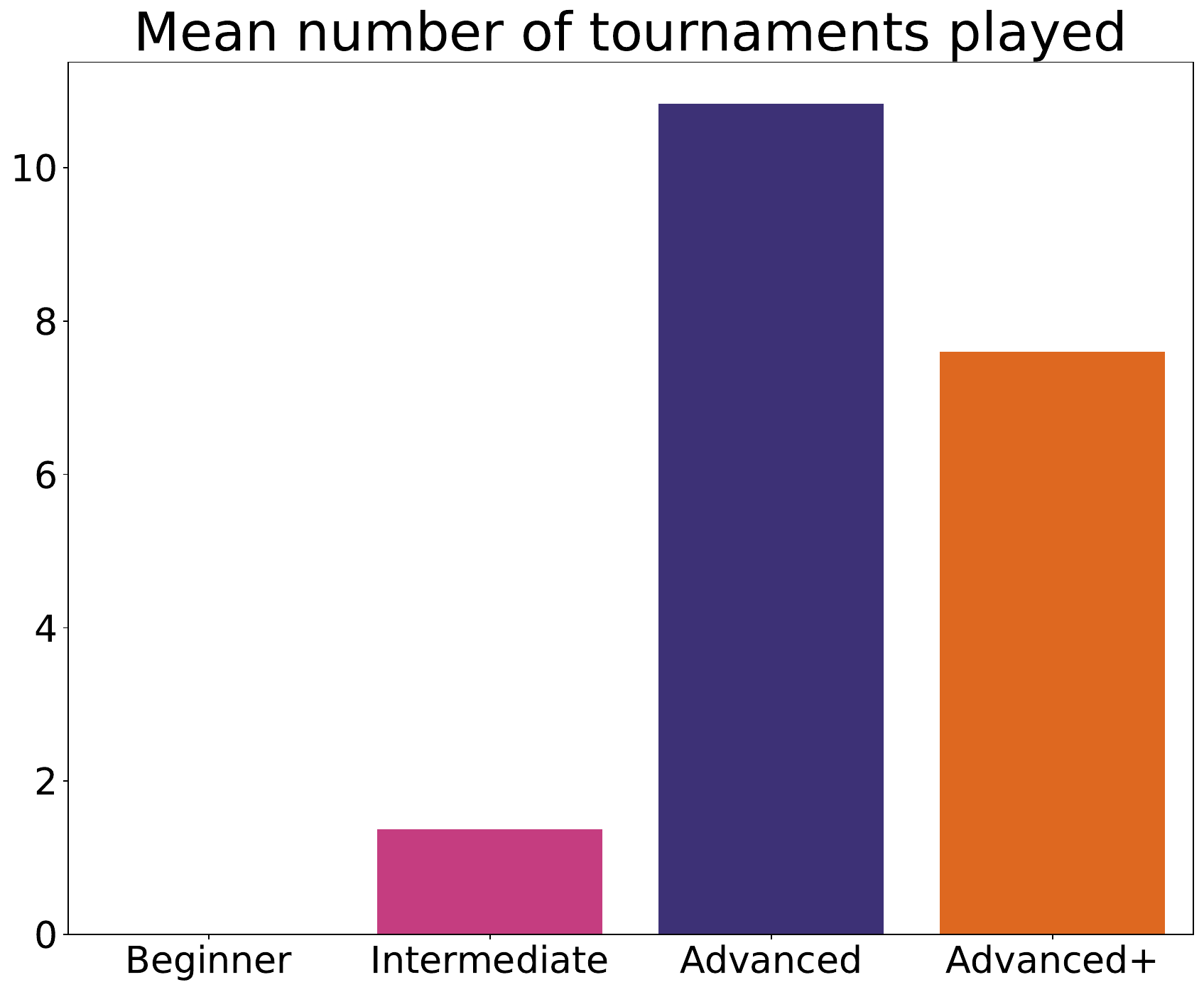}
    \includegraphics[width=0.25\textwidth]{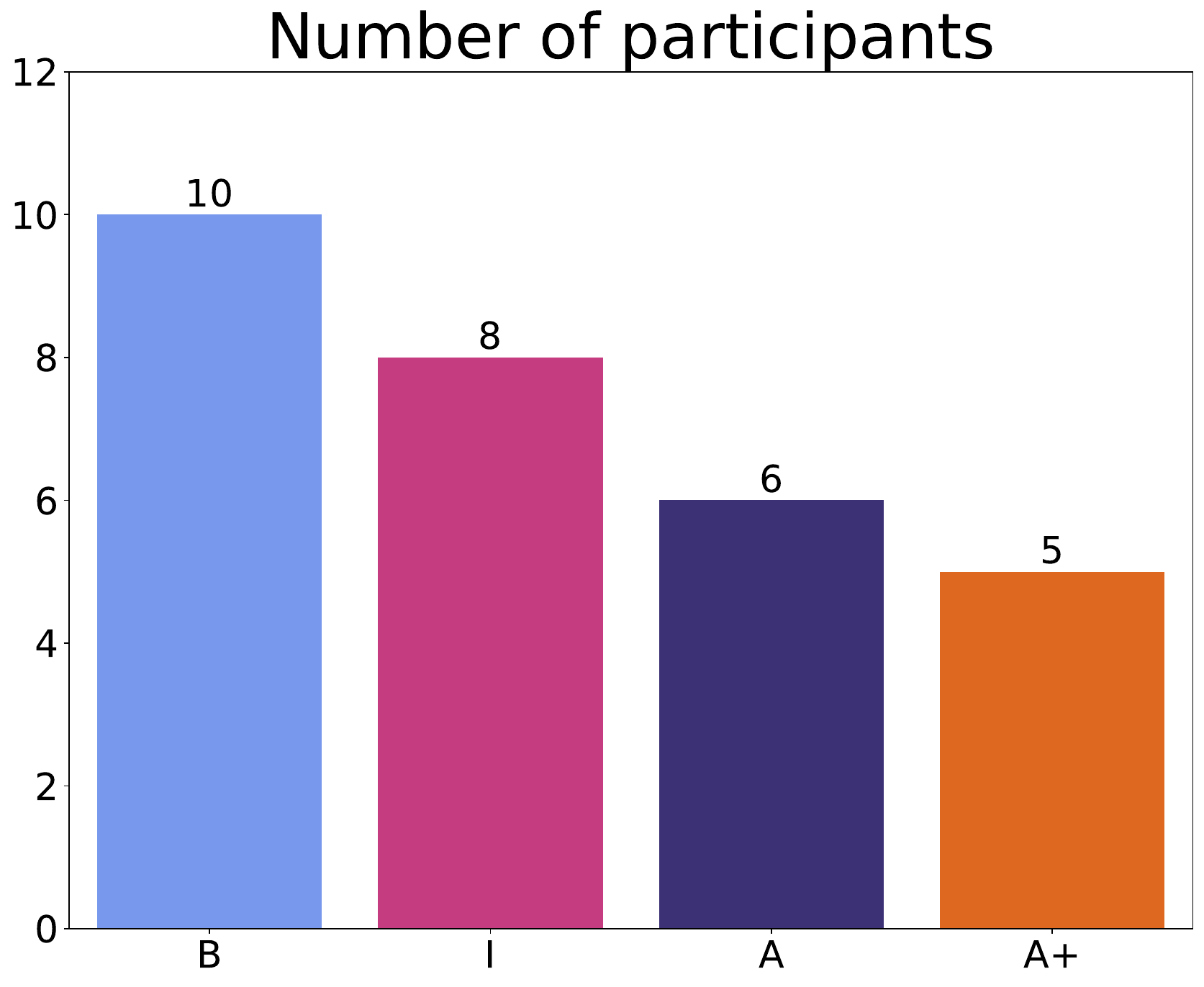}
    \includegraphics[width=0.25\textwidth]{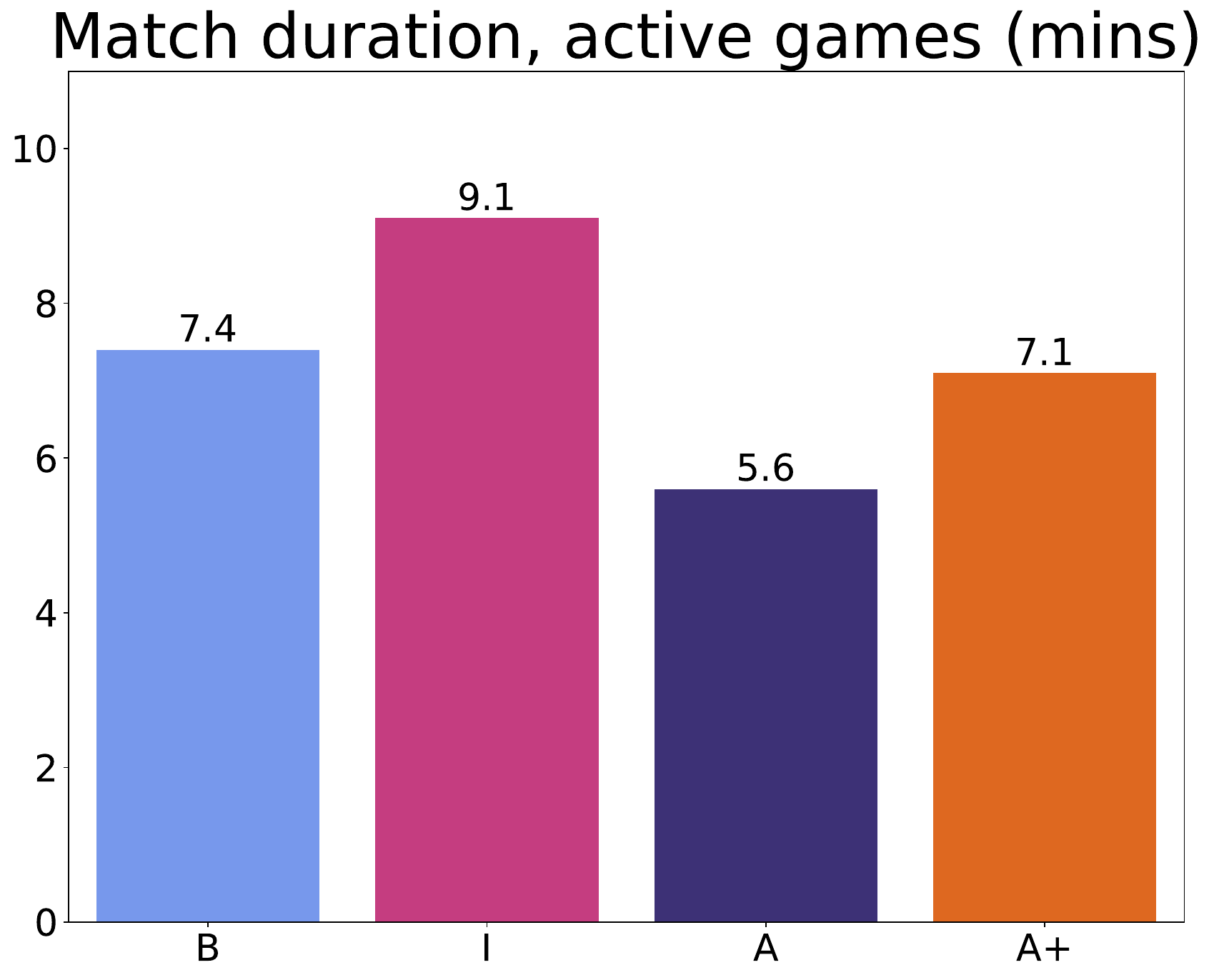}
    \caption{\textbf{Top row:} Participants' experience playing table tennis by group using the pre-study grouping. \textbf{Bottom row:} Matches played by skill group and average match duration using the pre-study grouping. Human opponent skill level: B = Beginner, I = Intermediate, A = Advanced, A+ = Advanced+ }
    \label{fig:group_stats_pre_study}
\end{figure*}

During the matches the referee checked that the pre-skill assessment accurately reflected a player's performance. The majority of players remained in the same category, 4 / 29 players changed category. 3 players with a pre-skill assessment of beginner were re-assessed to intermediate and one advanced player was re-assessed to advanced+. During the pre-skill assessment, the referee only had the opportunity to assess participants for a few minutes and it was not in a match setting, and so it was possible to mis-judge their skill level. Additionally, we think that a data entry mistake was made for the advanced player that was re-assessed to advanced+. 

The during match (i.e. post re-assessment) skill levels were used for the results reported in the main paper because this more accurately reflects a player's skill. Figure \ref{fig:match_summary_pre_study} presents match results using the pre-study assessment. All of the trends remain the same and none of the implications from the main results are affected.

Figure \ref{fig:group_stats_pre_study} presents players' table tennis experience by pre-study assessment skill group and the distinctions between groups are less clear compared with Figure \ref{fig:group_stats}. For example, there is a less clear distinction between beginner and intermediate players based on how long they have played table tennis and how often they play. This gives us an additional reason to think that the during match skill group assessment is the most appropriate to use.

\begin{table}[!h]
\centering
\begin{tabular}{ccccc}
\toprule
\multicolumn{2}{c}{} & \multicolumn{3}{c}{Score (Human-Robot)} \\
Skill Level & Winner & Game 1 & Game 2 & Game 3 \\ 
\midrule
Beginner & Robot & 5-11 & 5-11 & 6-11 \\ 
Beginner & Robot & 1-11 & 4-11 & 3-11 \\ 
Beginner & Robot & 0-11 & 2-11 & 3-11 \\ 
Beginner & Robot & 9-11 & 8-11 & 11-6 \\ 
Beginner & Robot & 0-11 & 1-11 & 4-11 \\ 
Beginner & Robot & 6-11 & 8-11 & 3-11 \\ 
Beginner & Robot & 0-11 & 2-11 & 5-11 \\
\midrule
Intermediate & Robot & 6-11 & 11-9 & 8-11 \\ 
Intermediate & Human & 6-11 & 11-4 & 11-8 \\
Intermediate & Human & 10-12 & 11-6 & 11-9 \\ 
Intermediate & Robot & 4-11 & 13-15 & 12-14 \\ 
Intermediate & Robot & 5-11 & 10-12 & 11-2 \\ 
Intermediate & Robot & 8-11 & 11-6 & 9-11 \\ 
Intermediate & Human & 10-12 & 11-6 & 11-5 \\ 
Intermediate & Robot & 8-11 & 11-9 & 6-11 \\ 
Intermediate & Human & 3-11 & 11-8 & 11-7 \\ 
Intermediate & Robot & 5-11 & 6-11 & 12-10 \\ 
Intermediate & Human & 9-11 & 11-2 & 11-6 \\ 
\midrule
Advanced & Human & 11-3 & 6-11 & 11-4 \\ 
Advanced & Human & 11-8 & 11-7 & 11-7 \\ 
Advanced & Human & 11-4 & 11-3 & 11-2 \\ 
Advanced & Human & 11-9 & 11-4 & 11-6 \\ 
Advanced & Human & 11-2 & 11-8 & 11-5 \\ 
\midrule
Advanced+ & Human & 11-7 & 11-9 & 11-5 \\ 
Advanced+ & Human & 11-3 & 11-2 & 11-2 \\ 
Advanced+ & Human & 11-2 & 11-1 & 12-10 \\ 
Advanced+ & Human & 11-9 & 13-11 & 11-4 \\ 
Advanced+ & Human & 7-11 & 11-9 & 11-5 \\ 
Advanced+ & Human & 11-3 & 11-5 & 11-4 \\ \bottomrule
\hline
\end{tabular}
\caption{Match details for games with the main rules.}
\label{app:scores_per_match}
\end{table}


\addtolength{\textheight}{-12cm}   

\end{document}